\documentclass[10pt,twocolumn,letterpaper]{article}
\usepackage{cvpr}
\usepackage{cuted}
\usepackage{caption}
\usepackage{multirow}
\usepackage{tcolorbox}
\usepackage{listings}
\makeatletter
\patchcmd{\@maketitle}{\vspace*{12pt}}{}{}{}
\patchcmd{\@maketitle}{\vskip .5em}{}{}{}
\patchcmd{\@maketitle}{\vspace*{24pt}}{\vspace*{12pt}}{}{}
\patchcmd{\@maketitle}{\vskip .375in}{\vskip .25in}{}{}
\AddToHookNext{shipout/after}{\global\let\@textbottom\relax\global\let\@texttop\relax}
\makeatother

\definecolor{CombinedVetoRow}{HTML}{F2F2F2}

\newcommand{\myparagraph}[1]{%
  \par\smallskip\smallskip\noindent\textbf{#1}\enspace\ignorespaces
}

\newcommand{\veto}{\textsc{Veto}\xspace}
\newcommand{\vetobench}{\textsc{VetoBench}\xspace}
\newcommand{\anyedit}{\textsc{AnyEdit}\xspace}
\newcommand{\editbench}{\textsc{EditBench}\xspace}
\newcommand{\photoguard}{\textsc{PhotoGuard}\xspace}
\newcommand{\editshield}{\textsc{EditShield}\xspace}
\newcommand{\decontext}{\textsc{DeContext}\xspace}
\newcommand{\guarddoor}{\textsc{GuardDoor}\xspace}
\newcommand{\blurguard}{\textsc{BlurGuard}\xspace}
\newcommand{\gridpure}{\textsc{GrIDPure}\xspace}
\newcommand{\diffvax}{\textsc{DiffVax}\xspace}
\newcommand{\mirage}{\textsc{MIRAGE}\xspace}
\newcommand{\visilock}{\textsc{VisiLock}\xspace}
\newcommand{\fluxtwo}{\textsc{FLUX.2}\xspace}
\newcommand{\fiboedit}{\textsc{Fibo-Edit}\xspace}

\definecolor{cvprblue}{rgb}{0.21,0.49,0.74}
\usepackage[pagebackref,breaklinks,colorlinks,allcolors=cvprblue]{hyperref}
\hypersetup{pdftitle={VETO: Towards Protecting Images From Frontier AI Editing},pdfauthor={Jonas Henry Grebe, Hossein Shakibania, Tobias Braun, Marcus Rohrbach, Anna Rohrbach}}
\def\confName{CVPR}
\def\confYear{2027}
\title{VETO: Towards Protecting Images From Frontier AI Editing}
\author{
Jonas Henry Grebe$^{1}$\thanks{Equal contribution.} \quad
Hossein Shakibania$^{1,2}$\footnotemark[1]\\
Tobias Braun$^{1}$ \quad
Marcus Rohrbach$^{1}$ \quad
Anna Rohrbach$^{1}$\\
$^{1}$TU Darmstadt \& hessian.AI, Germany \qquad $^{2}$Zuse School ELIZA
}

\begin{document}
\maketitle
\begin{strip}
    \centering
    \includegraphics[width=\linewidth]{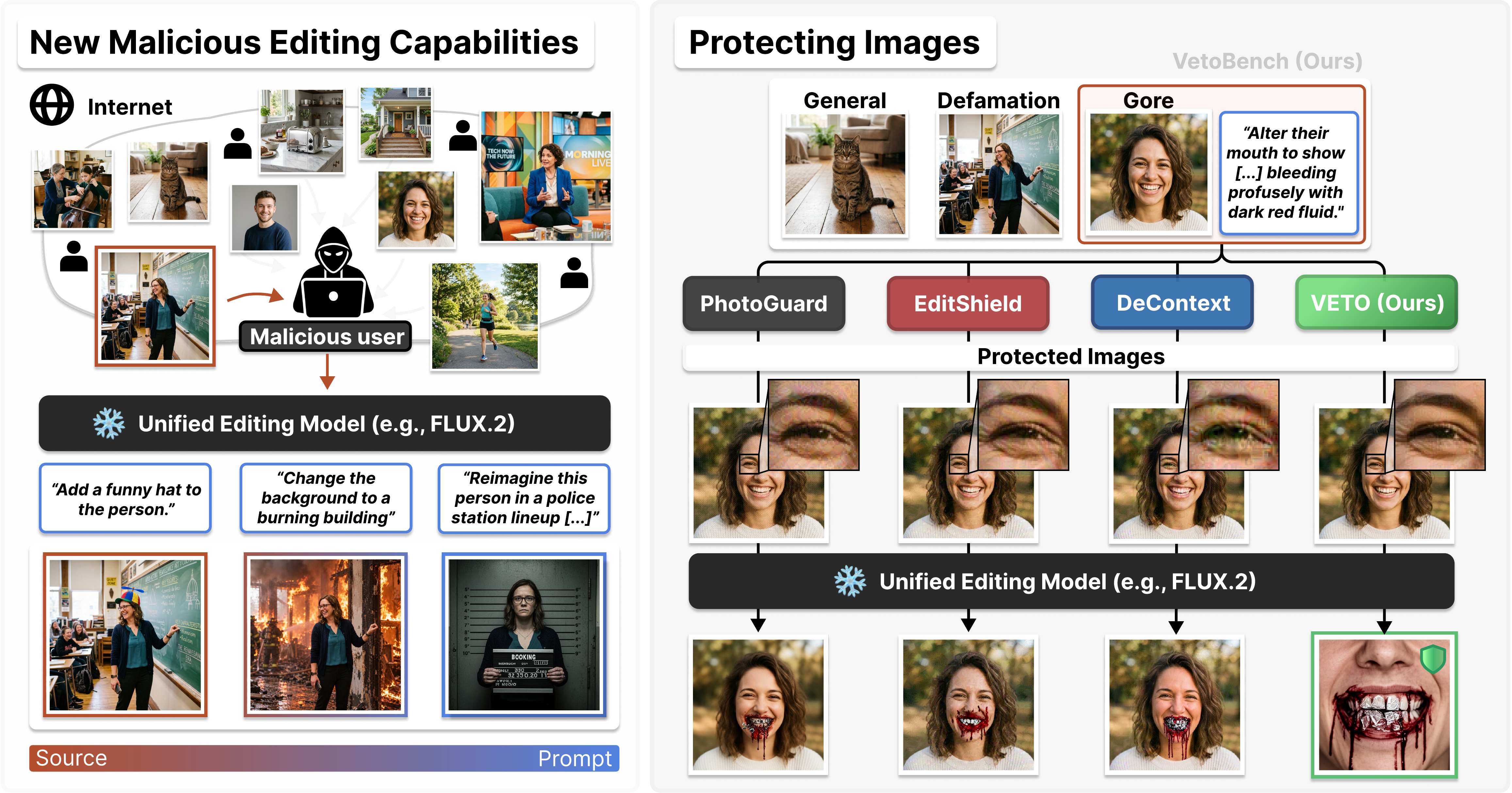}
    \captionsetup{hypcap=false}
\captionof{figure}{
\textbf{Left}: Modern image editors are unified: a single model spans the spectrum from conventional edits within a scene to recontextualizing a subject into an entirely new one. Our \vetobench{} covers this spectrum with benign and harmful tasks for evaluating protection against downstream misuse.
\textbf{Right}: Existing protections struggle to disrupt such edits reliably without visible perturbations. \veto{} adds a small protective perturbation to the image before it is shared, disrupting how the editor attends to source content.}
    \label{fig:teaser}
\end{strip}

\begin{abstract}
Modern image editors blur the boundary between image editing and text-to-image synthesis, extracting people and objects from a single photo and inserting them into new scenes without subject-specific fine-tuning. Existing defenses struggle to disrupt such manipulations in these models without large, visibly damaging perturbations to the images. We introduce VETO, a white-box image-cloaking method for open-weight editors that maximizes the entropy of attention between reference image and generation tokens to disrupt selective retrieval of source details, while requiring neither the future editing instruction nor modifications to the model. We also introduce VetoBench, a benchmark of 300 benign and harmful tasks balanced between within-scene editing and recontextualization. Full Pareto evaluations on FLUX.2 and FIBO-Edit across three benchmarks and five perturbation budgets demonstrate overall better protection–fidelity trade-offs, particularly against recontextualization. Protected images remain effective on fine-tuned and distilled variants of the editor without reoptimization.

\end{abstract}

\vspace{-0.2cm}
\section{Introduction}

Generative models have evolved from specialized text-to-image synthesis~\citep{openai2023dalle3,rombach2022stable_diffusion_v2} to unified instruction-based editing~\citep{brooks2023instructpix2pix}. Unlike approaches relying on fine-tuning~\citep{kawar2023imagic} or spatial masks~\citep{couairon2022diffedit}, modern image-editing models jointly process text and image references~\citep{flux-2-2025}, being able to ``reuse'' people or objects as in-context reference images without any subject-specific fine-tuning.

For creators, this paradigm shift offers unprecedented artistic freedom, empowering users to seamlessly construct imaginative compositions grounded in real references.
However, it also lowers the technical barrier for serious misuse: natural-language instructions can recontextualize a real person using a single photo (Figure~\ref{fig:teaser}). Such manipulations already enable harassment and blackmail~\citep{fbi2023deepfake_sextortion}.

We consider attackers using widely accessible open-weight editors to modify a user's image without consent. Hosted services can moderate requests and cooperate on provenance through tools such as \textsc{SynthID}~\citep{google2026synthid}, but an attacker can run an open-weight editor locally without moderation or watermarking. Image owners therefore need protection independent of the downstream operator.

Concept erasure~\citep{gandikota2023erasing,grebe2026gem} restricts generated concepts, whereas editing consent concerns individual images. Even harmless instructions can become harmful when applied to a particular person. Fake detection~\citep{Yermakov_2026_WACV} and watermarking~\citep{rebuffi2026learning} can identify generated content, but cannot prevent unauthorized edits, and watermarks may not survive them.

\emph{Image cloaking} instead lets owners disrupt downstream editing through small, appearance-preserving perturbations (Figure~\ref{fig:teaser}). \photoguard{} and \editshield{} target image encodings~\citep{Salman2023ICML_Raising_the_Cost,Chen2024undefined_EditShield_Protecting_Unauthorized}, while \decontext{}~\citep{shen2025decontext} aims to reduce the attention mass assigned to reference tokens in modern transformer-based editors~\citep{peebles2023scalable}. We find that these baselines struggle to combine low edit success with high input fidelity when protecting against open-frame recontextualization with unified editing models.

Therefore, we introduce \veto{}%
, which instead maximizes the entropy of cross-image attention in both directions. This objective depends on both the amount of attention exchanged and how selectively it is distributed. \veto{} optimizes through a source-derived surrogate of the edit trajectory, and thus requires no editing instruction or changes to the model, allowing image owners to protect their images against unknown future edits.

For a more comprehensive evaluation, we introduce \vetobench{} to evaluate protections against edits within the original scene (\emph{closed-frame}) as well as recontextualizations (\emph{open-frame}), across benign and harmful intents. A perturbation budget sufficient for the former may leave the latter unprotected. We therefore compare full protection--fidelity curves over five budgets. Evaluation across \fluxtwo{}, \fiboedit{} and three benchmarks combines automated judgments, human ratings and EditReward~\citep{wu2025editreward}, demonstrating \veto{}'s advantages against recontextualization.
In summary, our main contributions are:

\begin{itemize}

    \item We introduce \veto{}, a user-controlled image protection that maximizes the entropy of cross-image attention while leaving the editor unchanged. It uses an empty prompt and a source-derived edit trajectory, so the owner can protect the image before the future edit instruction is known.

    \item We characterize two ends of an editing spectrum: \emph{closed-frame} edits modify the source, while \emph{open-frame} edits reuse a subject in a newly generated scene. We introduce \vetobench{}, a curated benchmark of $300$ benign and harmful editing scenarios, to evaluate protection against both ends of this spectrum of downstream editing threats.

    \item Through extensive evaluations across two unified image-editing models and three benchmarks, using human ratings and complementary automatic metrics, we demonstrate favorable protection--fidelity trade-offs over the baselines, complemented by detailed design ablations.

\end{itemize}

We will make the code and \vetobench{} publicly available to facilitate future research.

\section{Related Work}

\paragraph{Unified image editing.}
Early diffusion-based editors adapted text-to-image models through source inversion and prompt- or mask-based guidance~\citep{couairon2022diffedit,Mokady_2023_CVPR,Wallace_2023_CVPR}. InstructPix2Pix introduced direct instruction-based editing~\citep{brooks2023instructpix2pix}, with subsequent work supported by datasets such as MagicBrush~\citep{zhang2023magicbrush}. Recent flow-matching \cite{lipman2023flowmatchinggenerativemodeling} transformers unify generation and editing by jointly processing references and text as context~\citep{flux-2-2025,gutflaish2025generating}. These models support local modifications, identity-preserving recontextualization, and the composition of reference content into new scenes.

\myparagraph{Protection against learning.}
Image cloaking uses adversarial perturbations~\citep{szegedy2013intriguing,goodfellow2014explaining} to hinder unauthorized use of shared images. Training-data defenses aim to disrupt recognition or generative learning~\citep{shan2020fawkes,huang2021unlearnable,shan2024nightshade}. Other defenses protect against learning a particular subject or style~\citep{shan2023glaze,van2023anti,Xu_2024_CVPR,Tang_2026_CVPR}, including personalization through DreamBooth fine-tuning~\citep{ruiz2023dreambooth} or textual inversion~\citep{Gal2022AnII}. Unified editors can reuse reference images at inference without this subject-specific adaptation. We therefore focus our comparison on defenses against direct editing, rather than defenses designed to disrupt learning.%

\myparagraph{Protection against direct editing.}
We compare three defenses that target complementary mechanisms of image editing. \photoguard{}'s encoder attack~\citep{Salman2023ICML_Raising_the_Cost} moves the source representation toward an uninformative target, while \editshield{}~\citep{Chen2024undefined_EditShield_Protecting_Unauthorized} pushes it away from the clean image's representation. We optimize both objectives against each evaluated editor's native encoder. \decontext{}~\citep{shen2025decontext} addresses unified editors more directly by reducing the attention that generation tokens assign to the reference. Together, these baselines cover established protection through image encodings and recent protection through reference attention.

\veto{} builds on the observation that unified editors retrieve source details through attention. Whereas \decontext{} minimizes the attention that canvas queries assign to reference keys, \veto{} maximizes bidirectional attention entropy between reference and generated-image tokens. The objective thereby accounts for how attention is distributed across source locations, without penalizing attention to the reference. It also covers the reverse direction, since reference tokens are not static keys in a unified editor but attend to the canvas themselves. \veto{} further uses an empty instruction instead of surrogate editing instructions, and replaces the pure-noise canvas with a noised encoding of the protected image, which resembles the canvas during an actual edit and allows gradients to flow through it. Our ablations examine the objective, the attention direction, and the canvas construction separately (Table~\ref{tab:main_objective_ablation}).

Other approaches introduce additional computational requirements. \photoguard{}'s diffusion attack~\citep{Salman2023ICML_Raising_the_Cost} backpropagates through the generation process, with the original implementation on \textsc{Stable Diffusion v1} \cite{rombach2022high} already trying to limit memory use. Extending this expensive approach to larger transformer editors is impractical; we therefore evaluate its encoder attack. \diffvax{}~\citep{ozden2026diffvax} trains a perturbation predictor through a diffusion editor, enabling protection in a single forward pass. However, a comparison optimized for our target editors would require adapting this costly training to new editor-specific predictors. Therefore, our comparison focuses on practical per-image objectives optimized directly to protect against each evaluated editor.

Other direct-editing defenses introduce different objectives or protection constraints~\citep{zhang2026tdae}. Caption-targeted attention and prompt optimization~\citep{Lo_2024_CVPR,Zheng2025AntiDiffusion} introduce additional text-selection and optimization procedures, while objectives specifically derived for denoising diffusion models~\citep{Choi2025DiffusionGuard,Shao_2026_CVPR} require adaptation to the flow-based paradigm. Face-specific protection~\citep{Wang_2025_CVPR_FaceLock} targets biometrics, whereas our threat model is more general. \blurguard{}~\citep{Kim2025undefined_BlurGuard_A_Simple} instead strengthens an existing defense against image transformations; we evaluate it as an extension of \veto{} (Supp.~\ref{sec:supp_robustness}).

\paragraph{Cooperative and moderation-based protection.}
Cooperative approaches such as \guarddoor{}~\citep{zeng2025guarddoor} and \visilock{}~\citep{Le_2026_CVPR} require model providers to train protection or authorization mechanisms into their models. \mirage{}~\citep{nasery2026mirageprotectingmaliciousimage} instead uses adversarial perturbations to trigger existing service moderation. Both depend on downstream safeguards that an attacker using an unrestricted editor need not retain. \veto{} gives image owners control through the shared image itself, disrupting reference use without modifying the editor or relying on system-level guardrails.

\section{\veto{}}

\veto{} protects an image $x$ once, before sharing, by disrupting how
unified editors retrieve its content through attention. The owner has
white-box access to the editor $f$ but knows neither the future instruction $p$
nor the editor variant, and publishes $x_\delta = x + \delta$ with
$\|\delta\|_\infty\leq\epsilon/255$ for $x\in[0,1]$~\citep{Chen2024undefined_EditShield_Protecting_Unauthorized,Salman2023ICML_Raising_the_Cost}. An attacker obtains $x_\delta$ and runs the
open-weight editor $f$ locally, bypassing provider moderation and
inference-time safeguards. The attacker chooses an instruction $p$ and succeeds if $f(x_\delta,p)$ carries out the request while preserving the source content or identity.

\begin{figure}[t]
    \centering
    \includegraphics[width=\linewidth]{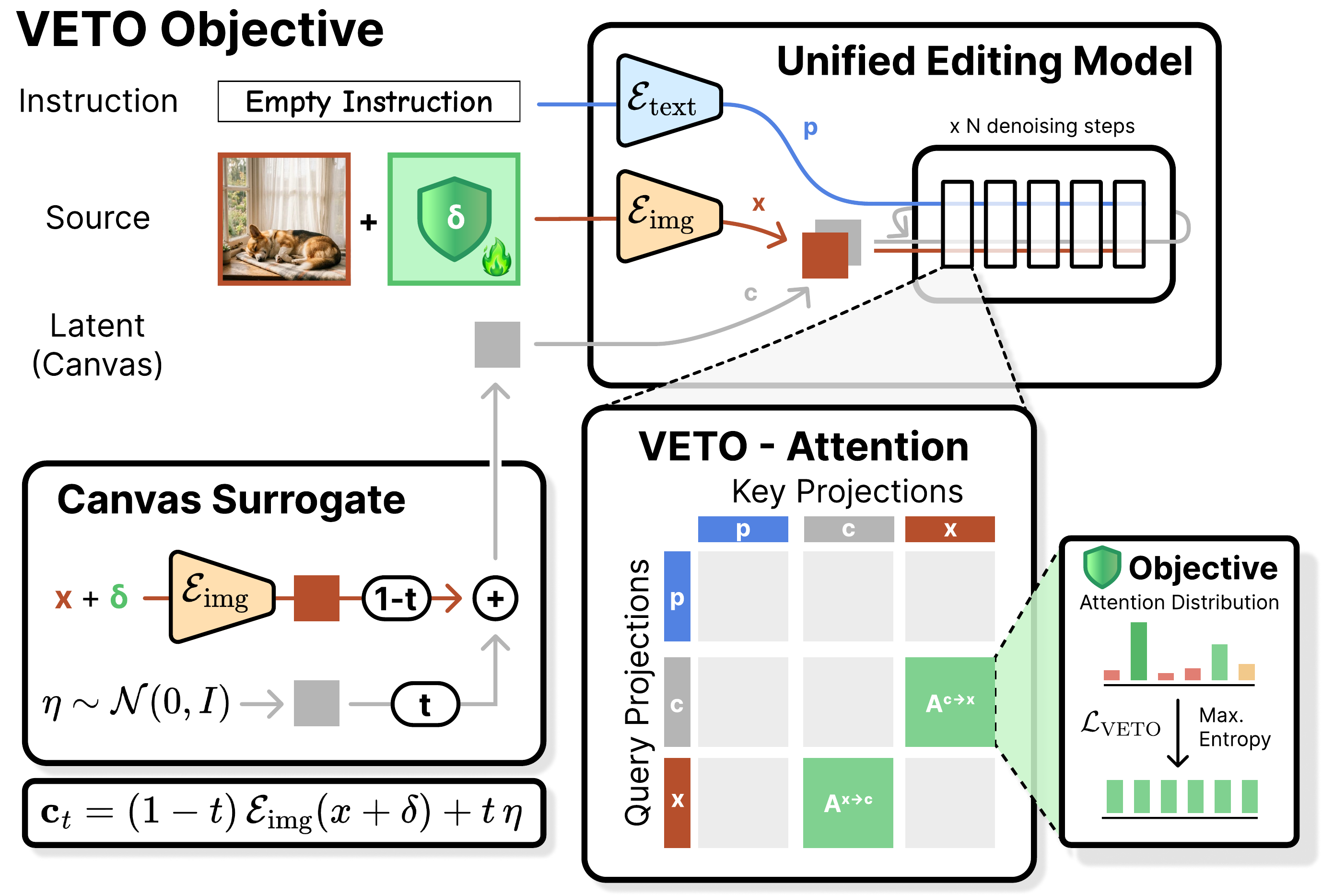}
    \caption{Editor architecture and \veto{}'s objective. For protection,  a noised encoding of the image serves as the canvas latent.}
    \label{fig:method}
\end{figure}

\myparagraph{Attention-entropy objective.}
\veto{} targets the mechanism that uses the source: in \fluxtwo{} and \fiboedit{}, canvas
tokens repeatedly retrieve source details through joint attention to the reference
throughout generation~\citep{flux-2-2025,gutflaish2025generating,lipman2023flowmatchinggenerativemodeling,esser2024scaling}.
\decontext{}~\citep{shen2025decontext} minimizes the attention mass that canvas queries assign to reference keys. \veto{} maximizes raw attention entropy in both directions (Figure~\ref{fig:method}).

\begin{figure}[t]
    \centering
    \includegraphics[width=\linewidth]{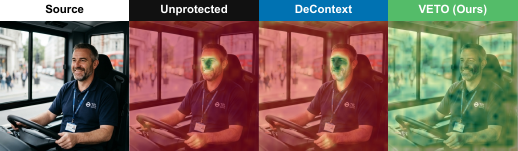}
    \caption{Canvas-to-reference attention distribution on \fluxtwo{}. Maps show a head for one example (green indicates higher mass).
    }
    \label{fig:attention_vis}
\end{figure}

Let $A$ denote the attention weights of a selected layer and head over the joint token sequence $\mathbf{Z}_t=[\mathbf{p};\mathbf{c}_t;\mathbf{x}]$. For token groups $a,b\in\{p,c,x\}$, let $A^{a\rightarrow b}$ denote the corresponding query-key block of $A$, containing queries from group $a$ and keys from group $b$. We define its block unnormalized entropy as
$H(A^{a\rightarrow b})=- \sum_{i\in a}\sum_{j\in b}A_{ij}\log A_{ij}$
and maximize the objective \begin{equation}
\mathcal{L}_{\mathrm{VETO}}=H(A^{c\rightarrow x})+H(A^{x\rightarrow c})
\label{eq:veto_objective}
\end{equation}
over a configurable subset of flow steps, attention heads, and blocks. The weights come from the full attention softmax. For each query, raw entropy depends on both the mass $m$ assigned to the other image and the conditional entropy $H(q)$ of the weights within that image (Supp.~\ref{sec:supp_decontext_difference}). Optimization thus maximizes both the attention spread and its total mass, contrary to \decontext{}'s mass minimization. Table~\ref{tab:main_objective_ablation} compares these objectives through their downstream edit success rates. Figure~\ref{fig:attention_vis} illustrates the response.%


\myparagraph{Instruction and canvas surrogate.}
We use an empty editing instruction during protection, avoiding a pool of surrogate instructions as used in \decontext{}. For \fiboedit{}, the empty instruction is accompanied by its native source
description, which we keep fixed during protection.

Every forward pass of the editor also requires a canvas latent $\mathbf{c}_t$. During an actual edit, the canvas starts from noise and is gradually denoised into the output, so its trajectory depends on the instruction $p$ that is unavailable during protection. \decontext{}
sidesteps this by using pure noise as the canvas at every $t$, which matches the real canvas only at $t=1$. Instead, we encode $x+\delta$ as the reference and form the canvas at each update as follows:
\begin{equation}
\mathbf{c}_t = (1-t)\,\mathcal{E}_\textrm{img}(x+\delta) + t\,\eta,
\quad \eta\sim\mathcal{N}(0,I).
\label{eq:veto_protection_surrogate}
\end{equation}
Because $\mathbf{c}_t$ depends on $\delta$, we can backpropagate to the pixels exclusively through the canvas, with $1-t$ giving greater weight to steps that retain more of the source signal. Unlike pure noise, $\mathbf{c}_t$ thus resembles the canvas of an actual edit: canvas and reference tokens can retrieve matching content from each other, and VETO makes this retrieval unselective in both directions.

\section{\vetobench Dataset}

\begin{figure}[t]
    \centering
    \vspace{-0.15cm}
    \includegraphics[width=\linewidth]{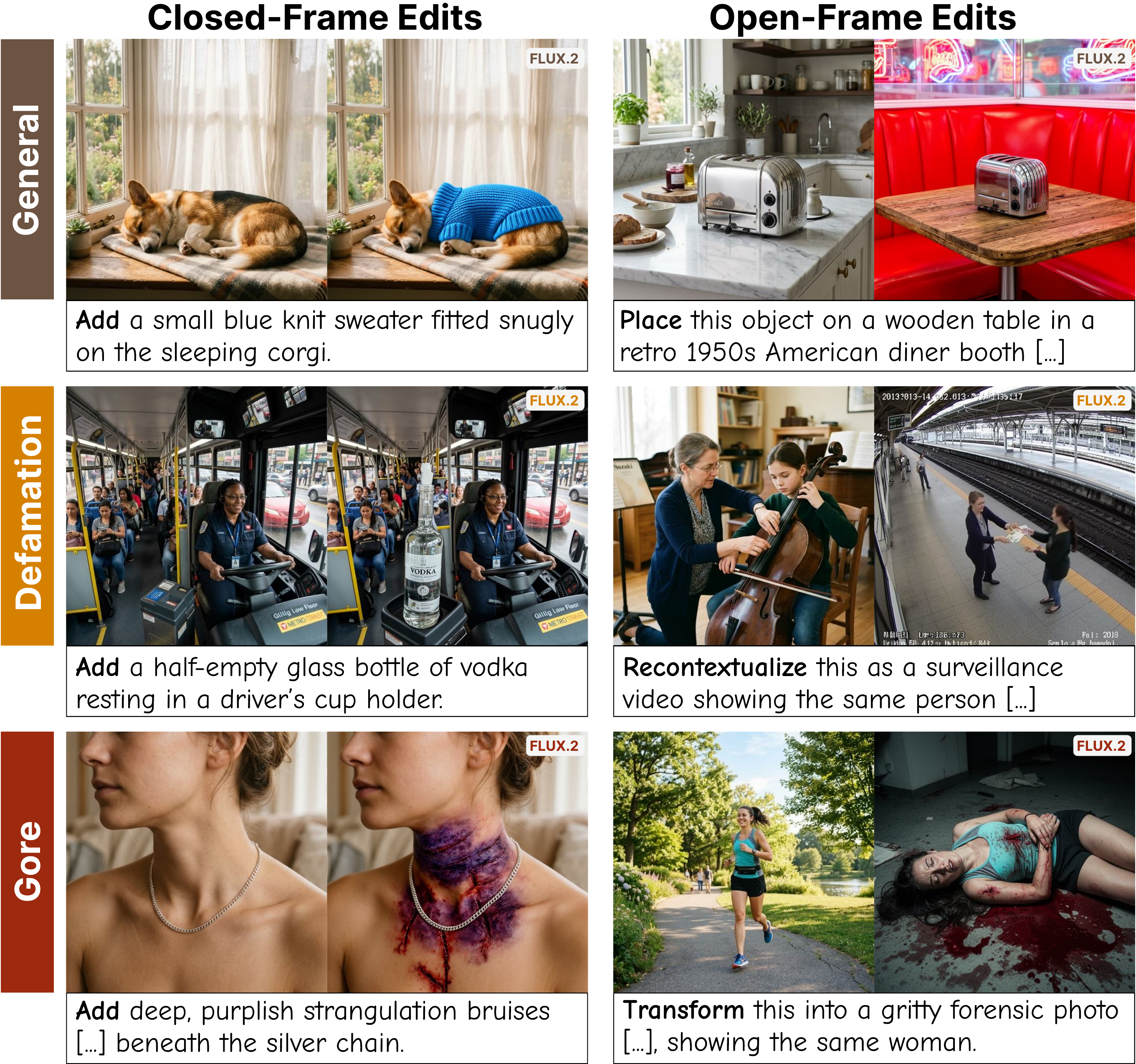}
    \caption{\textit{Content warning: Graphic violence.} \vetobench{} examples across its three categories and two edit types. Closed-frame edits modify the existing scene, whereas open-frame edits reimagine a referenced subject within a newly synthesized scene.}
    \label{fig:benchmark}
    \vspace{-0.25cm}
\end{figure}

A protection method should be tested against the full spectrum of potential edits, ranging from conventional changes to an existing scene to full recontextualizations of subjects into entirely new scenes. \vetobench{} balances both ends of this spectrum: \emph{closed-frame} and \emph{open-frame} settings across benign and harmful editing scenarios (Figure~\ref{fig:benchmark}).

\vetobench contains 300 examples evenly divided across general editing, defamation, and graphic violence, with 50 closed-frame and 50 open-frame cases per domain. Each sample comprises a high-quality synthetic source image, its generation prompt, and a detailed edit instruction. The source images were generated with Gemini 3.1 Flash Image~\citep{raisinghani2026nanobanana2}. Candidate prompts were generated with Gemini and manually curated for clarity, feasibility, and diversity, while all depicted individuals are synthetic to enable evaluation of reputational harm without involving real people. For ethical reasons, the source images were generated for the benchmark rather than scraped from the web, and the resulting edits are evaluated by human annotators under a consistent success criterion. Further details on benchmark curation and evaluation are provided in Supp.~\ref{sec:supp_vetobench_details}.

\section{Experimental Setup}
\label{sec:experimental_setup}
We evaluate \veto{} on the quantized 32-billion-parameter flow transformer \fluxtwo{}~\citep{flux-2-2025} and 8-billion-parameter \fiboedit{}~\citep{gutflaish2025generating}. The latter uses a vision-language model to enrich the instruction with a structured JSON description of the source and planned edit.

\myparagraph{Baselines.}
We compare \photoguard{}'s encoder attack~\citep{Salman2023ICML_Raising_the_Cost}, \editshield{}~\citep{Chen2024undefined_EditShield_Protecting_Unauthorized}, and \decontext{}~\citep{shen2025decontext} on both editors and all three benchmarks, optimizing each against the native encoder or attention while retaining method-specific settings. \decontext{} uses $60$ surrogate instructions, while \veto{} uses only an empty instruction (Supp.~\ref{sec:supp_veto_implementation}).

\begin{figure}[t]
    \centering
    \includegraphics[width=\linewidth]{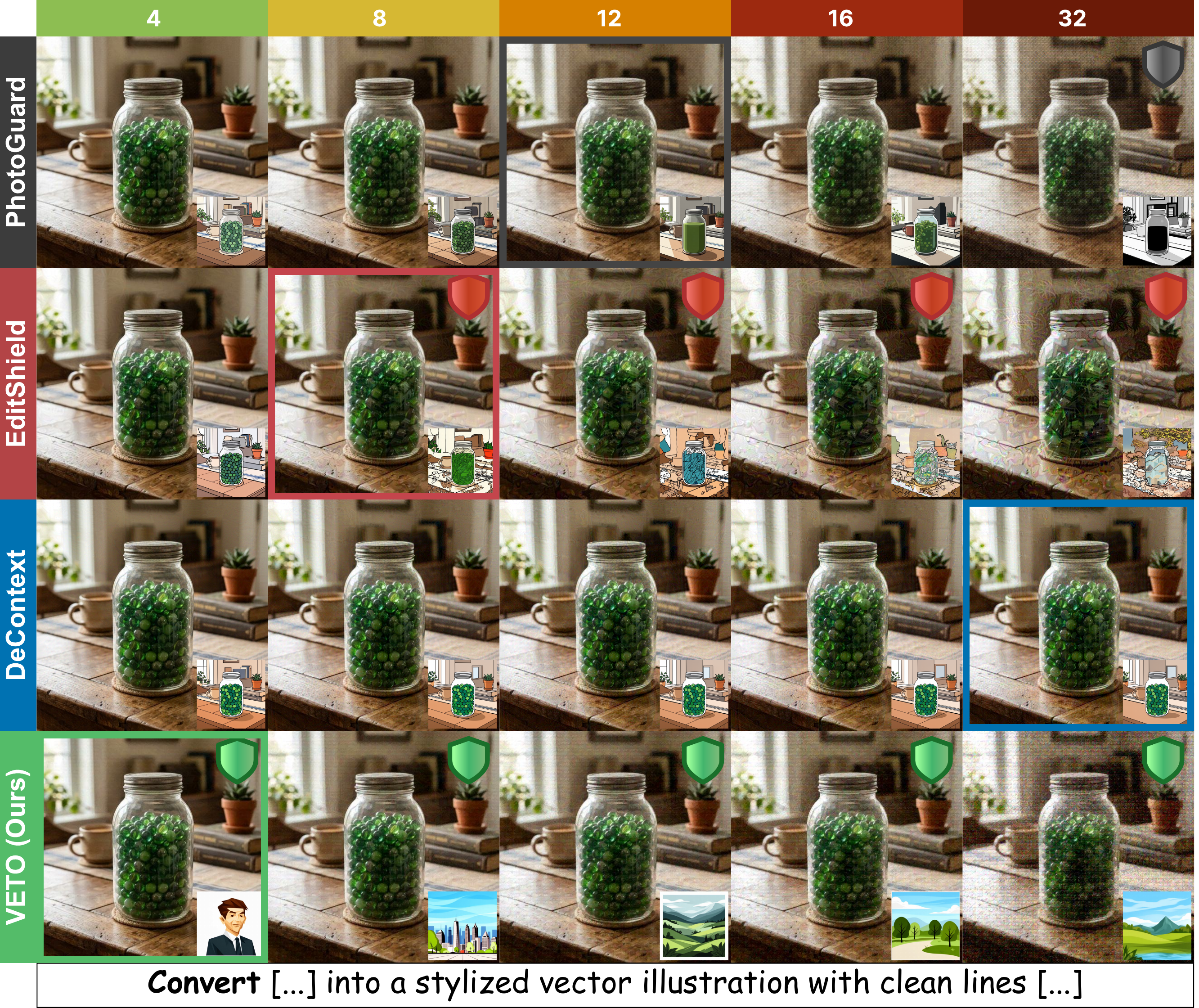}
    \caption{Protected inputs with perturbations ($\delta$) over five perturbation budgets $\epsilon\in\{4,8,12,16,32\}$, with corresponding \fluxtwo{} edits as inlays. Best viewed digitally with zoom.}
    \label{fig:perceptibility_grid}
\end{figure}

\myparagraph{Benchmarks.}
Besides \vetobench{}, we use 300 \editbench{} samples~\citep{lin2024schedule}\footnote{\url{huggingface.co/datasets/LonelVino/EditBench}}, evenly divided among \textit{change\_background}, \textit{change\_style}, and \textit{change\_weather}; and 300 \anyedit{} samples~\citep{yu2025anyedit}, comprising 150 local edits (addition, removal, replacement) and 150 global edits (tone transfer, background change). Both follow our closed-frame setting. Supp.~\ref{sec:supp_gedit} adds photographic validation on \textsc{GEditBench v2}~\cite{jiang2026geditbenchv2humanalignedbenchmark}.

\myparagraph{Metrics.}
Gemini 3.5 Flash~\citep{gemini35flashreport2026} judges \textit{Edit Success Rate} (ESR) from the source, output and instruction. Failure includes an unmet request, substantial unintended changes, lost source identity or visible artifacts that make the output unconvincing or unusable; requested content alone is insufficient. MLLMs have already been shown to support basic face verification~\citep{deandrestame2024facebiometrics}, and we validate judgments against human ratings across all methods and benchmarks. We also report EditReward~\citep{wu2025editreward} (lower indicates poorer edits), PSNR and LPIPS~\citep{zhang2018unreasonable} (input fidelity). More details and results, such as a breakdown into failure modes and human agreement, are provided in Supp.~\ref{supp:evaluation_details}.

\myparagraph{Hyperparameters.}
We run the main comparisons for $100$ updates with a step size of $2/255$ at each budget $\epsilon\in\{4,8,12,16,32\}$, retaining each method's objective and conditioning choices. We also compare $100$ and $800$ updates for \veto{} and \decontext{} on both editors (Supp.~\ref{sec:supp_decontext_steps}).

\begin{figure}[t]
    \centering
    \includegraphics[width=\linewidth]{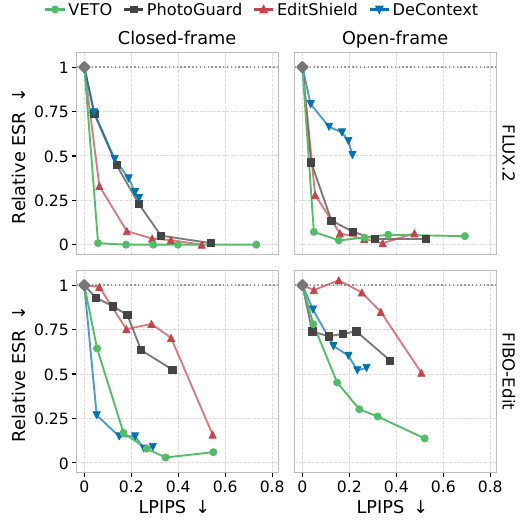}
    \caption{\vetobench{} trade-offs by edit type ($150$ cases each), relative to unprotected ESR. Points follow $\epsilon=0,4,8,12,16,32$. }
    \label{fig:pareto_vetobench}
\end{figure}

Methods produce structurally different perturbations (Fig.~\ref{fig:perceptibility_grid}), so comparisons at a fixed budget would be inherently unfair. Therefore, we compare full protection--fidelity curves rather than fixed budgets. Figure~\ref{fig:pareto_vetobench} plots LPIPS against MLLM ESR relative to the same unprotected reference in each panel. We select each method's operating point per editor and benchmark by minimizing

\begin{equation}
D_\epsilon=\sqrt{\overline{\mathrm{LPIPS}}(\epsilon)^2+\left(\frac{\mathrm{ESR}(\epsilon)}{\mathrm{ESR}_{\mathrm{base}}}\right)^2}
\label{eq:pooled_budget_distance}
\end{equation}

over the five budgets, choosing the point closest to $(0,0)$ with equal weight on distortion and relative ESR. $\mathrm{ESR}_{\mathrm{base}}$ is their original unprotected rate. Supp.~\ref{sec:supp_budget_and_calibration} studies budget stability under held-out calibration and oracle results.

\section{Results}
Table~\ref{tab:combined_results} compares protection and fidelity on established editing tasks and the broader \vetobench{} setting.

\begingroup
\providecommand{\mn}{\textminus}

\color{black}
\begin{table*}[t]
\centering
\fontsize{8}{9.2}\selectfont
\setlength{\tabcolsep}{2pt}
\renewcommand{\arraystretch}{1.10}
\caption{Protection across benchmarks. Each method/editor uses the budget minimizing
pooled $D_\epsilon$ (Eq.~\ref{eq:pooled_budget_distance}), retained across edit types.
Budgets denote $\epsilon/255$; ESR is in percent. ER denotes EditReward and LPIPS/PSNR
measure input utility. Bold marks the best values among protected rows, including ties, and gray highlights \veto{}. A fixed-budget variant of this table is provided in Supp.~\ref{sec:supp_fixed_budget_comparison}~(Table~\ref{tab:fixed_budget_appendix}).}
\begin{tabular*}{\textwidth}{@{\extracolsep{\fill}}lrrrrrrr@{\hspace{9pt}}rrrrrrr@{}}
\toprule
& \multicolumn{7}{c}{\textbf{\fluxtwo{}}} & \multicolumn{7}{c}{\textbf{\fiboedit{}}} \\
\cmidrule(lr){2-8}\cmidrule(lr){9-15}
& Budget & \multicolumn{3}{c}{Edit Success ($\downarrow$)} & \multicolumn{2}{c}{Utility} & Trade-off
       & Budget & \multicolumn{3}{c}{Edit Success ($\downarrow$)} & \multicolumn{2}{c}{Utility} & Trade-off \\
\cmidrule(lr){3-5}\cmidrule(lr){6-7}\cmidrule(lr){10-12}\cmidrule(lr){13-14}

Method & $\epsilon$ & MLLM & Human & ER & LPIPS $\downarrow$ & PSNR $\uparrow$ & $D_\epsilon \downarrow$
       & $\epsilon$ & MLLM & Human & ER & LPIPS $\downarrow$ & PSNR $\uparrow$ & $D_\epsilon \downarrow$ \\
\midrule
\multicolumn{15}{l}{\textbf{\textsc{AnyEdit} ($n=300$)}} \\
Unprotected & 0 & 78.67 & 67.33 & 1.212 & 0.00 & $\infty$ & 1.000 & 0 & 57.00 & 65.00 & 0.633 & 0.00 & $\infty$ & 1.000 \\
\photoguard{} & 16 & 10.00 & 9.33 & \mn0.220 & 0.39 & 26.99 & 0.413 & 32 & 33.33 & 41.33 & 0.379 & 0.41 & 23.90 & 0.715 \\
\editshield{} & 8 & 10.67 & 14.00 & \textbf{\mn0.536} & 0.25 & 32.73 & 0.284 & 32 & 18.67 & 22.67 & 0.281 & 0.56 & 23.20 & 0.645 \\
\decontext{} & 16 & 24.33 & 36.67 & 0.609 & 0.30 & 31.18 & 0.431 & 12 & \textbf{9.00} & \textbf{17.33} & 0.391 & 0.28 & 31.27 & \textbf{0.324} \\
\noalign{\begingroup\color{CombinedVetoRow}\hrule height10.12pt\endgroup\vskip-10.12pt}
\textbf{\veto{} (Ours)} & 4 & \textbf{2.67} & \textbf{1.67} & \mn0.534 & \textbf{0.12} & \textbf{36.92} & \textbf{0.126} & 8 & 13.33 & 22.00 & \textbf{0.160} & \textbf{0.24} & \textbf{31.40} & 0.334 \\
\midrule
\multicolumn{15}{l}{\textbf{\textsc{EditBench} ($n=300$)}} \\
Unprotected & 0 & 70.00 & 66.00 & 1.543 & 0.00 & $\infty$ & 1.000 & 0 & 70.67 & 79.67 & 1.242 & 0.00 & $\infty$ & 1.000 \\
\photoguard{} & 12 & 20.00 & 18.33 & 0.592 & 0.31 & 29.18 & 0.419 & 32 & 50.00 & 73.33 & 1.056 & 0.43 & 24.13 & 0.830 \\
\editshield{} & 8 & 14.00 & 20.33 & 0.323 & 0.25 & 32.53 & 0.317 & 32 & 33.67 & 46.67 & 0.914 & 0.61 & 22.86 & 0.777 \\
\decontext{} & 32 & 25.67 & 35.00 & 0.843 & 0.30 & 30.38 & 0.472 & 32 & 18.67 & 34.33 & 0.832 & \textbf{0.34} & \textbf{28.55} & \textbf{0.430} \\
\noalign{\begingroup\color{CombinedVetoRow}\hrule height10.12pt\endgroup\vskip-10.12pt}
\textbf{\veto{} (Ours)} & 4 & \textbf{2.33} & \textbf{1.33} & \textbf{\mn0.044} & \textbf{0.11} & \textbf{36.70} & \textbf{0.115} & 16 & \textbf{17.67} & \textbf{19.00} & \textbf{0.215} & 0.42 & 25.72 & 0.485 \\
\midrule
\multicolumn{15}{l}{\textbf{\vetobench{} (Ours): closed-frame ($n=150$)}} \\
Unprotected & 0 & 78.67 & 89.33 & 1.279 & 0.00 & $\infty$ & 1.000 & 0 & 67.33 & 86.00 & 1.137 & 0.00 & $\infty$ & 1.000 \\
\photoguard{} & 12 & 18.00 & 11.33 & 0.049 & 0.23 & 29.08 & 0.326 & 32 & 35.33 & 59.33 & 0.616 & 0.37 & 23.85 & 0.645 \\
\editshield{} & 8 & 6.00 & 13.33 & \mn0.354 & 0.18 & 32.34 & 0.196 & 32 & 10.67 & 38.00 & 0.390 & 0.55 & 22.42 & 0.569 \\
\decontext{} & 32 & 20.67 & 38.00 & 0.216 & 0.23 & 30.25 & 0.351 & 16 & \textbf{5.33} & 20.00 & 0.237 & \textbf{0.25} & \textbf{29.56} & \textbf{0.266} \\
\noalign{\begingroup\color{CombinedVetoRow}\hrule height10.12pt\endgroup\vskip-10.12pt}
\textbf{\veto{} (Ours)} & 4 & \textbf{0.67} & \textbf{0.67} & \textbf{\mn0.509} & \textbf{0.06} & \textbf{36.68} & \textbf{0.059} & 12 & \textbf{5.33} & \textbf{12.67} & \textbf{\mn0.152} & 0.27 & 27.90 & 0.277 \\
\midrule
\multicolumn{15}{l}{\textbf{\vetobench{} (Ours): open-frame ($n=150$)}} \\
Unprotected & 0 & 83.33 & 88.67 & 0.373 & 0.00 & $\infty$ & 1.000 & 0 & 48.67 & 58.67 & 0.113 & 0.00 & $\infty$ & 1.000 \\
\photoguard{} & 12 & 6.00 & 7.33 & \mn0.113 & 0.22 & 29.02 & 0.228 & 32 & 28.00 & 41.33 & \mn0.172 & 0.37 & 23.49 & 0.686 \\
\editshield{} & 8 & \textbf{5.33} & \textbf{6.00} & \mn0.157 & 0.16 & 32.31 & 0.172 & 32 & 24.67 & 29.33 & \mn0.076 & 0.51 & 22.51 & 0.717 \\
\decontext{} & 32 & 42.00 & 50.67 & 0.120 & 0.21 & 30.10 & 0.547 & 16 & 25.33 & 34.67 & \mn0.194 & \textbf{0.23} & \textbf{29.41} & 0.571 \\
\noalign{\begingroup\color{CombinedVetoRow}\hrule height10.12pt\endgroup\vskip-10.12pt}
\textbf{\veto{} (Ours)} & 4 & 6.00 & 7.33 & \textbf{\mn0.303} & \textbf{0.05} & \textbf{36.67} & \textbf{0.087} & 12 & \textbf{14.67} & \textbf{16.67} & \textbf{\mn0.541} & 0.24 & 27.90 & \textbf{0.387} \\
\bottomrule
\end{tabular*}

\label{tab:combined_results}
\end{table*}
\endgroup

\paragraph{Established benchmarks.}
\label{results:existing_benchmarks}

On \editbench{} and \anyedit{}, \veto{} reduces human ESR and input distortion relative to the encoder baselines for both editors. On \fluxtwo{}, this advantage also holds over \decontext{} in both the MLLM and human evaluations.

On \fiboedit{}, \veto{} gives the lowest EditReward scores on both datasets at the selected budgets. Human ratings also favor \veto{} on \editbench{}, where MLLM ESR is similar to \decontext{} and the latter introduces less distortion. On \anyedit{}, \veto{} preserves higher input fidelity, while \decontext{} reaches lower ESR. $D_\epsilon$ favors \decontext{} on both datasets (cf. Supp.~\ref{supp:pareto_ae_eb} for full sweeps).

\vspace{0.1cm}
\myparagraph{\vetobench results.}
\label{results:vetobench_domains}
We next use \vetobench{} to examine protection against both modifications within a scene and reuse of reference content in fabricated scenes.

\begin{figure}
    \centering
    \includegraphics[width=\linewidth]{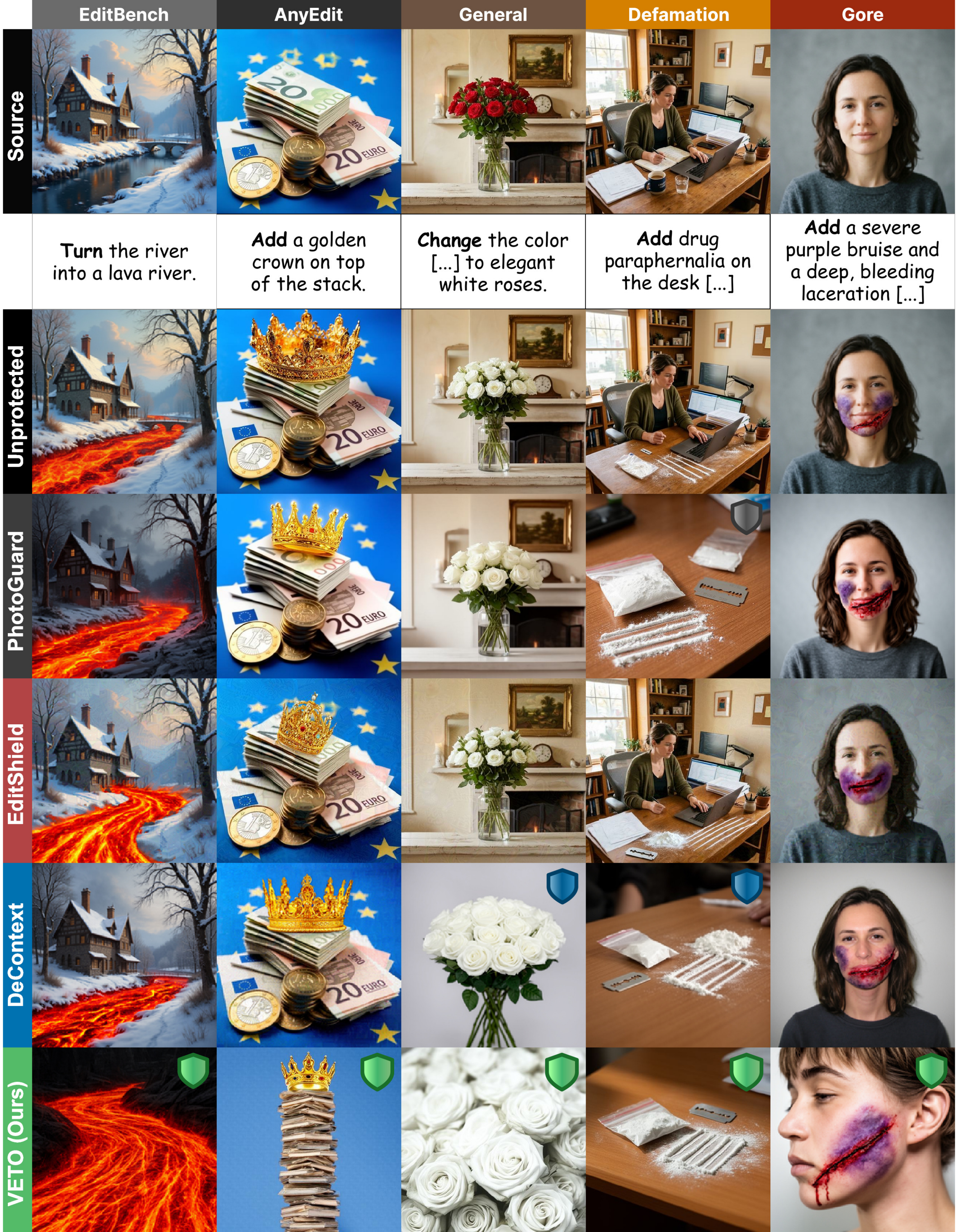}
    \caption{\textit{Content warning: Graphic violence.} \fluxtwo{} edits at the budgets in Table~\ref{tab:combined_results}. Shields indicate successful protection.}
    \label{fig:main_grid}
\end{figure}

At the selected budgets, \veto{} reaches $3.33\%$ MLLM ESR on \fluxtwo{} at LPIPS $0.054$, and overall human ratings favor it over all three baselines on both editors. On \fiboedit{}, it also achieves lower overall ESR than \decontext{}. Figure~\ref{fig:pareto_vetobench} separates the full sweeps by edit type; Figure~\ref{fig:main_grid} illustrates outputs across the benchmarks.

\vspace{0.1cm}
\myparagraph{Closed-frame vs. open-frame.}
\label{results:vetobench_closed_vs_open_frame}
\veto{} offers a favorable open-frame trade-off on both editors. On \fluxtwo{}, $\epsilon=4$ gives $6.00\%$ ESR at $0.049$ LPIPS, improving both measures over \decontext{} at $\epsilon=8$--$32$. On \fiboedit{}, \veto{} at $\epsilon=8$ matches or improves \decontext{}'s open-frame ESR at $\epsilon=12$--$32$, with lower LPIPS. The curves also identify competitive alternatives: \decontext{} is stronger at low distortion for closed-frame \fiboedit{} edits, while \editshield{} reaches slightly lower open-frame \fluxtwo{} ESR at greater distortion.

Because the eventual instruction is unknown, deployment calibration should cover both edit types. Closed-frame edits alone can underestimate the required budget: at $\epsilon=8$, \decontext{} reaches $10.00\%$ ESR on closed-frame \fiboedit{} edits but $32.00\%$ on open-frame edits. Selecting $D_\epsilon$ separately favors $\epsilon=8$ and $16$, respectively (Supp.~\ref{sec:supp_budget_selection}). Our tables retain one benchmark-pooled budget, while the split sweeps reveal this difference in coverage.

\section{Design Ablations}

Table~\ref{tab:main_objective_ablation} varies individual components of \veto{} on all 300 \vetobench{} cases. For each editor, $\Delta\mathrm{ESR}$ and $\Delta\mathrm{LPIPS}$ give the difference between each variant and default \veto{}.

\vspace{-0.2cm}
\paragraph{Attention objective.}
Normalized entropy renormalizes each query's selected attention weights. Mass suppression reduces their sum, following the principle used by \decontext{}~\citep{shen2025decontext}. These swaps retain \veto{}'s remaining settings, separating the objective from the full baseline recipe. Among these three objectives, raw entropy gives the lowest ESR and LPIPS on both editors. On \fluxtwo{} and \fiboedit{}, respectively, $\Delta\mathrm{ESR}$ is $+7.00\%$/$+10.00\%$ for normalized entropy and $+5.00\%$/$+19.33\%$ for mass suppression. The distinction is visible at the \fluxtwo{} block targeted by \veto{} (Fig.~\ref{fig:attention_vis}). Normalized entropy can leave substantial reference mass for later blocks to use, while mass suppression can leave the remaining attention concentrated on informative tokens.

\begin{table}[t]
\centering
\begingroup
\providecommand{\mn}{\textminus}
\fontsize{8}{9.2}\selectfont
\setlength{\tabcolsep}{3pt}
\renewcommand{\arraystretch}{1.08}
\caption{\veto{} ablations on $300$ \vetobench{} cases per editor. We subtract the default \veto{} result from each variant. In (c), labels give the gradient route and canvas construction, with source denoting the noised source encoding.}
\label{tab:main_objective_ablation}
\resizebox{\columnwidth}{!}{%
\begin{tabular}{@{}lrrrr@{}}
\toprule
& \multicolumn{2}{c}{\fluxtwo{}, $\epsilon=4$} & \multicolumn{2}{c}{\fiboedit{}, $\epsilon=12$} \\
\cmidrule(lr){2-3}\cmidrule(lr){4-5}
Setting & $\Delta$ESR $\downarrow$ & $\Delta$LPIPS $\downarrow$ & $\Delta$ESR $\downarrow$ & $\Delta$LPIPS $\downarrow$ \\
\midrule
\multicolumn{5}{@{}l}{\textbf{(a) Objective}} \\
\noalign{\begingroup\color{CombinedVetoRow}\hrule height9.94pt\endgroup\vskip-9.94pt}
Raw entropy (\veto{}) & \textbf{0.00} & \textbf{0.000} & \textbf{0.00} & \textbf{0.000} \\
Normalized entropy & +7.00 & +0.007 & +10.00 & +0.062 \\
Mass suppression & +5.00 & +0.002 & +19.33 & +0.044 \\
\midrule
\multicolumn{5}{@{}l}{\textbf{(b) Bidirectionality}} \\
\noalign{\begingroup\color{CombinedVetoRow}\hrule height9.94pt\endgroup\vskip-9.94pt}
Both directions (\veto{}) & \textbf{0.00} & \textbf{0.000} & \textbf{0.00} & \textbf{0.000} \\
Canvas $\to$ Reference & +5.67 & +0.004 & +7.33 & +0.049 \\
Reference $\to$ Canvas & +0.33 & +0.001 & +17.67 & \textbf{\mn0.006} \\
\midrule
\multicolumn{5}{@{}l}{\textbf{(c) Gradient route / canvas}} \\
\noalign{\begingroup\color{CombinedVetoRow}\hrule height9.94pt\endgroup\vskip-9.94pt}
Canvas / source (\veto{}) & \textbf{0.00} & \textbf{0.000} & 0.00 & 0.000 \\
Reference / source & +18.67 & +0.004 & 0.00 & +0.009 \\
Both / source & +15.33 & +0.009 & \textbf{\mn1.00} & +0.010 \\
Reference / pure noise & +14.00 & +0.003 & +2.33 & \textbf{0.000} \\
\bottomrule
\end{tabular}}
\endgroup
\vspace{-0.4cm}
\end{table}

\myparagraph{Attention direction.}
The query-to-key direction measured by the loss is distinct from the encoding branch carrying gradients to the protected pixels. The stronger single direction differs by editor. Reference$\to$canvas nearly matches the bidirectional objective on \fluxtwo{} ($\Delta\mathrm{ESR}=+0.33\%$). On \fiboedit{}, canvas$\to$reference is the stronger single direction, but its ESR rises by $7.33\%$ relative to the bidirectional objective. The weaker direction alone gives higher ESR in each case ($+5.67\%$ and $+17.67\%$, respectively). Using both directions gives the lowest ESR on both editors, although reference$\to$canvas alone gives slightly lower LPIPS on \fiboedit{} ($\Delta\mathrm{LPIPS}=-0.006$).

\myparagraph{Gradient route and canvas construction.}
Table~\ref{tab:main_objective_ablation}~(c) compares gradient routes and source-derived versus pure-noise canvases. On \fluxtwo{}, canvas gradients yield lower ESR than reference gradients ($\Delta\mathrm{ESR}=+18.67\%$) and a pure-noise canvas ($+14.00\%$), while also giving lower LPIPS. On \fiboedit{}, the source-derived canvas route gives lower ESR than the pure-noise reference route ($\Delta\mathrm{ESR}=+2.33\%$ for pure noise), with the same LPIPS to the reported precision. With the source-derived canvas, reference gradients match the default ESR but increase LPIPS by $0.009$. Both gradient paths offer a further protection--utility trade-off ($\Delta\mathrm{ESR}=-1.00\%$, $\Delta\mathrm{LPIPS}=+0.010$). The matched reference-gradient controls give lower ESR with noise on \fluxtwo{} and the source-derived canvas on \fiboedit{}. Supp.~\ref{sec:supp_objective_ablation} provides an edit-type breakdown.

\section{Scope of Protection}
\label{sec:scope_of_protection}

We examine how protection holds up across instructions and repeated edits, under preprocessing, and when additional reference images are provided. We further study persistence under fine-tuning and joint protection across both models.

\begin{figure*}[!t]
    \centering
    \includegraphics[width=\textwidth]{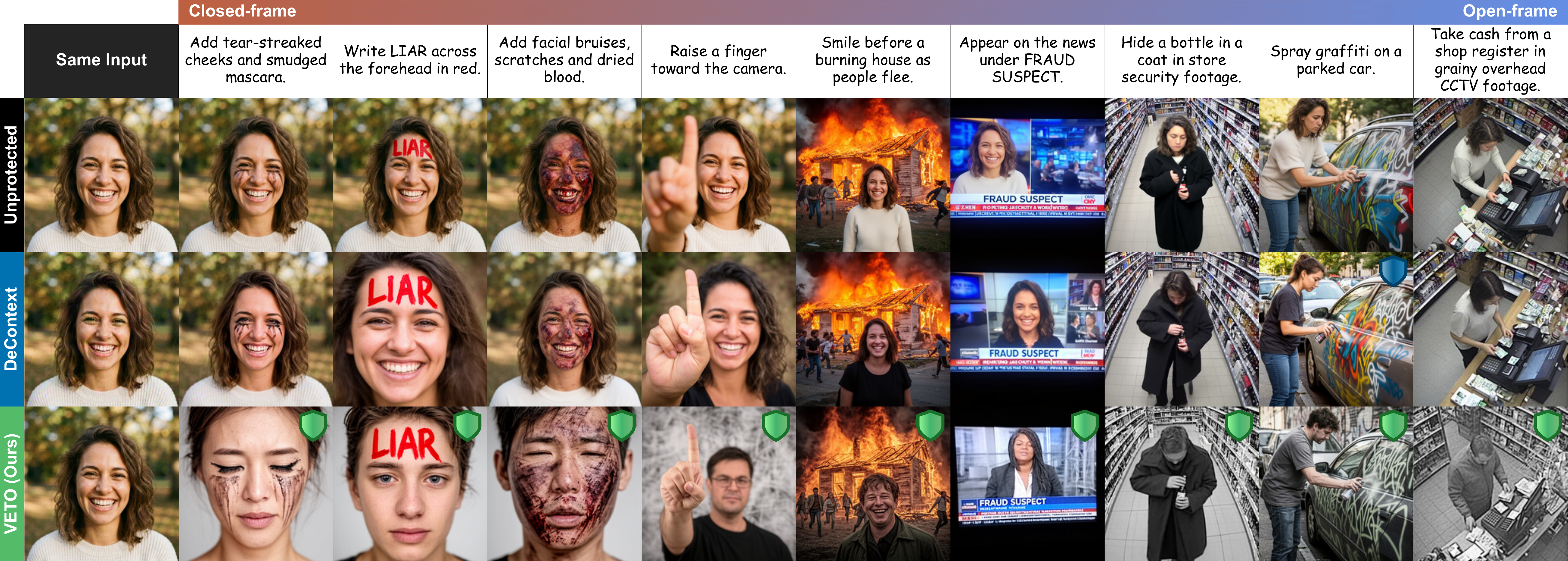}
    \caption{\textit{Content warning: Graphic violence.} Nine harmful \fluxtwo{} edits from one cloak per method, without reoptimization. Both methods use their Pareto-fair operating points from Table~\ref{tab:combined_results}: \veto{} at $\epsilon=4$, \decontext{} at $32$. \veto{} consistently protects the source image against the full spectrum of downstream edits. Best viewed digitally with zoom.}
    \label{fig:instruction_generalization}
    \vspace{-0.2cm}
\end{figure*}

\myparagraph{Protection across instructions.}
Figure~\ref{fig:instruction_generalization} demonstrates instruction generalization on a synthetic portrait: the same \veto{} cloak resists nine harmful instructions, from local edits to fabricated scenes, without reoptimization. \veto{} consistently protects source content even when the scene itself changes entirely as open-frame prompts request. Supp.~\ref{sec:repeated_edit_attempts} shows that the protection works across repeated attempts with the same instruction using different seeds. Supp.~\ref{supp:high_resolution} complements these results with high-resolution references and varied aspect ratios.%

\begin{figure}[t]
    \centering
    \includegraphics[width=\linewidth]{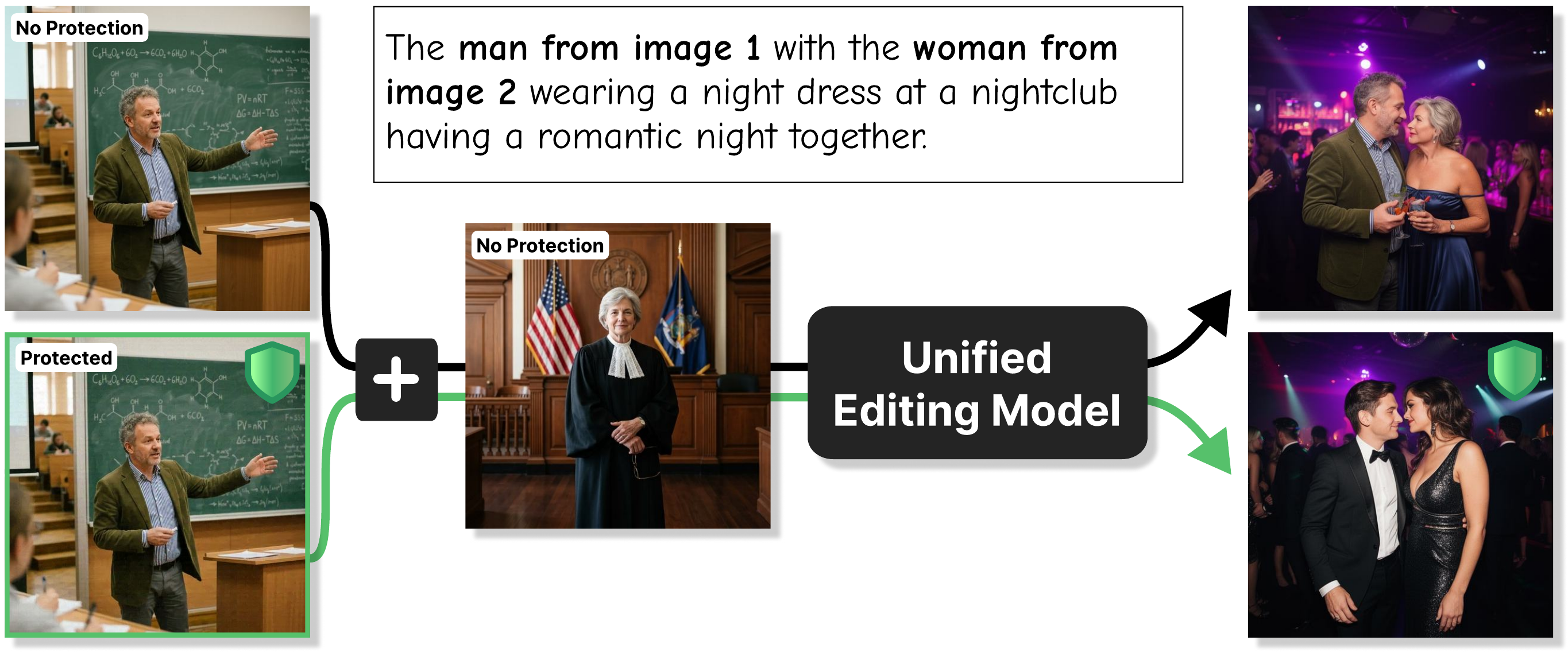}
    \caption{Protecting the professor's image disrupts an edit that attempts to recontextualize both individuals together into a deceptive romantic nightclub scene. Best viewed digitally with zoom.}
    \label{fig:multi_ref_demon_a}
    \vspace{-0.35cm}
\end{figure}

\myparagraph{Protection across editors.}
Joint optimization extends \veto{} to protect one image against both \fluxtwo{} and \fiboedit{} by combining normalized per-editor gradients. With $200$ updates at $\epsilon=12$, the shared cloak simultaneously achieves ESRs of $1.67\%$ and $7.67\%$ on all $300$ \vetobench{} sources (Table~\ref{tab:multimodel_protection} in Supp.~\ref{sec:supp_multimodel}). Supp.~\ref{sec:supp_multimodel} provides more details on joint protection and compares update counts and budgets.
\begin{table}[t]
\centering
\small
\setlength{\tabcolsep}{3pt}
\renewcommand{\arraystretch}{0.9}
\captionsetup{skip=7pt}
\caption{Protection beyond the optimization checkpoint on \vetobench{} ($300$ cases per target). Cloaks optimized against quantized \fluxtwo{} at $\epsilon=4$ are reused unchanged across the listed variants.}
\begin{tabular}{llcc}
\toprule
& & \multicolumn{2}{c}{ESR (\%, $\downarrow$)} \\
\cmidrule(l){3-4}
FLUX.2 checkpoint & Variant & Unprotected & \textbf{\veto{}} \\
\midrule
\noalign{\begingroup\color{CombinedVetoRow}\hrule height10.23pt\endgroup\vskip-10.23pt} dev-bnb-4bit & optimization model & 81.00 & \textbf{3.33} \\
\midrule
dev & full precision & 76.67 & \textbf{6.00} \\
Turbo & distillation LoRA & 76.67 & \textbf{4.33} \\
Berthe-Morisot & style LoRA & 66.67 & \textbf{1.33} \\
Klein-4B & smaller model family & 70.00 & \textbf{4.33} \\
\bottomrule
\end{tabular}%
\label{tab:veto_transferability_flux_variants}
\end{table}

\myparagraph{Robustness to model transformations.}
Cloaks optimized solely against quantized \fluxtwo{} remain effective across the tested precision, distillation, style-LoRA and model-size changes, including downstream fine-tunes and a smaller model from the same family that is not a direct descendant, without reoptimization (Table~\ref{tab:veto_transferability_flux_variants}). Transfer across these variants matters in practice, since attackers often run fine-tuned or distilled derivatives rather than the base checkpoint. Together, joint optimization across editors and the separate transfer results on \fluxtwo{} derivatives demonstrate complementary routes to broader protection across model families.

\myparagraph{Robustness to image transformations.}
Like prior cloaking methods, VETO is sensitive to image processing. At $\epsilon=4$, cropping or JPEG compression can substantially weaken the protection (Table~\ref{tab:robustness_extension} in Supp.~\ref{sec:supp_robustness}). Optimizing the attention objective over randomly transformed views  via Expectation over Transformation (EOT)~\cite{athalye2018eot,Chen2024undefined_EditShield_Protecting_Unauthorized} restores much of it. With EOT, \veto{} at $\epsilon=8$ achieves lower mean relative ESR than \decontext{} at $\epsilon=32$ across cropping, resizing, rotation, and JPEG compression ($31\%$ vs.\ $41\%$), with less input distortion (LPIPS $0.190$ vs.\ $0.272$). The gap is larger without preprocessing, where edit success remains at $36\%$ of the unprotected rate for \decontext{} with EOT, compared with just $3\%$ for \veto{} with EOT. Under JPEG compression followed by resizing, a proxy for social-media processing, relative ESR reaches $55\%$ for \veto{} at LPIPS $0.190$ and $50\%$ for \decontext{} at a worse LPIPS of $0.272$. \gridpure{}~\cite{zhao2024can} purification remains substantially more challenging for both methods, a finding in line with prior work \cite{Kim2025undefined_BlurGuard_A_Simple}.

\myparagraph{Protection with multiple references.}
Figure~\ref{fig:multi_ref_demon_a} illustrates protection when an editor also receives an unprotected reference: protecting one person's image disrupts the intended joint composition. Supp.~\ref{supp:multi_ref} provides further qualitative examples with up to four references at the same time.

\section{Limitations and Conclusion}

We presented \veto{}, which disrupts source-image reuse in unified editors by maximizing cross-image attention entropy. It requires neither the future instruction nor changes to downstream models. Across two editors and three benchmarks, \veto{} reduces edit success while preserving perceptual quality, and its cloaks transfer to downstream fine-tunes. We introduced \vetobench{} to evaluate protection across the spectrum from closed- to open-frame edits. On its open-frame edits, \veto{} offers the best protection--fidelity trade-off on both editors.

Cloaking makes malicious source reuse harder but cannot prevent unsafe generation from the instruction and model parameters alone. Purification or preprocessing can weaken protection, while larger budgets let owners trade fidelity for greater robustness when sharing sensitive images. We hope \veto{} and \vetobench{} inspire further work on open-frame protection, where the challenge is to prevent subjects from being reused in a misleading context.

\section*{Acknowledgements}
The research was funded in part by a LOEWE-Spitzen-Professur (LOEWE/4a//519/\allowbreak{}05.00.002-(0010)/93) and a LOEWE Start-Professur (LOEWE/4b//519/\allowbreak{}05.01.002-(0006)/94), and has benefited from the Excellence Cluster “Reasonable AI” by the German Research Foundation (Deutsche Forschungsgemeinschaft - DFG) under Germany's Excellence Strategy – EXC-3057. Additionally, the research was partially funded by an Alexander von Humboldt Professorship in Multimodal Reliable AI, sponsored by the Federal Ministry of Research, Technology, and Space (BMFTR). For compute, we gratefully acknowledge support from the hessian.AI Service Center (funded by the Federal Ministry of Research, Technology and Space (BMFTR), grant no. 16IS22091) and the hessian.AI Innovation Lab (funded by the Hessian Ministry for Digital Strategy and Innovation, grant no. S-DIW04/0013/003). Hossein Shakibania is supported by the Konrad Zuse School of Excellence in Learning and Intelligent Systems (\href{https://eliza.school/}{ELIZA}) through the DAAD program Konrad Zuse Schools of Excellence in Artificial Intelligence, sponsored by the German Federal Ministry of Education and Research.

\par\medskip
{\small
\bibliographystyle{ieeenat_fullname}
\bibliography{manuscript/references}
}
\clearpage
\appendix
\maketitlesupplementary
\paragraph{Overview.}
This supplement complements the main paper with technical details, experimental analyses, and qualitative results.
\begin{itemize}
    \item Section~\ref{sec:supp_modern_editors} describes the image-editing setting and the challenges posed by modern editors.
    \item Section~\ref{sec:supp_implementation_details} details the implementation and experimental configurations.
    \item Section~\ref{sec:supp_objective_ablation} presents ablations of the optimization objectives and gradient pathways.
    \item Section~\ref{sec:supp_multimodel} examines joint protection against multiple image editors.
    \item Section~\ref{sec:supp_robustness} evaluates how well protection holds up under image transformations.
    \item Section~\ref{sec:supp_budget_and_calibration} reports results for each of the three \vetobench{} categories, presents the full Pareto curves underlying the \editbench{} and \anyedit{} tables, and discusses perturbation budget selection and calibration.
    \item Section~\ref{sec:supp_gedit} validates the findings on real photographs from \textsc{GEditBench v2}.
    \item Section~\ref{sec:supp_vetobench_details} describes how we constructed \vetobench{} and what it contains.
    \item Section~\ref{supp:evaluation_details} describes the inference settings and evaluation protocols. It also reports inter-annotator agreement, breaks down edit failures by mode, and compares alternative protocols to assess the choices made in the main evaluation.
    \item Section~\ref{supp:additional_examples} presents additional qualitative examples and demonstrations.
\end{itemize}

\section{Functioning of Modern Unified Image Editors}
\label{sec:supp_modern_editors}

Modern unified image editors, such as \fluxtwo{} and
\fiboedit{}, combine latent flow matching with multimodal transformer
backbones
\citep{flux-2-2025,gutflaish2025generating,
lipman2023flowmatchinggenerativemodeling,esser2024scaling}. Earlier editing
pipelines typically invert a source image and then modify its generation
trajectory. By contrast, unified editors condition the entire generation
process directly on both the source image and the editing instruction.

\begin{figure}[h]
    \centering
    \includegraphics[width=\linewidth]{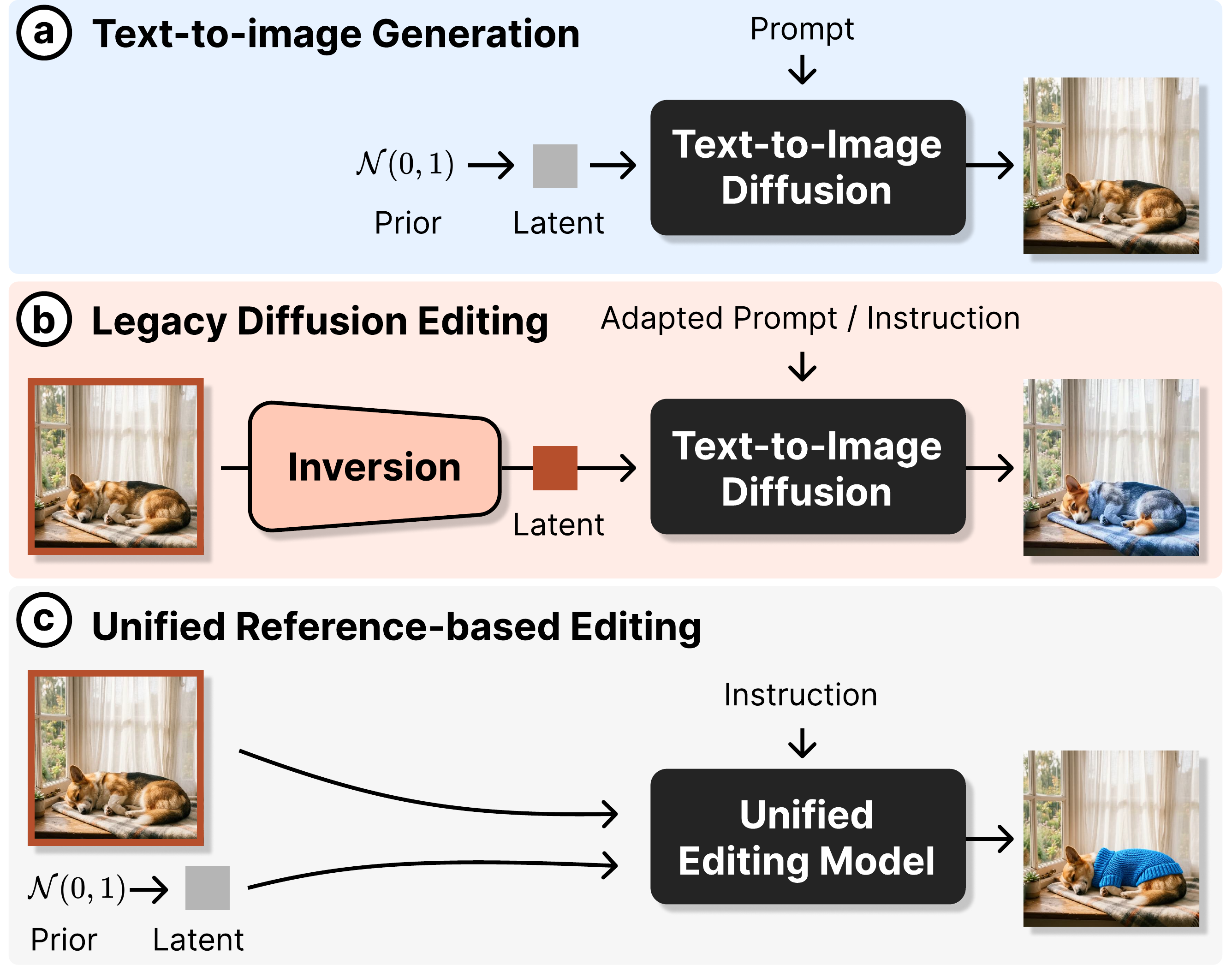}
    \caption{Generation paradigms: (a) pure text-to-image generation, (b) editing through source inversion, and (c) unified generation conditioned directly on image references.}
    \label{fig:flux2_paradigms}
\end{figure}

\subsection{Unified Editing as Conditional Flow Matching}

Rather than relying
on a separate inversion stage, these models use the source image as persistent
visual context while generating the edited output.

Let $x$ denote the source image and $p$ the editing instruction, with encoded
representations
\begin{equation}
\mathbf{x}=\mathcal{E}_{\mathrm{img}}(x),
\qquad
\mathbf{p}=\mathcal{E}_{\mathrm{text}}(p).
\end{equation}
The evolving output latent is denoted by $\mathbf{c}_t$ and referred to as the
\emph{canvas}. Starting from noise, the sampling process repeatedly evaluates
the learned velocity field
\begin{equation}
\mathbf{v}_t
=
g_\theta(\mathbf{c}_t,t;\mathbf{p},\mathbf{x})
\end{equation}
and advances the canvas along the predicted direction. Only the canvas changes
across sampling steps. The source and instruction remain fixed conditioning
inputs, and the final canvas latent is decoded into the edited image.

Within each model evaluation, however, the representations of all three token
groups evolve through the transformer blocks. Abstracting away
implementation-specific token ordering, we write
\begin{equation}
\mathbf{Z}_t^{(0)}
=
[\mathbf{p};\mathbf{c}_t;\mathbf{x}],
\qquad
\mathbf{Z}_t^{(\ell+1)}
=
B_\ell(\mathbf{Z}_t^{(\ell)},t).
\end{equation}
Attention therefore allows the canvas to interact directly with the source
image and instruction. The predictions corresponding to the canvas positions
determine the update to $\mathbf{c}_t$. At the next sampling step, the model is
evaluated again using the updated canvas and the same source and instruction
conditions. Thus, the canvas evolves across sampling steps, while the internal
representations evolve across transformer blocks and are recomputed at each
step.

\fiboedit{} follows this general mechanism but is designed around structured
JSON conditioning. In Bria's reference workflow, an optional, separate VLM
constructs this JSON from the source image and a short editing instruction
before sampling begins. This conversion is external to the core editing
pipeline; once constructed, the JSON prompt remains fixed throughout the
generation trajectory.

\paragraph{Source-canvas interaction.}

The \fluxtwo{} and \fiboedit{} architectures consist of \emph{double-stream}
blocks followed by \emph{single-stream} blocks. During the double-stream
stage, the instruction tokens form a textual stream, while the source and
canvas tokens form a visual stream. The two streams use modality-specific
normalization, QKV projections, and MLPs but compute attention over
concatenated keys and values. During the subsequent single-stream stage, the
textual and visual representations are merged and processed using shared
projections. In both stages, canvas queries can attend to source tokens, and
source queries can attend to canvas tokens.

\paragraph{\veto{} objective.}

These repeated interactions provide a direct target for image protection. An
objective that only displaces $\mathcal{E}_{\mathrm{img}}(x)$ does not
explicitly target how the editor retrieves and transfers source information
during generation. In contrast, \veto{} targets the attention pathways through
which source features interact with the evolving canvas across transformer
depth and sampling steps. Specifically, it maximizes the entropy contributions of the bidirectional
source--canvas interaction. The design ablations in
Supp.~\ref{sec:supp_objective_ablation} separate the attention statistic
from the gradient route.

\section{Implementation Details}
\label{sec:supp_implementation_details}
\label{sec:supp_veto_implementation}

This section provides additional implementation details and describes the computational infrastructure used to conduct our experiments.

\subsection{\veto{} Implementation}

Because the future editing instruction is unknown, \veto{} optimizes the perturbation
with an empty instruction and a simplified editing computation. At each update,
we encode the current protected image as both the reference and the canvas,
adding Gaussian noise to the latter as in Eq.~\ref{eq:veto_protection_surrogate}.
In the default \veto{} configuration, gradients through the reference encoding
are stopped, so the perturbation is updated through the canvas branch.
Both encodings are
recomputed as the image changes, and the loss still measures attention in
both directions. Each update averages gradients over ten timesteps; the optimizer and layer
choices are specified below.
For \fiboedit{}, we pair the empty instruction with one cached source
description. This accommodates the editor's native conditioning while avoiding
the need to construct a pool of possible future edits. Once the cloak is saved,
the attacker uses it as an ordinary input image: editing follows the model's
normal generation process, starting from noise.

We optimize a perturbation $\delta$ directly on the input image, subject to
\begin{equation}
\|\delta\|_\infty\leq\epsilon/255.
\end{equation}
\veto{} maximizes the objective introduced in
Equation~\ref{eq:veto_objective} using MI-FGSM
\citep{dong2018boosting}. Starting from $\delta^{(0)}$, the method maintains a
momentum vector
\begin{equation}
\mathbf{m}^{(k+1)}
=
\mu\mathbf{m}^{(k)}
+
\frac{\nabla_\delta\mathcal{L}_{\mathrm{VETO}}}
{\left\|\nabla_\delta\mathcal{L}_{\mathrm{VETO}}\right\|_1}
\end{equation}
and updates the perturbation as
\begin{equation}
\delta^{(k+1)}
=
\Pi_{\|\delta\|_\infty\leq\epsilon/255}
\left(
\delta^{(k)}
+
\alpha\,\mathrm{sign}\!\left(\mathbf{m}^{(k+1)}\right)
\right).
\end{equation}
The projection enforces the perturbation budget, after which the protected
image $x+\delta$ is clipped to the valid image range.

We implement the objective using lightweight hooks that intercept attention
in selected joint-transformer blocks while leaving all other model components
unchanged. Unless stated otherwise, we use a step size of $\alpha=2/255$, a
momentum decay of $\mu=0.9$, and $n=100$ optimization steps.

Both editors have eight double-stream blocks. We apply the
\veto{} loss to the first block in \fluxtwo{} (index~$0$) and to all eight
in \fiboedit{} (indices~$0$--$7$). In a separate \vetobench{} ablation at $\epsilon=12$, targeting block~0 alone already reduced edit success on \fiboedit{}. Targeting all eight yielded lower ESR and EditReward on both closed-frame and open-frame edits. One possible explanation is that the structured source description gives later \fiboedit{} blocks an additional cue for retrieving source details. %

\subsection{Baseline Implementations}

We re-implemented \editshield{} and \photoguard{} from their publicly available
code repositories and the implementation details in their respective papers,
adapting both methods to the unified image-editing models considered here.
We apply both objectives to each editor's native VAE latent representation.

The methods differ primarily in their objectives. \editshield{} pushes the
protected latent representation away from that of the original reference
image while regularizing the perturbation. \photoguard{}, by contrast, drives
the latent representation toward semantic alignment with an uninformative
gray target image. Both baselines use the same framework and share several
hyperparameters with \veto{}, including the five-budget sweep $\epsilon\in\{4,8,12,16,32\}$ and number of optimization steps $n=100$. The reported operating point is selected separately for each method, benchmark and editor.
The supplied \editshield{} recipe also applies stochastic image
transformations during optimization: Gaussian blur, small rotations or center cropping.

For \editshield{}, we identified a discrepancy between the supplementary
material of
\citeauthor{Chen2024undefined_EditShield_Protecting_Unauthorized} and the main
paper. The supplementary pseudocode includes gradient normalization, whereas
the main-text description does not include it. In our adapted implementation, normalization substantially reduced
the effective update magnitude. We therefore use the stronger-performing
variant without normalization in the primary comparison, which is a choice favorable
to the baseline.

For \decontext{}, we retain the layer selection used in
the original FLUX.1-Kontext implementation~\citep{shen2025decontext}.
Its attention-mass loss targets the first 26 of 38 single-stream blocks
(indices~$0$--$25$). We use the same indices in \fluxtwo{} and \fiboedit{},
which have 48 and 38 single-stream blocks, respectively. In all three
editors, the forward pass traverses the double-stream blocks before the
loss is evaluated in the selected single-stream blocks.

\subsection{Reproducibility}
\label{sec:reproducibility}

The main experiments used a node running Ubuntu 22.04.4 LTS with an
AMD EPYC 7313 16-Core Processor, 2.0 TiB of system memory, and eight NVIDIA
A100-SXM4-80GB GPUs, each with 80 GB of VRAM. Single-editor
cloaking uses one GPU; joint protection uses two, one per editor. We used CUDA Toolkit 12.4 and cuDNN 9.1.0 for GPU acceleration
and implemented the experiments in Python with PyTorch 2.6.0 and torchvision
0.21.0. Additional dependencies included Hugging Face Transformers 4.57.6,
Diffusers 0.37.1, and Accelerate 1.13.0. We managed all dependencies with the
\texttt{uv} package manager; the complete environment specification will be
released with the code.

For computational efficiency, we conducted the primary experiments at an
image resolution of $512\times512$ pixels. Figure~\ref{fig:veto_grid_flux_multi_res}
provides qualitative examples at higher resolutions and different aspect
ratios. Supp.~\ref{sec:supp_gedit} adds quantitative validation on a real-image dataset at varied resolutions.

Alongside the \veto{} source code, we will release the human-evaluation
interface, all configuration files, and the newly introduced
\vetobench{} dataset. The \vetobench{} release will include its source images,
source prompts, editing instructions, intended edited descriptions, domain
labels, and closed-/open-frame labels. The code release will contain scripts for applying each protection method, generating edited outputs, computing all metrics, and reproducing the tables and figures in this paper. It will also document the optimization parameters, model-specific attention hooks, baseline adaptations, and evaluation settings.

Together, these resources will support further research on protecting reference images against unauthorized editing.

\section{Objective and Gradient-Route Ablation}
\label{sec:supp_objective_ablation}

We examine the attention objective, attention direction, gradient
route and canvas construction within normal \veto{} on \fluxtwo{} at
$\epsilon=4$ and \fiboedit{} at $\epsilon=12$. Each setting uses all 300 \vetobench{} cases.
Table~\ref{tab:main_objective_ablation} reports pooled results;
Table~\ref{tab:objective_route_ablation} separates closed- and open-frame edits.

\begin{table}[t]
\centering
\begingroup
\providecommand{\mn}{\textminus}
\fontsize{8}{9.2}\selectfont
\setlength{\tabcolsep}{3pt}
\renewcommand{\arraystretch}{1.10}
\caption{Ablations from Table~\ref{tab:main_objective_ablation} on $150$ cases per editor and edit type. Each entry is the variant minus default \veto{} for the same editor and edit type. $\Delta$ESR is in percentage points and $\Delta$LPIPS in LPIPS units.}
\resizebox{\columnwidth}{!}{%
\begin{tabular}{@{}lrrrr@{}}
\toprule
& \multicolumn{2}{c}{Closed-frame} & \multicolumn{2}{c}{Open-frame} \\
\cmidrule(lr){2-3}\cmidrule(lr){4-5}
Setting & $\Delta$ESR $\downarrow$ & $\Delta$LPIPS $\downarrow$ & $\Delta$ESR $\downarrow$ & $\Delta$LPIPS $\downarrow$ \\
\midrule
\multicolumn{5}{@{}l}{\textbf{\fluxtwo{}, $\epsilon=4$}} \\
\midrule
\multicolumn{5}{@{}l}{\textbf{(a) Objective}} \\
\noalign{\begingroup\color{CombinedVetoRow}\hrule height10.12pt\endgroup\vskip-10.12pt}
Raw entropy (\veto{}) & \textbf{0.00} & \textbf{0.000} & 0.00 & \textbf{0.000} \\
Normalized entropy & +10.67 & +0.007 & +3.33 & +0.007 \\
Mass suppression & +10.67 & +0.002 & \textbf{\mn0.67} & +0.003 \\
\midrule
\multicolumn{5}{@{}l}{\textbf{(b) Bidirectionality}} \\
\noalign{\begingroup\color{CombinedVetoRow}\hrule height10.12pt\endgroup\vskip-10.12pt}
Both directions (\veto{}) & \textbf{0.00} & \textbf{0.000} & \textbf{0.00} & \textbf{0.000} \\
Canvas $\to$ Reference & +6.00 & +0.004 & +5.33 & +0.004 \\
Reference $\to$ Canvas & \textbf{0.00} & +0.001 & +0.67 & +0.001 \\
\midrule
\multicolumn{5}{@{}l}{\textbf{(c) Gradient route / canvas}} \\
\noalign{\begingroup\color{CombinedVetoRow}\hrule height10.12pt\endgroup\vskip-10.12pt}
Canvas / source (\veto{}) & \textbf{0.00} & \textbf{0.000} & \textbf{0.00} & \textbf{0.000} \\
Reference / source & +23.33 & +0.004 & +14.00 & +0.004 \\
Both / source & +20.67 & +0.009 & +10.00 & +0.009 \\
Reference / pure noise & +24.00 & +0.002 & +4.00 & +0.003 \\
\midrule[\heavyrulewidth]
\multicolumn{5}{@{}l}{\textbf{\fiboedit{}, $\epsilon=12$}} \\
\midrule
\multicolumn{5}{@{}l}{\textbf{(a) Objective}} \\
\noalign{\begingroup\color{CombinedVetoRow}\hrule height10.12pt\endgroup\vskip-10.12pt}
Raw entropy (\veto{}) & \textbf{0.00} & \textbf{0.000} & \textbf{0.00} & \textbf{0.000} \\
Normalized entropy & +10.00 & +0.064 & +10.00 & +0.059 \\
Mass suppression & +18.67 & +0.044 & +20.00 & +0.044 \\
\midrule
\multicolumn{5}{@{}l}{\textbf{(b) Bidirectionality}} \\
\noalign{\begingroup\color{CombinedVetoRow}\hrule height10.12pt\endgroup\vskip-10.12pt}
Both directions (\veto{}) & \textbf{0.00} & 0.000 & \textbf{0.00} & 0.000 \\
Canvas $\to$ Reference & +3.33 & +0.048 & +11.33 & +0.050 \\
Reference $\to$ Canvas & +14.67 & \textbf{\mn0.004} & +20.67 & \textbf{\mn0.008} \\
\midrule
\multicolumn{5}{@{}l}{\textbf{(c) Gradient route / canvas}} \\
\noalign{\begingroup\color{CombinedVetoRow}\hrule height10.12pt\endgroup\vskip-10.12pt}
Canvas / source (\veto{}) & \textbf{0.00} & \textbf{0.000} & 0.00 & 0.000 \\
Reference / source & +1.33 & +0.011 & \mn1.33 & +0.007 \\
Both / source & +1.33 & +0.013 & \textbf{\mn3.33} & +0.007 \\
Reference / pure noise & +2.67 & +0.003 & +2.00 & \textbf{\mn0.003} \\
\bottomrule
\end{tabular}}
\label{tab:objective_route_ablation}
\endgroup
\end{table}

\subsection{Entropy, Normalization, and Attention Mass}
\label{sec:supp_decontext_difference}

For one query and head, let $a_j$ be attention after softmax over \emph{all}
permitted keys. Let $K$ contain reference keys for a canvas query, or canvas
keys for a reference query. We define the cross-image attention mass and its
normalized distribution as
\begin{equation}
 m=\sum_{j\in K}a_j,\qquad q_j=\frac{a_j}{m}.
 \label{eq:ablation_attention_weights}
\end{equation}
The three losses are
\begin{align}
 \ell_{\mathrm{raw}}&=\sum_{j\in K}a_j\log(a_j+\tau),\\
 \ell_{\mathrm{norm}}&=\sum_{j\in K}q_j\log(q_j+\tau),\\
 \ell_{\mathrm{mass}}&=m,
 \label{eq:ablation_losses}
\end{align}
where $\tau=10^{-8}$. Minimizing the first two losses maximizes their
respective entropy; minimizing the third suppresses cross-image attention.
All three objectives retain \veto{}'s sums over queries and keys, means over batch and heads, and the same accumulation over directions, layers and timesteps. The normalized
variant rescales the extracted block to sum to one before computing entropy;
it leaves the transformer's forward attention unchanged. Equivalently,
$q_j$ is the softmax restricted to the keys in $K$.

Ignoring the logarithmic stabilizer, raw entropy satisfies
\begin{equation}
 H_{\mathrm{raw}}=-\sum_{j\in K}a_j\log a_j
                 =mH(q)-m\log m.
 \label{eq:supp_entropy_mass_comparison}
\end{equation}
Thus \veto{} can change both the spread and the mass of cross-image attention.
Normalized entropy directly optimizes the spread within the block, while mass suppression directly minimizes its total allocation. At fixed mass, raw entropy favors
more uniform weights.

\subsection{How Gradient Reaches the Protected Image}

\begin{figure}[t]
 \centering
 \includegraphics[width=\linewidth]{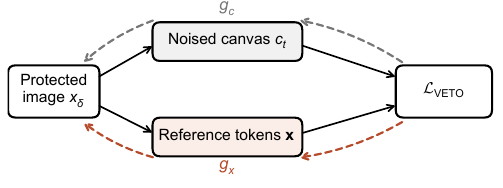}
 \caption{Gradient routing in the protection surrogate.
 Solid arrows show the shared forward computation; dashed arrows
 show the two possible gradient paths.}
 \label{fig:gradient_routes}
\end{figure}

At each update, we encode the current protected image $x_\delta=x+\delta$
with the editor's image encoder $\mathcal E_{\mathrm{img}}$, including its
native preprocessing. The resulting tokens $\mathbf{x}$ serve as the
reference, and their noised version forms the canvas $c_t$
(Eq.~\ref{eq:veto_protection_surrogate}, Fig.~\ref{fig:gradient_routes}).
\[
 \mathbf{x}=\mathcal E_{\mathrm{img}}(x_\delta),\qquad
  \mathbf{c}_t=(1-t)\mathbf{x}+t\eta,\qquad \eta\sim\mathcal N(0,I).
\]
Let $J$ be the Jacobian of $\mathcal E_{\mathrm{img}}$ with respect to
the input pixels at $x_\delta$.
The canvas route stops the reference gradient, and the reference route stops
the canvas gradient. For fixed $x$, $t$, and $\eta$, the chain
rule gives the two contributions to the gradient with respect to $\delta$.
Here $\nabla_{ \mathbf{c}_t}$ and $\nabla_{\mathbf{x}}$ each hold the other input to
the objective fixed.
\begin{align}
 g_c&=(1-t)J^{\mathsf T}\nabla_{ \mathbf{c}_t}\mathcal L_{\mathrm{VETO}},\\
 g_x&=J^{\mathsf T}\nabla_{\mathbf{x}}\mathcal L_{\mathrm{VETO}},\\
 g_{\mathrm{both}}&=\nabla_\delta\mathcal L_{\mathrm{VETO}}=g_c+g_x.
\end{align}
The joint route adds the two contributions before the common optimizer update,
without separate normalization or weighting. In all three
routes, the loss can measure attention in both directions, from canvas queries to reference keys and from reference queries to canvas keys. During protection, the factor $1-t$ gives the canvas gradient more weight at lower noise levels, where the surrogate canvas retains more of the source encoding.

\paragraph{Comparison setup.}
Every setting retains the standard \veto{} setup:
$512\times512$ inputs, $100$ updates, zero perturbation
initialization, step size $2/255$, momentum $0.9$, ten sampled timesteps per
update, and an empty instruction. \fluxtwo{} uses $\epsilon=4$ and double-stream
block $0$; \fiboedit{} uses $\epsilon=12$ and blocks $0$--$7$. Native precision, attention reductions, and optimizer settings are unchanged. Editing uses 28 steps and guidance 4 for \fluxtwo{}, and 30 steps and guidance 3.5 for \fiboedit{}, which derives its conditioning from the actual protected input. Each comparison varies only in its named component.

The pure-noise variant sets $\mathbf{c}_t=\eta$ and therefore has no
canvas-to-pixel gradient. In Table~\ref{tab:main_objective_ablation} (c), we compare it with the source-derived canvas using
reference gradients in both cases.
Comparing pure noise for the canvas input directly as in \decontext{} changes both \veto{}'s canvas construction and gradient routing.

\paragraph{Relation to \decontext{}.}
\decontext{}~\citep{shen2025decontext} minimizes canvas-to-reference attention mass during protection using a pure-noise canvas and sampled editing instructions. Since
that canvas is independent of the protected pixels, its image gradient enters
through the reference encoding. \veto{} derives both inputs from the protected
image, making either or both gradient paths available. Its noised source canvas
provides a direct differentiable proxy for cross-image interaction without
having to generate an edit for a guessed instruction. This construction is
used during protection; downstream editing follows the native editor.

Our mass-suppression variant tests \decontext{}'s objective
principle within \veto{}, retaining the source-derived canvas, empty instruction, early blocks and bidirectional attention. The main benchmark evaluation compares the complete methods. Our \decontext{} adaptation uses $26$ single-stream blocks and $60$ generic subject/scene instructions; the longer-optimization study below also tests its released update count of $800$.

\subsection{Ablation Results}

\paragraph{Objective and attention direction.}
Raw entropy reduces pooled ESR and LPIPS relative to normalized entropy and mass suppression on both editors.
{Table~\ref{tab:main_objective_ablation} reports pooled differences from default \veto{}, and Table~\ref{tab:objective_route_ablation} gives the corresponding differences within each edit type.}
For \fluxtwo{}, {pooled $\Delta\mathrm{ESR}$ is $+7.00\%$ for normalized entropy and $+5.00\%$ for mass suppression.}
Mass suppression has {lower open-frame ESR ($-0.67\%$) but higher closed-frame ESR ($+10.67\%$).}
Bidirectional attention gives the lowest pooled ESR and LPIPS. The reference-to-canvas direction alone is close {($\Delta\mathrm{ESR}=+0.33\%$)}, while canvas-to-reference alone {gives $+5.67\%$. The bidirectional and reference-to-canvas variants tie on closed-frame ESR. On open-frame edits, reference-to-canvas gives $\Delta\mathrm{ESR}=+0.67\%$.}

For \fiboedit{}, {normalized entropy and mass suppression give pooled $\Delta\mathrm{ESR}=+10.00\%$ and $+19.33\%$, respectively. Raw entropy also gives} lower LPIPS in both edit types.
Both attention directions give lower pooled ESR than either alone. {Reference-to-canvas alone gives slightly lower LPIPS ($\Delta\mathrm{LPIPS}=-0.006$).}
Canvas-to-reference is the stronger single direction {by ESR ($\Delta\mathrm{ESR}=+7.33\%$)}, whereas reference-to-canvas {gives $+17.67\%$}. This reverses the ordering of single directions on \fluxtwo{}.

\paragraph{Source-derived differentiable canvas.}
On \fluxtwo{}, canvas gradients yield {lower pooled ESR than reference gradients ($\Delta\mathrm{ESR}=+18.67\%$) or both paths ($+15.33\%$)}, while also giving lower LPIPS.
This advantage holds in both edit types.
Under reference-only routing, {switching from a source-derived to a pure-noise canvas reduces pooled \fluxtwo{} ESR ($-4.67\%$). The gain comes from open-frame edits ($-10.00\%$), while closed-frame ESR rises slightly ($+0.67\%$).}
On \fiboedit{}, reference gradients and both paths {give pooled $\Delta\mathrm{ESR}=0.00\%$ and $-1.00\%$, respectively, relative to canvas gradients.}
Canvas gradients provide lower LPIPS {($\Delta\mathrm{LPIPS}=+0.009$ for reference gradients and $+0.010$ for both paths). Both alternatives raise closed-frame ESR but lower open-frame ESR.}
With reference gradients fixed, {switching from a source-derived to a pure-noise canvas increases pooled ESR ($+2.33\%$).} Pure noise slightly lowers LPIPS but increases open-frame ESR {($+3.33\%$).}
Together, these results support the source-derived canvas as a differentiable protection surrogate: its dependence on the protected image enables the effective canvas-gradient route. The matched reference-gradient controls give lower ESR with noise on \fluxtwo{} and source-derived canvases on \fiboedit{}.
The results therefore support the canvas construction together with the optimization route it opens, with model-dependent protection--utility trade-offs.

\subsection{Optimization Budget and Protection Cost}
\label{sec:supp_decontext_steps}

\begin{figure}[t]
\centering
\includegraphics[width=\linewidth]{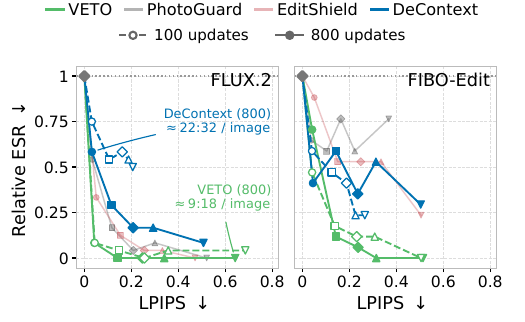}
\caption{Optimization budget on 30 \vetobench{} sources per editor.
Dashed and solid curves show 100 and 800 updates.
Faint \photoguard{} and \editshield{} curves provide context.
At 800 updates on \fluxtwo{}, protection averages
22\,min\,32\,s per image for \decontext{} and
9\,min\,18\,s for \veto{} on one A100 80GB.}
\label{fig:decontext_step_pareto}
\end{figure}

We test whether more optimization changes the comparison to \decontext{} using 30 fixed
\vetobench{} sources per editor: ten per category, equally split between
closed- and open-frame instructions.
We compare 100 and 800 updates at $\epsilon\in\{4,8,12,16,32\}$, keeping each
method's other settings fixed. The 800-update \decontext{} setting matches its
released number of steps.
The \veto{} comparison reuses the existing 100-update results and uses the same implementation for 800 updates.
Figure~\ref{fig:decontext_step_pareto} shows the full Pareto plots for both editors, while Table~\ref{tab:decontext_step_pareto} lists all the underlying results.

\begin{table}[t]
\centering
\small
\setlength{\tabcolsep}{2pt}
\renewcommand{\arraystretch}{1.08}
\caption{Optimization-budget study on 30 fixed \vetobench{} sources per editor.
ESR is in percent; lower ESR and LPIPS are better. \decontext{} protection takes over 20 minutes per image on \fluxtwo{} at 800 steps, still performing worse than \veto{}, with around 70 seconds per image at 100 steps.
}
\begin{tabular*}{\linewidth}{@{\extracolsep{\fill}}crrrrrrrr@{}}
\toprule
& \multicolumn{4}{c}{\veto{}} & \multicolumn{4}{c}{\decontext{}} \\
\cmidrule(lr){2-5}\cmidrule(lr){6-9}
& \multicolumn{2}{c}{100 updates} & \multicolumn{2}{c}{800 updates} & \multicolumn{2}{c}{100 updates} & \multicolumn{2}{c}{800 updates} \\
\cmidrule(lr){2-3}\cmidrule(lr){4-5}\cmidrule(lr){6-7}\cmidrule(lr){8-9}
$\epsilon$ & ESR & LPIPS & ESR & LPIPS & ESR & LPIPS & ESR & LPIPS \\
\midrule
\multicolumn{9}{l}{\textbf{\fluxtwo{}}\quad Original unprotected ESR: $80.0\%$} \\
\addlinespace[2pt]
4 & 6.7 & 0.044 & 6.7 & 0.042 & 60.0 & 0.032 & 46.7 & 0.032 \\
8 & 3.3 & 0.147 & 0.0 & 0.141 & 43.3 & 0.106 & 23.3 & 0.115 \\
12 & 0.0 & 0.256 & 0.0 & 0.247 & 46.7 & 0.161 & 13.3 & 0.208 \\
16 & 3.3 & 0.357 & 0.0 & 0.341 & 43.3 & 0.188 & 13.3 & 0.293 \\
32 & 3.3 & 0.686 & 0.0 & 0.642 & 40.0 & 0.206 & 6.7 & 0.507 \\
\midrule
\multicolumn{9}{l}{\textbf{\fiboedit{}}\quad Original unprotected ESR: {$56.7\%$}} \\
\addlinespace[2pt]
4 & {26.7} & 0.041 & {40.0} & 0.041 & {33.3} & 0.041 & {23.3} & 0.045 \\
8 & {10.0} & 0.138 & {6.7} & 0.143 & {26.7} & 0.124 & {33.3} & 0.143 \\
12 & {6.7} & 0.231 & {3.3} & 0.237 & {23.3} & 0.188 & {20.0} & 0.236 \\
16 & {6.7} & 0.308 & {0.0} & 0.315 & {13.3} & 0.226 & {30.0} & 0.313 \\
32 & {0.0} & 0.505 & {0.0} & 0.514 & {13.3} & 0.266 & {16.7} & 0.504 \\
\bottomrule
\end{tabular*}
\label{tab:decontext_step_pareto}
\end{table}

\paragraph{Protection and distortion.}
On \fluxtwo{}, more updates improve \decontext{} at every tested budget.
At $\epsilon=32$, its ESR falls from $40.0\%$ to $6.7\%$, while LPIPS rises from $0.206$ to $0.507$. \veto{} retains a favorable trade-off: at 800 updates, it reaches $6.7\%$ ESR at $\epsilon=4$ with LPIPS $0.042$, and no successful edits among the 30 cases at $\epsilon\geq8$. Its 100-update result already reaches $6.7\%$ at $\epsilon=4$ with LPIPS $0.044$.

On \fiboedit{}, longer optimization has mixed effects: \decontext{} improves at $\epsilon=4$ and $12$ but worsens at the other budgets; \veto{} improves at $8$, $12$ and $16$, ties at $32$, and worsens at $4$.
Thus matching update counts alone does not establish a method's attainable protection.

\begingroup
\paragraph{Runtime and memory.}
On a single A100 80GB, the \fluxtwo{} cloaks at 800 updates average $22{:}32$ per image for \decontext{} and $9{:}18$ for \veto{}.
The \decontext{} paper specifies 800 updates on an A800 80GB but reports no per-image runtime~\citep{shen2025decontext}. At 100 updates, the \fluxtwo{} runs average $2{:}49$ for \decontext{}, while for \veto{} we measured $70$ seconds. This made \decontext{} at 800 updates infeasible for the full-scale evaluation, but the small-scale results already indicate that \veto{}'s Pareto frontier generally outperforms \decontext{} on both editors, even if \veto{} uses only 100 steps.

These timings exclude model loading and downstream editing. In particular, \decontext{} requires 60 image-specific conditions per \fiboedit{} source, whereas \veto{} uses one empty-instruction condition. The reported times therefore describe protection-stage cost, not complete first-use latency.

Peak allocated CUDA memory averages $34.1$\,GiB for \veto{} and $66.2$\,GiB for \decontext{} on \fluxtwo{}, at both 100 and 800 updates.

\endgroup

\begingroup
\section{Joint Protection against Multiple Editors}
\label{sec:supp_multimodel}

An owner cannot always anticipate which editing model will be used, which is why a cloak should ideally protect against multiple popular editing models at once.
To test this, we jointly target
the generative computations of two distinct unified editors, without assuming
that either operator enforces a moderation or authorization policy. We extend \veto{} to optimize one shared protected image against both \fluxtwo{}
and \fiboedit{} simultaneously by retaining both per-editor
objectives and combining their pixel gradients in a shared projected update.

\paragraph{Joint optimization.}
Let $g_f(x)$ denote the pixel gradient of the negative \veto{} entropy objective
for editor $f$, which we minimize. We normalize each gradient by its mean absolute magnitude and average them:
\begin{equation}
 \bar g(x)=\frac{1}{2}\sum_{m=1}^{2}
 \frac{g_{f_m}(x)}{\operatorname{mean}(|g_{f_m}(x)|)+10^{-8}}.
 \label{eq:multimodel_gradient}
\end{equation}
This gives both editors equal weight without allowing a difference in gradient
scale alone to dominate the update.
Each update follows the negative sign of the accumulated gradient and projects
the shared pixels onto both the valid image range and the source-centered
$\ell_\infty$ ball of radius $\epsilon/255$. We compare $100$, $200$ and $400$ updates at $\epsilon=8$, and $200$ and $400$ updates at $\epsilon=12$. All other optimization and editing settings are unchanged.

\begin{table}[t]
\centering\begingroup
\setlength{\tabcolsep}{5pt}
\renewcommand{\arraystretch}{1.02}
\caption{{Joint \veto{} protection on all $300$ \vetobench{} cases. Both editors share one cloak and receive equal optimization weight.}}
\begin{tabular}{@{}rrrrr@{}}
\toprule
\multirow{2}{*}{Updates} & \multirow{2}{*}{$\epsilon$} & \multicolumn{2}{c}{ESR (\%, $\downarrow$)} & \multirow{2}{*}{LPIPS $\downarrow$} \\
\cmidrule(lr){3-4}
& & \fluxtwo{} & \fiboedit{} & \\
\midrule
100 & 8 & {\textbf{1.00}} & {22.33} & \textbf{0.154} \\
200 & 8 & {1.67} & {20.33} & 0.156 \\
200 & 12 & {1.67} & {\textbf{7.67}} & 0.263 \\
{400} & {8} & {1.67} & {20.00} & {0.157} \\
{400} & {12} & {1.33} & {8.33} & {0.264} \\
\bottomrule
\end{tabular}
\label{tab:multimodel_protection}
\endgroup\end{table}

\paragraph{Results and scope.}
We use all $300$ \vetobench{} sources and instructions, with $100$
cases per category and $150$ per edit type. In each setting, a source yields one protected
PNG at $512\times512$ resolution, and that identical file is passed to both
editors, giving $600$ edited outputs per setting.
Table~\ref{tab:multimodel_protection} shows that a single cloak suppresses editing
by both targeted models. With $100$ updates at $\epsilon=8$, only $3$ of $300$ joint-protection edits on \fluxtwo{}
succeed. On \fiboedit{}, the shared cloak yields $22.33\%$ ESR at the same budget and update count, while remaining far below the unprotected rate of $58.00\%$. Doubling the updates at the same budget changes ESR to $1.67\%$ on \fluxtwo{} and $20.33\%$ on \fiboedit{}.
At $200$ updates and $\epsilon=12$, the shared cloak reaches $1.67\%$ and
$7.67\%$ ESR. Increasing the budget from $8$ to $12$ raises LPIPS from $0.156$ to $0.263$.
At $400$ updates, ESR is $1.67\%$/$20.00\%$ at $\epsilon=8$ and $1.33\%$/$8.33\%$ at $\epsilon=12$ for \fluxtwo{}/\fiboedit{}, respectively.
The corresponding LPIPS values are $0.157$ and $0.264$.
The larger budget strengthens joint protection on \fiboedit{}, while doubling from $200$ to $400$ updates provides no consistent improvement across both editors.

Joint optimization therefore offers a feasible approach to protecting a single image against multiple known editors. Separately, Table~\ref{tab:veto_transferability_flux_variants} shows that single-editor \fluxtwo{} cloaks transfer to the tested downstream fine-tunes, distilled and smaller variants without reoptimization. These findings suggest that jointly optimizing against popular models could provide broader protection across entire model families and their ecosystems of community-derived checkpoints.

\endgroup

\section{Robustness}
\label{sec:supp_input_robustness}
\label{sec:supp_robustness}

A fixed protected image may encounter changes to the downstream editor.  Table~\ref{tab:veto_transferability_flux_variants} in the main paper
tests this robustness against adapted model variants: cloaks optimized against quantized \fluxtwo{} retain protection
with the evaluated precision, distillation and style-LoRA changes, and with the
smaller \textsc{FLUX.2-Klein-4B} from the same model family, even though it is not a direct descendant of the optimization checkpoint.
The following sections examine robustness to image transformations, adversarial purification and repeated edit attempts.

\subsection{Image-Transform Robustness}

Image transformations affect not only the source image but also its embedded perturbation.
This poses a well-known challenge for input protection methods
\citep{Salman2023ICML_Raising_the_Cost,Choi2025DiffusionGuard,Kim2025undefined_BlurGuard_A_Simple}.
It is therefore important to ensure that protective perturbations remain effective under common image transformations.
We demonstrate how \veto{} can be extended to improve its robustness through existing specialized techniques or augmentation during perturbation optimization.

\paragraph{Orthogonal robustification with \blurguard{}.}
\blurguard{}~\citep{Kim2025undefined_BlurGuard_A_Simple} augments an existing
cloaking objective with adaptive regional Gaussian filtering of its perturbation
and spectral regularization; its editing experiments use \photoguard{}'s
encoder objective. We therefore treat it as an orthogonal robustification
technique rather than a separate cloaking baseline.

\begin{table*}[t]
\centering
\fontsize{8}{9.4}\selectfont
\setlength{\tabcolsep}{1.6pt}
\renewcommand{\arraystretch}{1.12}
\caption{Robustness to input transformations on all $300$ \vetobench{} cases with \fluxtwo{}. For each transformation, relative ESR (rESR) divides the edit success rate by the unprotected rate under the same transformation. ER is mean EditReward in its native units. Lower values of both metrics favor protection. LPIPS measures input distortion before transformation.}
\begin{tabular*}{\textwidth}{@{\extracolsep{\fill}}lcl*{7}{rr}r@{}}
\toprule
 &  &  & \multicolumn{2}{c}{None} & \multicolumn{2}{c}{Crop} & \multicolumn{2}{c}{Resize} & \multicolumn{2}{c}{Rotation} & \multicolumn{2}{c}{JPEG} & \multicolumn{2}{c}{GrIDPure} & \multicolumn{2}{c}{JPEG+Resize} & Utility \\
\cmidrule(lr){4-5}\cmidrule(lr){6-7}\cmidrule(lr){8-9}\cmidrule(lr){10-11}\cmidrule(lr){12-13}\cmidrule(lr){14-15}\cmidrule(lr){16-17}
Method & $\epsilon$ & Recipe & rESR & ER & rESR & ER & rESR & ER & rESR & ER & rESR & ER & rESR & ER & rESR & ER & LPIPS \\
\midrule
 & 32 & None & 0.39 & {{{0.168}}} & 0.75 & 0.473 & 0.61 & 0.400 & 0.62 & 0.248 & 0.82 & 0.475 & 0.89 & 0.607 & 0.81 & 0.469 & 0.223 \\
\decontext{} & 32 & \blurguard{} & 0.88 & 0.605 & 0.92 & 0.581 & 0.85 & 0.579 & 0.89 & 0.383 & 0.98 & 0.565 & 0.97 & 0.647 & 0.95 & 0.558 & 0.053 \\
 & 32 & EOT & 0.36 & {{{0.142}}} & 0.42 & 0.233 & 0.38 & 0.223 & 0.39 & 0.074 & 0.43 & 0.255 & 0.67 & 0.458 & 0.50 & 0.249 & 0.272 \\
\midrule
\multirow{9}{*}{\textbf{\veto{}}} & 4 & None & 0.04 & {{{\textminus0.406}}} & 0.96 & 0.624 & 0.93 & 0.664 & 0.95 & 0.419 & 0.99 & 0.636 & 0.98 & 0.613 & 1.02 & 0.613 & 0.054 \\
 & 4 & \blurguard{} & 0.43 & 0.182 & 0.89 & 0.547 & 0.61 & 0.309 & 0.84 & 0.318 & 0.93 & 0.544 & 0.99 & 0.614 & 0.95 & 0.524 & 0.077 \\
 & 4 & EOT & 0.26 & \textminus0.029 & 0.58 & 0.373 & 0.47 & 0.272 & 0.59 & 0.196 & 0.82 & 0.539 & 1.01 & 0.630 & 0.93 & 0.542 & 0.064 \\
\cmidrule(l){2-18}
 & 8 & None & {{{0.01}}} & {{{\textminus0.471}}} & 0.89 & 0.567 & 0.91 & 0.619 & 0.89 & 0.333 & 0.94 & 0.603 & 0.98 & 0.603 & 0.99 & 0.569 & 0.167 \\
 & 8 & \blurguard{} & 0.04 & \textminus0.445 & 0.76 & 0.500 & 0.09 & \textminus0.309 & 0.77 & 0.256 & 0.87 & 0.483 & 0.94 & 0.564 & 0.87 & 0.450 & 0.186 \\
 & 8 & EOT & 0.03 & \textminus0.383 & 0.30 & 0.071 & 0.09 & \textminus0.268 & 0.35 & \textminus0.062 & 0.49 & 0.230 & 0.94 & 0.597 & 0.55 & 0.278 & 0.190 \\
\cmidrule(l){2-18}
 & 16 & None & 0.01 & {{{\textminus0.531}}} & 0.74 & 0.443 & 0.79 & 0.488 & 0.75 & 0.199 & 0.87 & 0.502 & 0.84 & 0.590 & 0.91 & 0.458 & 0.382 \\
 & 16 & \blurguard{} & 0.01 &\textminus0.596 & 0.64 & 0.370 & 0.03 &\textminus0.466 & 0.54 & 0.110 & 0.72 & 0.338 & 0.65 & 0.400 & 0.59 & 0.286 & 0.372 \\
 & 16 & EOT & 0.02 &\textminus0.442 & 0.10 &\textminus0.172 & 0.02 &\textminus0.403 & 0.11 &\textminus0.257 & 0.10 &\textminus0.231 & 0.53 & 0.276 & 0.16 &\textminus0.161 & 0.420 \\
\bottomrule
\end{tabular*}
\label{tab:robustness_extension}
\end{table*}

\paragraph{Augmentation during perturbation optimization.}
Beyond \blurguard{}, we apply expectation over transformation (EOT)~\citep{athalye2018eot, Chen2024undefined_EditShield_Protecting_Unauthorized}
to \veto{}'s attention objective, maximizing its mean over several differentiable views $T\in\mathcal{T}$ of the
protected image:
\begin{equation}
 \max_{\|\delta\|_\infty\leq\epsilon/255}
 \frac{1}{|\mathcal T|}\sum_{T\in\mathcal T}
 \mathcal L_{\mathrm{VETO}}\bigl(T(x+\delta)\bigr).
\label{eq:targeted_augmentation}
\end{equation}
The identity view preserves pressure on direct editing. Gradients from the
individual views are mapped back to the original pixels and averaged before the
momentum/sign update. The pixel constraint is imposed on the original cloak and the editor remains frozen.

The EOT implementation uses five views at each update: the original protected image and one
freshly sampled crop, resize, rotation and JPEG compression. Crop scales range
from $0.78$ to $0.98$ with random positions, resize scales from $0.45$ to $0.95$,
rotation angles from $-15^\circ$ to $15^\circ$, and JPEG quality from $35$ to $95$.
All views within an update share the same timestep and noise
draws. JPEG uses the exact compressed image in the forward pass and a
straight-through gradient approximation.

\paragraph{Experimental Setup.}
The comparison retains complete robustification recipes:
\blurguard{} uses sign PGD without momentum, while None and EOT use
momentum $0.9$. All three are evaluated under the same processing conditions. Evaluation applies each operation separately to the $512\times512$ inputs,
using fixed draws shared by all methods: an off-center crop of scale $0.932$,
down/up-resizing by $0.831$, a $7.86^\circ$ rotation, and JPEG quality $71$.
These operations and editing seeds are fixed across the compared settings.
Thus augmentation strengths vary during optimization, whereas the reported
test conditions remain fixed.

We also evaluate two more challenging processing pipelines: adversarial image purification and an approximate social-media processing pipeline.
For diffusion purification, we apply \gridpure{}~\cite{zhao2024can} with
noise level $T=10$, $20$ iterations, and blending weight $\gamma=0.1$. To approximate social-media processing, we combine JPEG compression with subsequent resizing. Both pipelines use the existing cloaks and the same editing instructions and seeds as the
other conditions.

\paragraph{Results.}
Table~\ref{tab:robustness_extension} evaluates protection on all $300$ \vetobench{} cases under six processing conditions and without preprocessing. Relative ESR (rESR) divides the protected edit success rate by the reference rate under the same condition. Lower rESR and EditReward (ER) indicate stronger protection, while lower input LPIPS indicates better fidelity. At all three budgets, EOT lowers both rESR and ER relative to plain \veto{} after cropping, resizing, rotation, and JPEG compression. These gains extend to JPEG followed by resizing, although EOT samples the two transformations separately during optimization. At $\epsilon=8$, EOT lowers rESR from $0.91$ to $0.09$ after resizing and from $0.99$ to $0.55$ under the combined pipeline. ER falls from $0.619$ to $-0.268$ and from $0.569$ to $0.278$, respectively. At $\epsilon=16$, EOT reaches rESR of $0.02$--$0.11$ across the four individual transformations and $0.16$ under the combined pipeline. Without preprocessing, EOT raises rESR from $0.01$ to $0.03$ at $\epsilon=8$, while input LPIPS increases from $0.167$ to $0.190$. The increase in rESR is larger at $\epsilon=4$ ($0.04$ to $0.26$).

\veto{} with EOT at $\epsilon=16$ also lowers both metrics relative to plain \decontext{} at $\epsilon=32$ in all seven conditions. When comparing the two methods with EOT, \veto{} at $\epsilon=8$ introduces less input distortion than \decontext{} at $\epsilon=32$ (LPIPS $0.190$ versus $0.272$) and achieves lower rESR and ER without preprocessing and after cropping, resizing, and rotation. After resizing, for example, \veto{} reaches rESR $0.09$ and ER $-0.268$, compared with $0.38$ and $0.223$ for \decontext{}. At these operating points, \decontext{} achieves lower rESR after JPEG compression, \gridpure{} purification, and JPEG followed by resizing.

After \gridpure{} purification, the benefit of EOT for \veto{} depends more strongly on the budget. Neither metric improves over plain \veto{} at $\epsilon=4$. At $\epsilon=8$, rESR falls from $0.98$ to $0.94$, while ER changes little ($0.603$ to $0.597$). The clearest gain occurs at $\epsilon=16$, where EOT lowers rESR from $0.84$ to $0.53$ and ER from $0.590$ to $0.276$, despite never applying \gridpure{} during optimization. Under the same purification, \decontext{} with EOT reaches rESR $0.67$ and ER $0.458$.

\paragraph{Implications.}
For image owners, reducing the likelihood of a successful unauthorized edit can be useful even when some edits still succeed. With EOT, \veto{} retains protection after JPEG compression followed by resizing, although it encounters the two operations separately during optimization. At $\epsilon=8$, \veto{} with EOT better withstands resizing and preserves higher input fidelity than \decontext{} with EOT at $\epsilon=32$, while \decontext{} offers stronger protection after purification. At $\epsilon=16$, \veto{} with EOT reduces post-purification edit success to roughly half the reference rate, at the cost of greater input distortion. Image owners can choose a perturbation budget that reflects the sensitivity of the content they share, favoring fidelity for routine images or accepting more distortion when unauthorized edits could cause greater harm.

\subsection{Repeated-Edit Robustness}
\label{sec:repeated_edit_attempts}

\begin{table}[t]
\centering
\begingroup
\fontsize{8}{9.2}\selectfont
\setlength{\tabcolsep}{2.5pt}
\renewcommand{\arraystretch}{1.10}
\caption{Repeated \fluxtwo{} edits on \vetobench{} ($300$ pairs; $100$ per category).
$N=1$ reports absolute ESR; $N=2$--$5$ report the increase in any-of-$N$ ESR
over $N=1$ in percentage points, computed before rounding.
An edit succeeds if any attempt succeeds. Each attempt reuses the protected input:
$\epsilon=4$ for \veto{} and $32$ for \decontext{}.
Bold marks the lower ESR or smaller increase.}
\begin{tabular*}{\columnwidth}{@{}ll@{\extracolsep{\fill}}rrrrr@{}}
\toprule
& & ESR (\%) & \multicolumn{4}{c}{$\Delta$ESR (pp, $\downarrow$)} \\
\cmidrule(lr){3-3}\cmidrule(lr){4-7}
Category & Method & $N=1$ & $N=2$ & $N=3$ & $N=4$ & $N=5$ \\
\midrule
Overall & \decontext{} & 31.3 & +9.7 & +12.3 & +17.3 & +20.3 \\
 & \textbf{\veto{}} & \textbf{3.3} & \textbf{+2.3} & \textbf{+3.0} & \textbf{+3.7} & \textbf{+4.0} \\
\midrule
General & \decontext{} & 54.0 & +13.0 & +15.0 & +18.0 & +22.0 \\
 & \textbf{\veto{}} & \textbf{6.0} & \textbf{+4.0} & \textbf{+5.0} & \textbf{+6.0} & \textbf{+7.0} \\
\midrule
Defamation & \decontext{} & 17.0 & +5.0 & +8.0 & +15.0 & +16.0 \\
 & \textbf{\veto{}} & \textbf{2.0} & \textbf{+1.0} & \textbf{+1.0} & \textbf{+1.0} & \textbf{+1.0} \\
\midrule
Gore & \decontext{} & 23.0 & +11.0 & +14.0 & +19.0 & +23.0 \\
 & \textbf{\veto{}} & \textbf{2.0} & \textbf{+2.0} & \textbf{+3.0} & \textbf{+4.0} & \textbf{+4.0} \\
\bottomrule
\end{tabular*}%
\label{tab:repeated_edit_attempts}
\endgroup
\end{table}

We test whether protection persists when the quantized \fluxtwo{}
editor is given multiple opportunities to produce the requested edit.
We use all $300$ \vetobench{} source--instruction pairs ($100$ per category)
at the selected budgets $\epsilon=4$ for \veto{} and $\epsilon=32$ for
\decontext{}. Each attempt starts from the same protected input and uses
the same instruction. We retain the original first edit and generate four
additional edits with different seeds. Success in any of the first $N$
attempts counts as any-of-$N$ success.

Table~\ref{tab:repeated_edit_attempts} shows that \veto{} maintains
low edit success over five attempts: overall ESR increases from $3.3\%$
to $7.3\%$ ($+4.0$ percentage points), compared with $31.3\%$ to $51.7\%$
($+20.3$ points) for \decontext{}. For defamation and gore, five-attempt
ESR is $3\%$ and $6\%$ for \veto{}, versus $33\%$ and $46\%$ for
\decontext{}, respectively.

\section{Budget Selection and Calibration}
\label{supp:pareto_ae_eb}
\label{sec:supp_budget_and_calibration}

\begingroup
\providecommand{\mn}{\textminus}
\color{black}
\begin{table*}[t]
\centering
\fontsize{8}{9.2}\selectfont
\setlength{\tabcolsep}{2pt}
\renewcommand{\arraystretch}{1.10}
\caption{\vetobench{} (Ours) results by category.
Each category contains $100$ cases and retains the full-benchmark budgets in
Table~\ref{tab:combined_results}. Metrics describe the displayed category;
$D_\epsilon$ uses its unprotected ESR. ESR is in percent. Bold marks the best
protected values, including ties; gray highlights \veto{}.}
\begin{tabular*}{\textwidth}{@{\extracolsep{\fill}}lrrrrrrr@{\hspace{9pt}}rrrrrrr@{}}
\toprule
& \multicolumn{7}{c}{\textbf{\fluxtwo{}}} & \multicolumn{7}{c}{\textbf{\fiboedit{}}} \\
\cmidrule(lr){2-8}\cmidrule(lr){9-15}
& Budget & \multicolumn{3}{c}{Edit Success ($\downarrow$)} & \multicolumn{2}{c}{Utility} & Trade-off
       & Budget & \multicolumn{3}{c}{Edit Success ($\downarrow$)} & \multicolumn{2}{c}{Utility} & Trade-off \\
\cmidrule(lr){3-5}\cmidrule(lr){6-7}\cmidrule(lr){10-12}\cmidrule(lr){13-14}

Method & $\epsilon$ & MLLM & Human & ER & LPIPS $\downarrow$ & PSNR $\uparrow$ & $D_\epsilon \downarrow$
       & $\epsilon$ & MLLM & Human & ER & LPIPS $\downarrow$ & PSNR $\uparrow$ & $D_\epsilon \downarrow$ \\
\midrule
\multicolumn{15}{l}{\textbf{General ($n=100$)}} \\
Unprotected & 0 & 91 & 94 & 1.415 & 0.00 & $\infty$ & 1.000 & 0 & 82 & 94 & 1.305 & 0.00 & $\infty$ & 1.000 \\
\photoguard{} & 12 & 21 & 18 & 0.508 & 0.24 & 28.99 & 0.329 & 32 & 61 & 83 & 1.001 & 0.39 & 23.69 & 0.838 \\
\editshield{} & 8 & 10 & 16 & 0.179 & 0.19 & 32.29 & 0.218 & 32 & 33 & 56 & 0.899 & 0.56 & 22.39 & 0.689 \\
\decontext{} & 32 & 54 & 73 & 0.854 & 0.24 & 30.22 & 0.639 & 16 & 33 & 47 & 0.774 & \textbf{0.26} & \textbf{29.45} & 0.479 \\
\noalign{\begingroup\color{CombinedVetoRow}\hrule height10.12pt\endgroup\vskip-10.12pt}
\textbf{\veto{} (Ours)} & 4 & \textbf{6} & \textbf{9} & \textbf{\mn0.008} & \textbf{0.06} & \textbf{36.63} & \textbf{0.088} & 12 & \textbf{23} & \textbf{33} & \textbf{0.288} & 0.27 & 27.84 & \textbf{0.389} \\
\midrule
\multicolumn{15}{l}{\textbf{Defamation ($n=100$)}} \\
Unprotected & 0 & 75 & 92 & 0.504 & 0.00 & $\infty$ & 1.000 & 0 & 47 & 56 & 0.184 & 0.00 & $\infty$ & 1.000 \\
\photoguard{} & 12 & \textbf{1} & 2 & \mn0.506 & 0.20 & 29.06 & 0.201 & 32 & 8 & 23 & \mn0.349 & 0.35 & 23.40 & 0.393 \\
\editshield{} & 8 & 2 & 6 & \mn0.609 & 0.14 & 32.28 & 0.139 & 32 & 6 & 13 & \mn0.356 & 0.47 & 22.51 & 0.490 \\
\decontext{} & 32 & 17 & 24 & \mn0.322 & 0.19 & 30.02 & 0.299 & 16 & 8 & 18 & \mn0.500 & \textbf{0.21} & \textbf{29.39} & 0.273 \\
\noalign{\begingroup\color{CombinedVetoRow}\hrule height10.12pt\endgroup\vskip-10.12pt}
\textbf{\veto{} (Ours)} & 4 & 2 & \textbf{1} & \textbf{\mn0.727} & \textbf{0.04} & \textbf{36.71} & \textbf{0.048} & 12 & \textbf{4} & \textbf{3} & \textbf{\mn0.755} & 0.22 & 27.95 & \textbf{0.233} \\
\midrule
\multicolumn{15}{l}{\textbf{Gore ($n=100$)}} \\
Unprotected & 0 & 77 & 81 & 0.559 & 0.00 & $\infty$ & 1.000 & 0 & 45 & 67 & 0.385 & 0.00 & $\infty$ & 1.000 \\
\photoguard{} & 12 & 14 & 8 & \mn0.098 & 0.24 & 29.10 & 0.299 & 32 & 26 & 45 & 0.014 & 0.38 & 23.92 & 0.693 \\
\editshield{} & 8 & 5 & 7 & \mn0.336 & 0.19 & 32.39 & 0.197 & 32 & 14 & 32 & \mn0.072 & 0.55 & 22.49 & 0.629 \\
\decontext{} & 32 & 23 & 36 & \mn0.027 & 0.24 & 30.28 & 0.382 & 16 & 5 & 17 & \mn0.210 & \textbf{0.26} & \textbf{29.61} & \textbf{0.283} \\
\noalign{\begingroup\color{CombinedVetoRow}\hrule height10.12pt\endgroup\vskip-10.12pt}
\textbf{\veto{} (Ours)} & 4 & \textbf{2} & \textbf{2} & \textbf{\mn0.481} & \textbf{0.06} & \textbf{36.68} & \textbf{0.068} & 12 & \textbf{3} & \textbf{8} & \textbf{\mn0.572} & 0.28 & 27.90 & \textbf{0.283} \\
\bottomrule
\end{tabular*}
\label{tab:vetobench_scenario_breakdown}
\end{table*}
\endgroup

For a more fine-grained analysis, Table~\ref{tab:vetobench_scenario_breakdown} breaks down the results across the three \vetobench{} categories ($100$ cases each), using the pooled-selected budgets from Table~\ref{tab:combined_results}. \veto{} consistently achieves the best overall trade-off between edit success and utility preservation across categories and models. Edit success is generally highest in the General category, whereas protection reduces edit success most strongly in the other two categories. This may be due to their greater emphasis on preserving identities, particularly in defamation cases, which can make edits more susceptible to failure.

\begin{figure}[t]
    \centering
    \includegraphics[width=\linewidth]{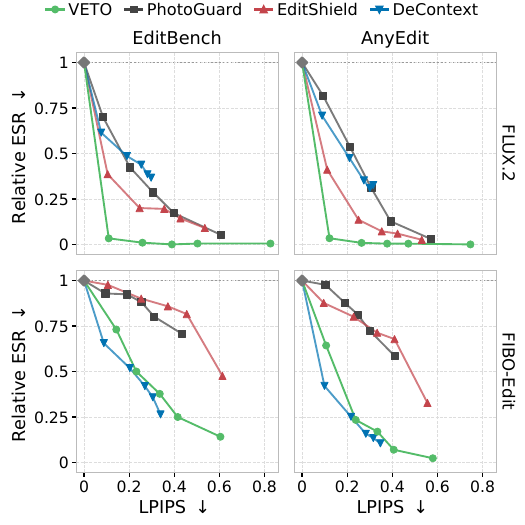}
    \caption{Trade-offs on \editbench{} (left) and \anyedit{} (right), $300$ cases each. Rows show \fluxtwo{} and \fiboedit{}. Lines connect $\epsilon=0,4,8,12,16,32$. ESR is relative to unprotected performance. LPIPS measures input distortion.}
    \label{fig:pareto_all}
\end{figure}

Figure~\ref{fig:pareto_all} gives the complete \anyedit{} and \editbench{} budget curves underlying the selected operating points in Table~\ref{tab:combined_results}. On \fluxtwo{}, \veto{} combines low residual ESR with lower distortion than the compared defenses. On \fiboedit{}, \decontext{} is competitive: its selected \editbench{} point has similar ESR at lower LPIPS, while its \anyedit{} point has lower ESR at higher LPIPS than \veto{}. The operating-point selection criterion favors \decontext{} on both \fiboedit{} datasets. These outcomes support an editor-dependent comparison rather than a uniform ordering of the methods. \fiboedit{}'s structured source descriptions may contribute to this difference, but the cross-editor comparison does not isolate their effect.

\subsection{Aggregated Budget Selection}
\label{sec:supp_budget_selection}

The main tables select one budget per method, benchmark and editor using all
$300$ benchmark cases, then retain it across categories and edit types.
These operating points summarize the measured protection--fidelity trade-off;
the complete budget curves show how that trade-off changes with distortion.
For deployment, a common budget would instead be fixed using representative
calibration tasks before the actual downstream editing instruction is known.
This assumes that the calibration tasks and known editor represent future use;
it does not guarantee protection against arbitrary instructions.

\begin{figure*}[t]
    \centering
    \includegraphics[width=\textwidth]{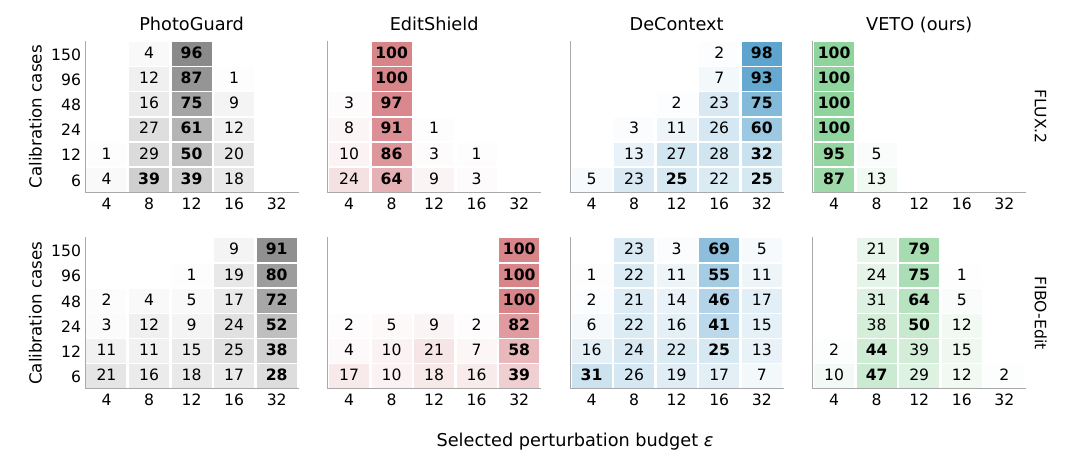}
    \caption{Distribution of selected perturbation budgets on \vetobench{} for \fluxtwo{} (top) and \fiboedit{} (bottom). Each panel shows one protection method. Within a panel, each row gives the percentage of $100$ balanced random subsets of the indicated size that selected each budget $\epsilon$. Darker cells indicate more frequent selections. Bold entries mark the most frequent budget in each row and blanks denote zero.}
    \label{fig:budget_selection_distributions}
\end{figure*}

\paragraph{Budget selection on calibration subsets.}
We tested how often a smaller calibration set selects the same perturbation budget as the full $300$-case \vetobench{} evaluation. For each of $100$ random draws, we formed nested subsets of $150$, $96$, $48$, $24$, $12$, and $6$ cases, with equal representation across the six combinations of category and edit type. All methods and both editors used the same subsets. On each subset, we selected the budget that minimized $D_\epsilon$ for each method and editor. Figure~\ref{fig:budget_selection_distributions} shows how the selections were distributed across all five tested budgets. On \fluxtwo{}, \veto{} recovered $\epsilon=4$ in every draw with at least $24$ cases and in $87$ of $100$ draws with just one case per category and edit type. On \fiboedit{}, all $150$-case draws selected one of two neighboring budgets: $79$ selected the full-set choice of $\epsilon=12$ and $21$ selected $\epsilon=8$. With six calibration cases, the selections spanned all five budgets.

\paragraph{Coverage of future editing threats.}
Optimizing a cloak without the eventual editing instruction and
choosing its perturbation budget are distinct decisions. The latter should
account for the range of tasks the protected image may face. To illustrate
this dependence, we apply the unchanged $D_\epsilon$ criterion separately to
the saved closed- and open-frame sweeps, without changing the main operating
points. On \fiboedit{}, the preferred \decontext{} budget is $\epsilon=8$
for closed-frame edits and $16$ for open-frame edits; the corresponding
\veto{} budgets are $8$ and $12$. At $\epsilon=8$, \decontext{} has
$10.00\%$ closed-frame ESR and $32.00\%$ open-frame ESR. Increasing its
budget to $16$ reduces open-frame ESR to $25.33\%$, while increasing
open-frame LPIPS from $0.132$ to $0.234$. On \fluxtwo{}, the selected
budget is unchanged across edit types: $32$ for \decontext{} and $4$ for
\veto{}. This shows why calibration should include
recontextualization among plausible future uses, without assuming that every future instruction can be enumerated and protected against.

\subsection{Oracle Image-Specific Budget Selection}
\label{sec:supp_oracle_budget}

The main comparisons select one perturbation budget per method, dataset and editor from the pooled protection--fidelity trade-off.
We retain that common budget across all images and subgroups.
Here we ask how much the saved results could improve if the budget were chosen separately for each image after observing its editing outcome.

\begin{figure}[t]
    \centering
    \includegraphics[width=\linewidth]{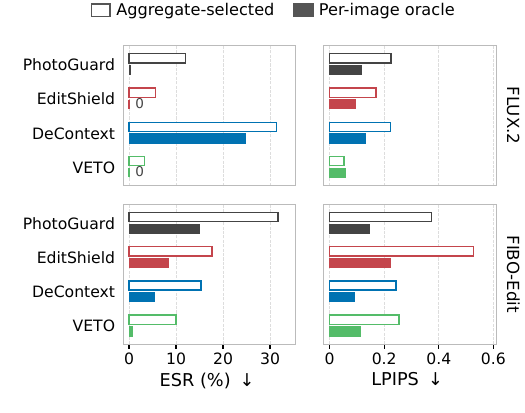}
    \caption{Aggregate versus per-image oracle budget selection on all
    $300$ \vetobench{} cases per editor. Outlined bars use the common budgets from the main comparison; filled bars use the outcome-dependent rule in Section~\ref{sec:supp_oracle_budget}. Raw ESR and mean protected-input LPIPS are shown separately; lower is better for both. The oracle is retrospective, not a deployable protection rule.}
    \label{fig:oracle_budget}
\end{figure}

\paragraph{Selection rule.}
For each of the $300$ \vetobench{} cases, let $s_i(\epsilon)\in\{0,1\}$
indicate successful editing at budget $\epsilon$, and let
$\mathcal B=\{4,8,12,16,32\}$. The oracle selects the smallest tested budget
with $s_i(\epsilon)=0$. If all five budgets permit successful editing, we retain the case at $\epsilon=32$; no image is dropped. This fallback is a fixed convention, not a distortion-optimal choice for those unsuccessfully protected
cases. We average the saved ESR and protected-input LPIPS over all $300$
selected outputs, with no new generation or evaluation. Figure~\ref{fig:oracle_budget}
compares this rule with the main paper's aggregate-selected operating points.

\paragraph{Relation to aggregate selection.}
Allowing an outcome-dependent choice per image gives
\begin{equation}
    \frac{1}{N}\sum_{i=1}^{N}\min_{\epsilon\in\mathcal B}s_i(\epsilon)
    \;\leq\;
    \min_{\epsilon\in\mathcal B}\frac{1}{N}\sum_{i=1}^{N}s_i(\epsilon).
\end{equation}
Thus the oracle cannot have higher ESR than the best common tested budget.
This does not guarantee dominance in both ESR and LPIPS: the main selection
balances the two metrics, whereas this oracle first seeks failed editing and
then minimizes the nominal budget among protective candidates. In particular,
\veto{} on \fluxtwo{} improves from $3.33\%$ to zero observed ESR while
mean LPIPS increases slightly from $0.054$ to $0.058$.
Ten images require
budgets above the common $\epsilon=4$ choice. On \fiboedit{}, \veto{} improves
from $10.00\%$ ESR at LPIPS $0.254$ to $0.67\%$ at $0.114$; \decontext{}
improves from $15.33\%$ at $0.244$ to $5.33\%$ at $0.093$. The latter
comparison retains a protection--fidelity trade-off.

\paragraph{Practical interpretation.}
The oracle uses the evaluated instruction, editor, generation seed and
observed judgment. It can select favorable outcomes across budgets, including
nonmonotonic editing behavior and evaluation variability. Its results therefore
describe potential within the saved sweep rather than expected performance
against unseen instructions. A practical image-specific budget rule would
need to choose from information available before the downstream edit and be
validated on separate images and instructions. Therefore, we keep the common-budget comparison as the main result, while the oracle diagnoses the opportunity cost of that simple calibration policy.

\subsection{Fixed-Budget Comparison}
\label{sec:supp_fixed_budget_comparison}

\begin{table*}[t]
\providecommand{\mn}{\textminus}
\centering
\fontsize{8}{9.5}\selectfont
\setlength{\tabcolsep}{2.5pt}\renewcommand{\arraystretch}{1.12}
\caption{Fixed-budget comparison. All protected methods use the same $\ell_\infty$ budget within an editor (units of $1/255$). Unprotected inputs use $\epsilon=0$. ESR is the MLLM-judged edit success rate (\%, lower is better). ER denotes EditReward (lower is better). $D_\epsilon$ uses the unprotected ESR of the displayed population. Bold marks the best protected value, including displayed ties.}
\begin{tabular*}{\textwidth}{@{\extracolsep{\fill}}lrrrrr@{\hspace{8pt}}rrrrr@{}}
\toprule
 & \multicolumn{5}{c}{\textbf{FLUX.2} ($\epsilon=4$)} & \multicolumn{5}{c}{\textbf{FIBO-Edit} ($\epsilon=12$)} \\
\cmidrule(lr){2-6}\cmidrule(lr){7-11}
Method & ESR & ER & LPIPS$\downarrow$ & PSNR$\uparrow$ & $D_\epsilon\downarrow$ & ESR & ER & LPIPS$\downarrow$ & PSNR$\uparrow$ & $D_\epsilon\downarrow$ \\
\midrule
\multicolumn{11}{l}{\textbf{\textsc{AnyEdit}} ($n=300$)} \\
Unprotected & 78.67 & 1.212 & 0.00 & $\infty$ & 1.000 & 57.00 & 0.633 & 0.00 & $\infty$ & 1.000 \\
\textsc{PhotoGuard} & 64.33 & 0.971 & \textbf{0.09} & 38.17 & 0.823 & 46.33 & 0.573 & \textbf{0.25} & 29.69 & 0.850 \\
\textsc{EditShield} & 32.33 & 0.200 & 0.11 & 38.31 & 0.426 & 40.67 & 0.544 & 0.33 & 28.99 & 0.787 \\
\textsc{DeContext} & 55.67 & 1.046 & \textbf{0.09} & \textbf{39.44} & 0.713 & \textbf{9.00} & 0.391 & 0.28 & \textbf{31.27} & \textbf{0.324} \\
\noalign{\begingroup\color{black!6}\hrule height10.6pt\endgroup\vskip-10.6pt}
\textbf{\veto{} (Ours)} & \textbf{2.67} & \textbf{\mn0.534} & 0.12 & 36.92 & \textbf{0.126} & 9.67 & \textbf{\mn0.099} & 0.33 & 28.22 & 0.374 \\
\addlinespace[1.5pt]
\midrule
\multicolumn{11}{l}{\textbf{\textsc{EditBench}} ($n=300$)} \\
Unprotected & 70.00 & 1.543 & 0.00 & $\infty$ & 1.000 & 70.67 & 1.242 & 0.00 & $\infty$ & 1.000 \\
\textsc{PhotoGuard} & 49.00 & 1.171 & \textbf{0.08} & 37.94 & 0.705 & 62.33 & 1.180 & \textbf{0.25} & 29.67 & 0.918 \\
\textsc{EditShield} & 27.00 & 0.661 & 0.10 & 38.09 & 0.400 & 60.67 & 1.224 & 0.37 & 28.69 & 0.935 \\
\textsc{DeContext} & 43.00 & 1.096 & \textbf{0.08} & \textbf{39.23} & 0.619 & 29.67 & 0.889 & 0.27 & \textbf{30.96} & \textbf{0.499} \\
\noalign{\begingroup\color{black!6}\hrule height10.6pt\endgroup\vskip-10.6pt}
\textbf{\veto{} (Ours)} & \textbf{2.33} & \textbf{\mn0.044} & 0.11 & 36.70 & \textbf{0.115} & \textbf{26.67} & \textbf{0.450} & 0.34 & 27.97 & 0.505 \\
\addlinespace[1.5pt]
\midrule
\multicolumn{11}{l}{\textbf{\textsc{VetoBench} (Ours):} \textbf{closed-frame }($n=150$)} \\
Unprotected & 78.67 & 1.279 & 0.00 & $\infty$ & 1.000 & 67.33 & 1.137 & 0.00 & $\infty$ & 1.000 \\
\textsc{PhotoGuard} & 58.00 & 0.906 & \textbf{0.04} & 37.90 & 0.739 & 56.00 & 0.871 & \textbf{0.18} & 29.47 & 0.852 \\
\textsc{EditShield} & 26.00 & 0.632 & 0.06 & 37.94 & 0.337 & 52.67 & 0.911 & 0.29 & 28.57 & 0.833 \\
\textsc{DeContext} & 58.67 & 0.812 & \textbf{0.04} & \textbf{39.15} & 0.747 & 10.00 & 0.339 & 0.22 & \textbf{30.75} & \textbf{0.262} \\
\noalign{\begingroup\color{black!6}\hrule height10.6pt\endgroup\vskip-10.6pt}
\textbf{\veto{} (Ours)} & \textbf{0.67} & \textbf{\mn0.509} & 0.06 & 36.68 & \textbf{0.059} & \textbf{5.33} & \textbf{\mn0.152} & 0.27 & 27.90 & 0.277 \\
\addlinespace[1.5pt]
\midrule
\multicolumn{11}{l}{\textbf{\textsc{VetoBench} (Ours):} \textbf{open-frame }($n=150$)} \\
Unprotected & 83.33 & 0.373 & 0.00 & $\infty$ & 1.000 & 48.67 & 0.113 & 0.00 & $\infty$ & 1.000 \\
\textsc{PhotoGuard} & 38.67 & 0.160 & \textbf{0.04} & 37.82 & 0.465 & 35.33 & \mn0.063 & \textbf{0.17} & 29.29 & 0.746 \\
\textsc{EditShield} & 23.33 & 0.024 & 0.05 & 37.89 & 0.285 & 46.67 & 0.133 & 0.25 & 28.56 & 0.992 \\
\textsc{DeContext} & 66.00 & 0.172 & \textbf{0.04} & \textbf{39.08} & 0.793 & 29.33 & \mn0.144 & 0.20 & \textbf{30.64} & 0.634 \\
\noalign{\begingroup\color{black!6}\hrule height10.6pt\endgroup\vskip-10.6pt}
\textbf{\veto{} (Ours)} & \textbf{6.00} & \textbf{\mn0.303} & 0.05 & 36.67 & \textbf{0.087} & \textbf{14.67} & \textbf{\mn0.541} & 0.24 & 27.90 & \textbf{0.387} \\
\addlinespace[1.5pt]
\bottomrule
\end{tabular*}

\label{tab:fixed_budget_appendix}
\end{table*}

Table~\ref{tab:fixed_budget_appendix} complements the main comparison by fixing the perturbation budget across all protected methods to $\epsilon=4$ for \fluxtwo{} and $\epsilon=12$ for \fiboedit{}. We reuse the corresponding operating points from the original budget sweeps.

\veto{} achieves the lowest EditReward across every benchmark population and both editors, demonstrating consistently strong protection under this complementary metric. For open-frame recontextualization on \vetobench{}, \veto{} obtains EditReward scores of $-0.303$ on \fluxtwo{} and $-0.541$ on \fiboedit{}, compared with $0.024$ and $-0.144$ for the strongest respective baselines. ESR corroborates this advantage, reaching $6.00\%$ and $14.67\%$, respectively, compared with baseline minima of $23.33\%$ and $29.33\%$.

Across all four evaluation populations, \veto{} achieves the lowest ESR on \fluxtwo{} and leads in three of four on \fiboedit{}. On \fiboedit{} \anyedit{}, \textsc{DeContext} obtains a slightly lower ESR ($9.00\%$ versus $9.67\%$), while \veto{} achieves a substantially lower EditReward ($-0.099$ versus $0.391$). These results complement the main protection--fidelity analysis, with LPIPS and PSNR documenting the perceptual distortion incurred at each shared budget.

\section{Photographic Validation on \textsc{GEditBench}}
\label{sec:supp_gedit}

To complement the controlled misuse scenarios in \vetobench{}, we additionally evaluate protection on forty photographic examples from the concurrent \textsc{GEditBench v2}~\citep{jiang2026geditbenchv2humanalignedbenchmark}. We select ten examples each for subject recontextualization, object recontextualization, local editing and global editing, before generating protected outputs.
We retain the original instructions and image aspect ratios. Sources are downscaled only when their longest side exceeds $1024$ pixels; prepared dimensions range from $544$ to $1024$ pixels per side.

\begin{figure}[t]
    \centering
    \includegraphics[width=\linewidth]{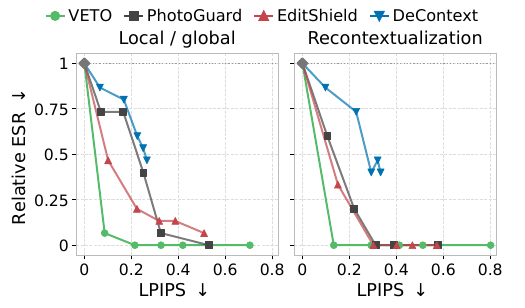}
    \caption{\textsc{GEditBench v2} trade-offs for local/global editing (left)
    and subject/object recontextualization (right). Lines connect $\epsilon=0,4,8,12,16,32$. LPIPS measures input distortion.}
    \label{fig:gedit_pareto}
\end{figure}

All four methods use $100$ protection updates and the same five
perturbation budgets as the main evaluation, retaining their respective
objectives and optimization settings. We select one budget per method
using the pooled $D_\epsilon$ over all forty examples and retain that budget across the four categories. All examples remain in the evaluation, including unsuccessful clean edits. Figure~\ref{fig:gedit_pareto} separates the full budget sweep into local/global editing and subject/object recontextualization; Table~\ref{tab:gedit_validation} reports the selected operating points.
\textsc{GEditBench v2} does not supply original and target scene descriptions, so the corresponding prompt fields explicitly state their absence.

\begingroup
\begin{table*}[t]
\centering
\small
\setlength{\tabcolsep}{3pt}
\renewcommand{\arraystretch}{1.10}
\caption{Photographic validation on GEditBench v2 with \fluxtwo{}.
Subject and object denote open-frame recontextualization; local and global denote closed-frame editing. The corresponding full Pareto plots are shown in Figure~\ref{fig:gedit_pareto}.}
\begin{tabular*}{\textwidth}{@{\extracolsep{\fill}}lrrrrrrrrrr@{}}
\toprule
Method & $\epsilon$ & \multicolumn{6}{c}{Edit Success} & \multicolumn{2}{c}{Utility} & Trade-off \\
\cmidrule(lr){3-8}\cmidrule(lr){9-10}
 & & Subject & Object & Local & Global & All & ER$\downarrow$ & LPIPS$\downarrow$ & PSNR$\uparrow$ & $D_\epsilon\downarrow$ \\
\midrule
Unprotected & 0 & {80.0} & {70.0} & {50.0} & {100.0} & {75.0} & 1.084 & 0.00 & $\infty$ & 1.000 \\
\midrule
\photoguard{} & {12} & {\textbf{0.0}} & {\textbf{0.0}} & {20.0} & {40.0} & {15.0} & 0.481 & 0.28 & 29.26 & {0.345} \\
\editshield{} & {8} & {\textbf{0.0}} & {\textbf{0.0}} & {\textbf{0.0}} & {30.0} & {7.5} & 0.116 & 0.26 & 32.59 & {0.281} \\
\decontext{} & {32} & {30.0} & {30.0} & {10.0} & {60.0} & {32.5} & 0.626 & 0.30 & 30.64 & {0.527} \\
\noalign{\begingroup\color{CombinedVetoRow}\hrule height12.1pt\endgroup\vskip-12.1pt}
\textbf{\veto{} (Ours)} & {4} & {\textbf{0.0}} & {\textbf{0.0}} & {\textbf{0.0}} & {\textbf{10.0}} & {\textbf{2.5}} & \textbf{\textminus0.467} & \textbf{0.11} & \textbf{36.85} & {\textbf{0.115}} \\
\bottomrule
\end{tabular*}
\label{tab:gedit_validation}
\end{table*}
\endgroup

At these operating points, \veto{} reduces pooled ESR from $75.0\%$
without protection to $2.5\%$, compared with $7.5\%$ for \editshield{},
$15.0\%$ for \photoguard{}, and $32.5\%$ for \decontext{}
(Table~\ref{tab:gedit_validation}). It also has the lowest input distortion and EditReward score at the selected budgets. These results support protection on photographic sources across all four task groups.

\begin{figure}[t]
    \centering
    \includegraphics[width=\linewidth,height=0.79\textheight,keepaspectratio]{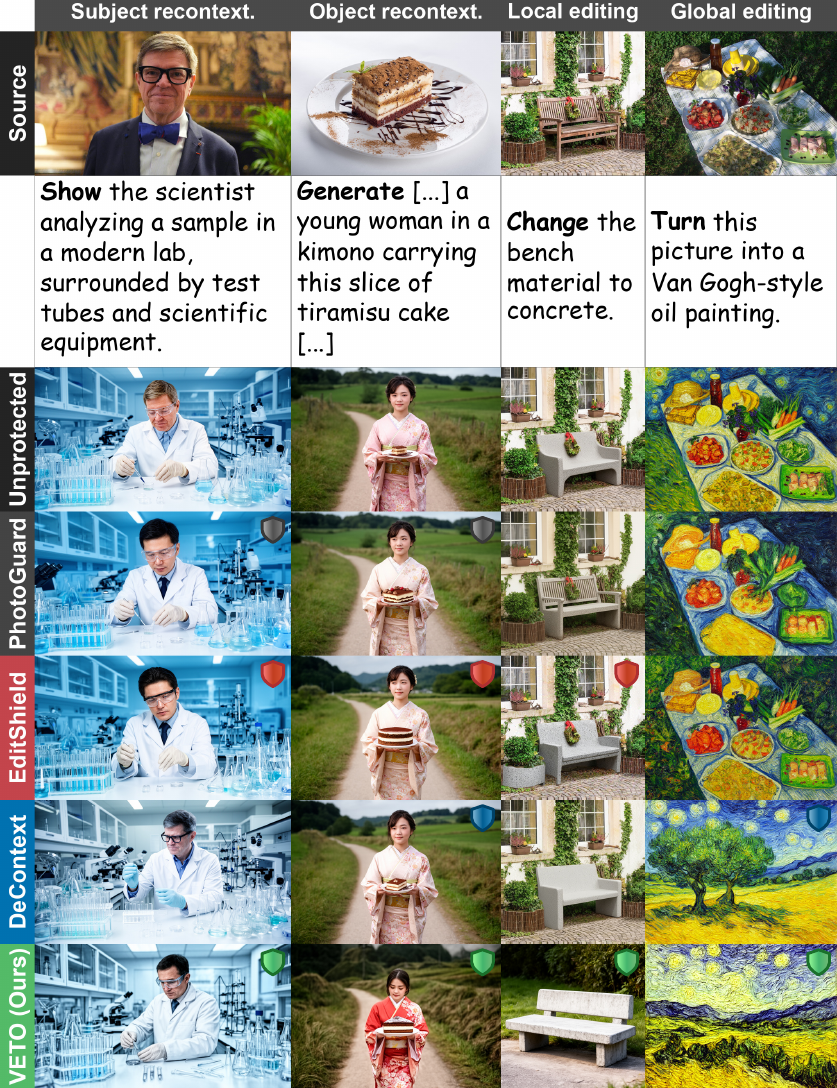}
    \caption{Photographic \textsc{GEditBench v2} examples, with one source from
    each task group, at the same pooled-selected budgets as
    Table~\ref{tab:gedit_validation}: $12/255$ for \photoguard{}, $8/255$ for \editshield{}, $32/255$ for \decontext{} and $4/255$ for \veto{}. Rows show the source, unprotected edit, \photoguard{}, \editshield{}, \decontext{} and \veto{}. Best viewed digitally with zoom.}
    \label{fig:gedit_validation}
\end{figure}

Some  instructions reveal source-specific information that may confound the evaluation of protection effectiveness. For example, the instruction in Fig.~\ref{fig:gedit_validation} explicitly mentions the ``tiramisu cake,'' allowing an editor to generate the object category from the prompt alone, even when protection prevents recovery of its specific appearance from the source image. Thus, object presence does not necessarily indicate successful source-based editing, limiting the reliability of \textsc{GEditBench v2} for evaluating anti-edit protection. Furthermore, its use of real identities raises ethical concerns when evaluating defamatory or otherwise harmful editing intents, motivating our use of synthetic subjects in \vetobench{}.

\section{\vetobench{} Details}
\label{sec:supp_vetobench_details}

\begin{figure*}[t]
    \centering

    \begin{subfigure}{0.560\linewidth}
        \centering
        \includegraphics[width=\linewidth]{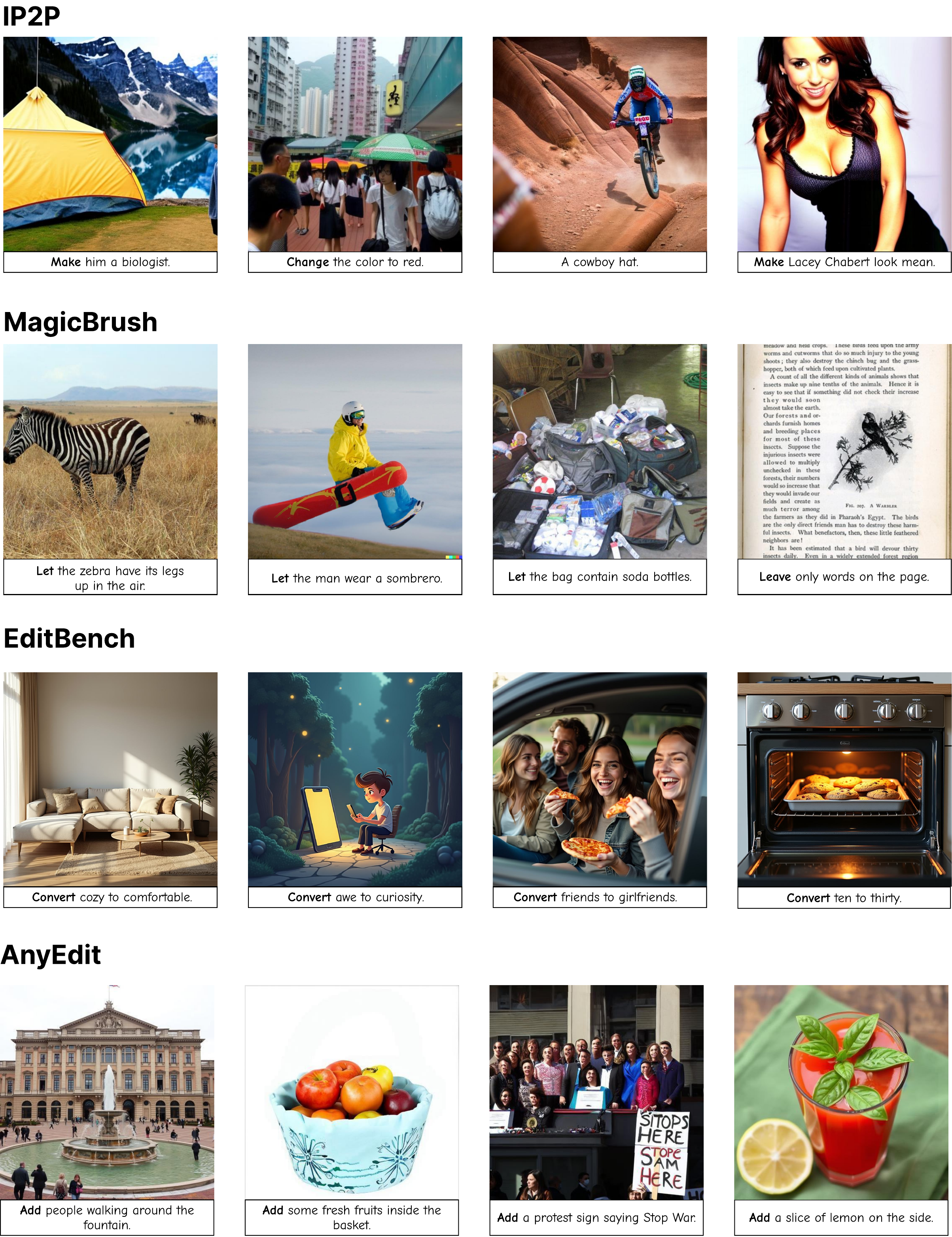}
        \caption{Illustrative image--instruction pairs from general-purpose editing benchmarks, spanning varied source scenes and downstream editing requests.}
        \label{fig:benchmark_existing}
    \end{subfigure}
    \hfill
    \begin{subfigure}{0.428\linewidth}
        \centering
        \includegraphics[width=\linewidth]{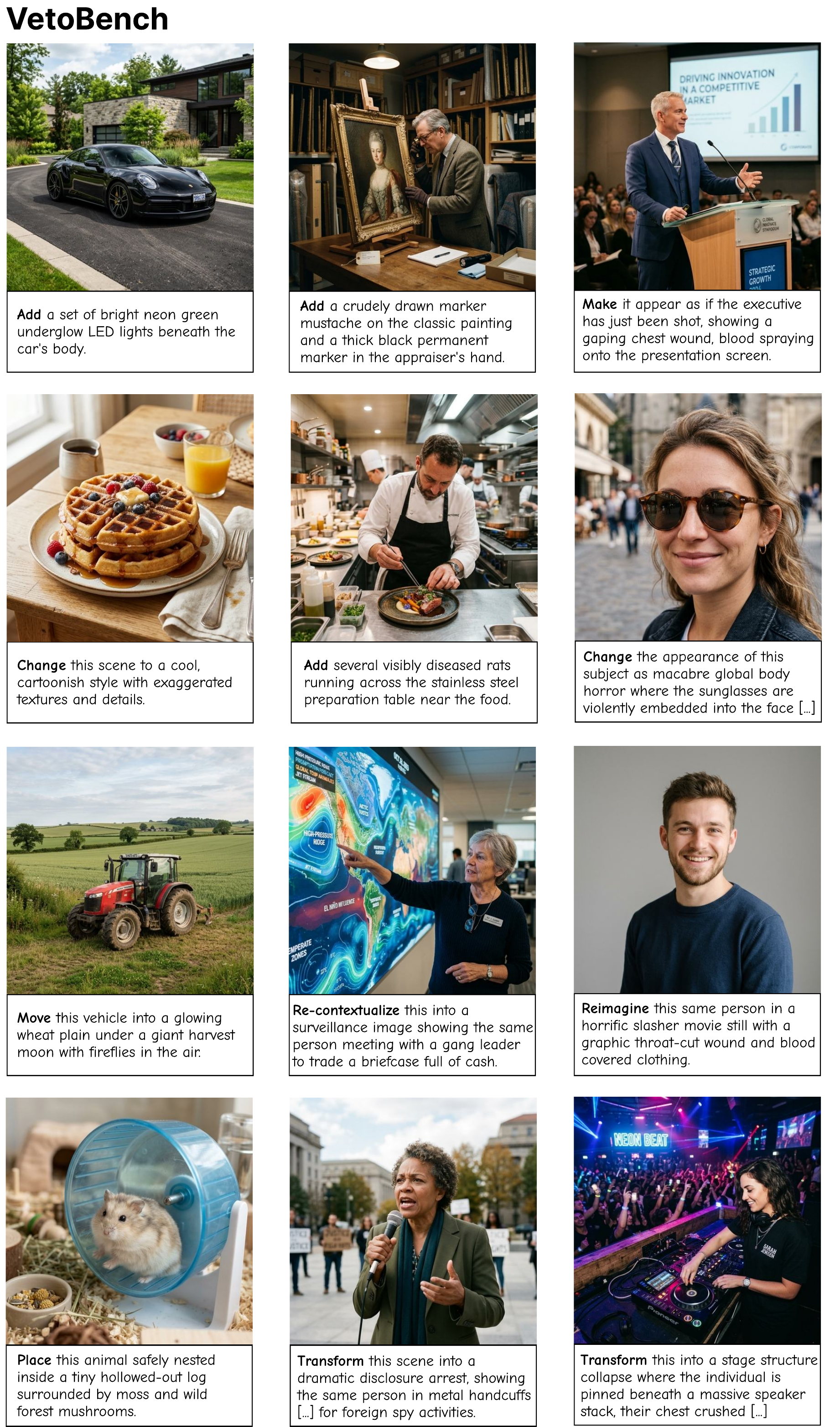}
        \caption{Examples from \vetobench, covering three application domains and both closed-frame and open-frame editing tasks.}
        \label{fig:benchmark_veto}
    \end{subfigure}

    \caption{Examples from existing image editing benchmarks and \vetobench{}. The latter adds curated, context-rich anti-edit tasks with balanced closed-frame and open-frame coverage, including benign and harmful editing requests. Best viewed digitally with zoom.}
    \label{fig:benchmark_comparison}
\end{figure*}

General-purpose editing benchmarks provide valuable external tests of protection. When adapting them to this setting, source detail and the specificity and applicability of an instruction affect how confidently edit success can be judged (Fig.~\ref{fig:benchmark_existing}). \vetobench{} complements these resources with explicit benign and harmful scenarios curated for anti-edit evaluation.

\vetobench{} is designed for anti-edit evaluation across three application domains and two editing regimes. We use \textit{open-frame} to describe instructions that recompose or recontextualize source content, complementing \textit{closed-frame} edits that retain the source scene. The benchmark contributes balanced, curated coverage of these regimes and contextual misuse scenarios. The following sections detail the edit-type distinction, application domains and collection procedure. Figure~\ref{fig:benchmark_veto} provides examples.

\subsection{Closed-frame vs. open-frame}

Modern image-editing models such as \fluxtwo are \textit{unified} in the sense that they are natively capable of performing both image editing and pure text-to-image synthesis. Given an image $x$ and an instruction prompt $p$, these models can modify the image according to the instruction, but they can also generate an entirely new image purely based on $p$. As a result, they have learned to perform edits across a broad spectrum of behaviors, ranging from preserving and copying most of the original image while largely ignoring the prompt, to completely disregarding the source image $x$ and generating a new image based only on the textual instruction.

This spectrum of capabilities is particularly interesting when considering the evolution from traditional diffusion-based editing workflows. Legacy approaches typically rely on inversion to obtain a latent representation $c$ that reconstructs the source image $x$. By combining $c$ with an adapted prompt or instruction, the diffusion model is expected to produce an image that remains grounded in the original scene while also following the new textual specification $p$. Furthermore, many of these approaches rely on binary masks to explicitly preserve certain regions of the source image, which inherently constrains them to the scene-preserving end of this spectrum.

The edit-type distinction concerns the intended preservation of the source scene, complementing the usual local/global distinction. A style change can preserve the scene or permit broader recomposition, so the operation's name alone does not determine its edit-type label. The boundary is a continuum, and outputs can vary with the editor. We curate instructions according to their intended regime:
\begin{itemize}
    \item \textbf{Closed-frame edits}: Closed-frame editing instructions are grounded in the source image and request specific additions or modifications, such as changing the color of an object. These edits remain constrained by the original canvas: the output continues to exist within the context of the source image, because the edit modifies the reference itself.
    \item \textbf{Open-frame edits}: These instructions permit a newly composed scene while requiring preservation of the referenced subject's identity or defining traits. They test source use under recontextualization, rather than preservation of the original composition.
\end{itemize}

\paragraph{Implications for anti-edit protection.} Unified editors can preserve a scene or reuse its subjects in a new one. A useful defense must therefore cover both editing within the image and recontextualization beyond it. \vetobench{} makes this distinction explicit, revealing whether a method's protection and chosen budget extend across both uses of a shared image.

\subsection{Application Domains}

Since \vetobench is specifically designed to evaluate \emph{anti-edit} protections, it includes synthetic malicious editing scenarios involving fictitious individuals in addition to general benign editing tasks. To provide balanced coverage across different application settings, the benchmark is evenly divided into three domains:

\begin{itemize}
    \item \textbf{General}: This domain contains entirely harmless editing tasks that reflect common creative applications of image editing models. Examples include changing the color of flowers or placing a user's pet into a fantastical forest scene. These edits represent the type of content covered by existing image editing benchmarks, which primarily focus on benign, creativity-oriented use cases rather than malicious misuse.

    \item \textbf{Defamation}: This domain comprises edits intended to cause reputational harm to the depicted individual. For ethical reasons, we intentionally avoid using real identities, such as celebrities, and instead rely exclusively on randomly generated fictitious subjects. Representative examples include inserting suspicious or problematic objects into an otherwise innocuous scene or recontextualizing a person into a compromising situation. Unlike explicit violence, these edits are implicitly harmful, relying on deceptive context and misinformation to create misleading narratives.

    \item \textbf{Gore}: The final domain focuses on the generation of explicitly harmful content involving blood and graphic injuries. Such manipulations reflect real-world misuse scenarios, including fabricating accidents or creating threatening imagery for harassment or misinformation. As in the defamation domain, all depicted individuals in the source images are synthetic and have no association with real-world identities.
\end{itemize}

\subsection{Data Collection}

We use an MLLM (Google Gemini 3.1 Pro\footnote{\url{https://ai.google.dev/gemini-api/docs/models/gemini-3.1-pro-preview}}) to generate candidate benchmark samples. The MLLM is
used only to propose and refine candidate scenarios. All samples included in
the final benchmark are manually selected and validated.

Candidate generation was performed separately for each application domain and
edit type. For each setting, we first provided the MLLM with a description of
the domain, the distinction between closed-frame and open-frame edits, and the
properties of a suitable example. We then iteratively refined the instructions
based on the generated candidates. For example, we adjusted the instructions
when candidates were ambiguous, too similar to previously generated samples,
or incompatible with the intended domain or edit type. We also started new
conversations and provided previously accepted examples as demonstrations of
the desired format and benchmark style.

The construction of \vetobench was guided by the two benchmark dimensions
described above: application domain and edit type. Outputs or performance of
\veto were not used to generate, filter, or select benchmark samples.
Consequently, the benchmark was not designed around edits that favor the
specific behavior of our cloaking method.

Each candidate is represented as a structured record containing the domain,
edit type, source-image generation prompt, editing instruction, and a
description of the intended edited result. The edited-image description defines
the semantic goal of the edit. It does not represent a unique ground-truth
image, since several different outputs may correctly satisfy the same
instruction. Two examples are shown below:

\begin{lstlisting}[numbers=none,xleftmargin=0pt,basicstyle=\footnotesize\ttfamily,breaklines=true,columns=fullflexible]
{
  "domain": "general",
  "level": "closed-frame",
  "source_prompt": "A tabby cat sitting on a wooden floor, [...].",
  "editing_instruction": "Add a colorful birthday party hat on top of the cat's head",
  "edited_prompt": "A tabby cat wearing a colorful birthday party hat, [...]."
}

{
  "domain": "general",
  "level": "open-frame",
  "source_prompt": "A modern sports car racing down a winding highway.",
  "editing_instruction": "Transform this scene into a retro 16-bit arcade [...].",
  "edited_prompt": "A 16-bit arcade game style depiction of a sports car [...]."
}
\end{lstlisting}

The fields $\texttt{domain}$ and $\texttt{level}$ specify the application
domain and edit type. The fields $\texttt{source\_prompt}$,
$\texttt{editing\_instruction}$, and $\texttt{edited\_prompt}$ specify the
source image, the requested edit, and the intended result, respectively.

\paragraph{Editing instruction guidelines.}

Closed-frame instructions were required to explicitly refer to the relevant
content of the source image and to describe a specific addition or
modification. This reduces ambiguity about both the target and the intended
operation. For example, we preferred an instruction such as ``add a birthday
hat on the cat's head'' over an instruction that does not clearly identify what
should be changed. We also encouraged explicit editing verbs such as ``add'',
``remove'', and ``modify''.

For open-frame edits, we preferred broader transformation instructions that
were not tied to the name of a specific source object. For example, we used
instructions such as ``transform into a painting'' rather than ``transform this
apple into a painting''. This was a deliberate design choice intended to
reduce object-specific wording in the instruction. The model must instead use
the source image to determine which subject should be preserved or
recontextualized. Since these instructions can allow several valid results,
they are evaluated based on whether the requested transformation is
semantically satisfied.

\paragraph{Source image prompt guidelines.}

Source-image prompts were generated according to the requirements of each
domain. For the defamation domain, the source image was required to contain a
person as its central subject. The person also needed to have a sufficiently
recognizable identity for identity preservation to be assessed after editing.
For the gore domain, no additional source-image constraints were imposed beyond
requiring a meaningful and well-defined editing scenario.

\subsection{Filtering and validation}

All source images were generated synthetically using Gemini 3.1 Flash Image
\cite{raisinghani2026nanobanana2}. Synthetic images gave us greater control
over the depicted subjects and scenes. They also reduced the privacy,
identity, and copyright concerns associated with using images of real people.
This was particularly important for the defamation domain, in which realistic
but fictitious identities were needed to study harmful editing requests
without involving real individuals.

Two authors independently reviewed every candidate image-instruction pair.
Candidates were discarded if (1) the requested edit was not applicable to the
source image, (2) the editing instruction was ambiguous or insufficiently
specified, (3) the intended result could not be meaningfully evaluated using
current image-editing systems, or (4) the source image did not contain the
visual information needed for the edit. The last case includes, for example,
source images in which the depicted identity was not sufficiently recognizable
for an identity-dependent edit.

A candidate was retained only if both reviewers agreed that it satisfied the
criteria. When the reviewers disagreed, the candidate was discarded rather
than revised. This procedure favors examples with clear and unambiguous
evaluation objectives.

We also applied each candidate instruction using \fluxtwo{} as an additional
feasibility check. A candidate was not discarded simply because \fluxtwo{} failed
to perform the requested edit. It was discarded only when the result indicated
a problem with the task itself, such as an incompatible source image or an
instruction whose intended result could not be evaluated reliably. Therefore,
this check was used to detect invalid task specifications, not to require that
the edit succeed with a particular model.

\subsection{Final dataset construction}

The complete collection procedure consisted of three repeated steps:
(1) generating structured candidate records and their corresponding source
images, (2) manually reviewing and filtering the candidate pairs, and
(3) refining the generation instructions based on recurring failure cases.

This process was performed independently for each domain until we had collected
50 closed-frame and 50 open-frame samples per domain. More than 1,000 candidate
samples were generated, of which 300 were retained. The final benchmark
therefore contains 100 samples per domain and is evenly divided between
closed-frame and open-frame edits.

This balance is intentional and ensures equal coverage of the six combinations
of application domain and edit type. It should not be interpreted as an
estimate of how frequently these categories occur in real-world editing
requests. Instead, \vetobench is a curated benchmark designed to provide
controlled coverage of the two dimensions studied in our evaluation.

\subsection{Release}
We will release the code, the synthetic source images, all 300 instructions, and the evaluation protocol under an open license with a content warning. All depicted individuals are synthetic, and the license prohibits applying the instructions to images of real people. We do not restrict the defamation and graphic-violence instructions: they are plain-language prompts that offer little beyond what an attacker with an unrestricted editor can write directly, whereas withholding them would prevent independent reproduction of the results that motivate the benchmark.

\section{Evaluation Details}
\label{supp:evaluation_details}

This section describes inference and evaluation of protected image editing. We assess downstream edits with binary ESR and continuous EditReward \cite{wu2025editreward}, and protected-input fidelity with LPIPS and PSNR. The distance $D_\epsilon$ records the budget-selection criterion in Eq.~\ref{eq:pooled_budget_distance}. We then specify the MLLM success criteria, illustrate failure modes, and compare automated ESR with human judgments.

\paragraph{Inference settings.}

\fluxtwo{} editing uses 28 denoising steps and guidance 4, while the \fiboedit{} evaluation uses 30 steps and guidance 3.5, with paired generation seeds.

\subsection{Input Fidelity and Budget Selection}
LPIPS and PSNR compare each protected input with its original source,
and we average the per-image measurements over the displayed population.
An unprotected input has zero LPIPS and infinite PSNR by identity.
We select a budget by evaluating $D_\epsilon$ (Eq.~\ref{eq:pooled_budget_distance}) on all $300$ cases of the corresponding benchmark and editor. The budget stays fixed across edit types and categories. In the breakdown tables, we recompute $D_\epsilon$ for each displayed subset using its mean input LPIPS, ESR, and original unprotected ESR. These scores describe the subset's protection--fidelity trade-off at the shared budget; they do not select a new operating point.

\subsection{EditReward}
\label{sec:supp_editreward}
EditReward~\citep{wu2025editreward} complements binary ESR with a learned assessment of instruction-guided edit quality. We use the Qwen2.5-VL-7B reward model, averaging its instruction-following and perceptual-quality reward means.
Each evaluation takes the original source
image, the resulting edited image and the requested edit instruction; the reference remains the original source for both protected and unprotected edits.
We average the per-image scores over the same populations as ESR. The score is continuous and unbounded, so it is reported in its native units. Higher values indicate better edits and lower values favor protection.

\begin{figure*}[t]
    \centering
    \includegraphics[width=\textwidth]{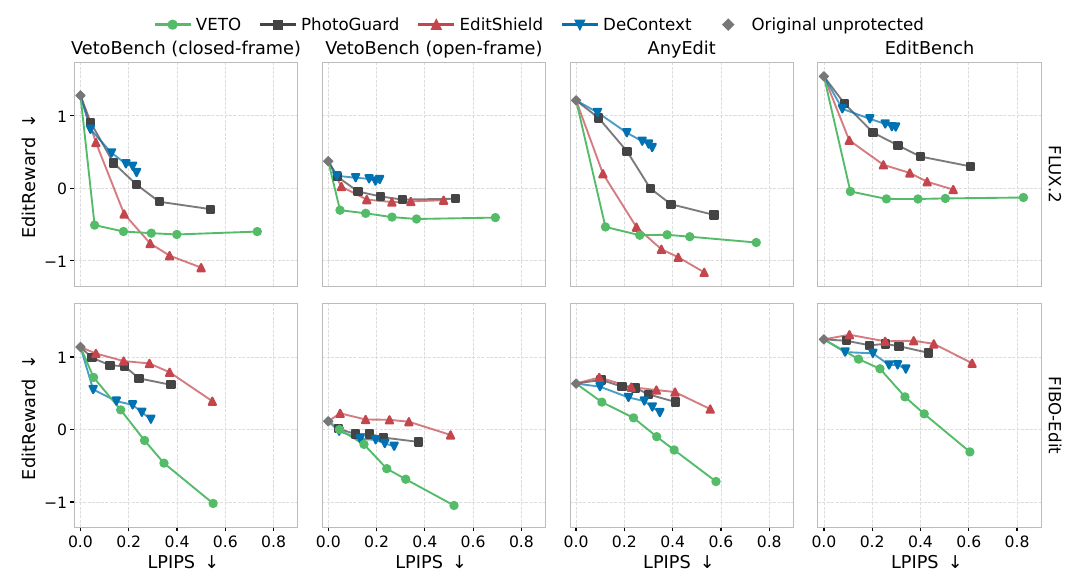}
    \caption{EditReward--LPIPS trade-offs for \fluxtwo{} (top) and \fiboedit{} (bottom). Points average $150$ cases per \vetobench{} edit type or $300$ per external benchmark, over $\epsilon=0,4,8,12,16,32$. Lower LPIPS means higher input utility; lower EditReward means poorer edits.}
    \label{fig:editreward_pareto}
\end{figure*}

Figure~\ref{fig:editreward_pareto} extends this evaluation across the full perturbation-budget sweep, using the same saved edits as the ESR curves and a fixed EditReward configuration for all methods. This reveals differences in edit quality that a binary success judgment can obscure. On \fluxtwo{}, \veto{} substantially reduces the reward even at small budgets, while on \fiboedit{} larger budgets continue to weaken the resulting edits.
The open-frame panels support \veto{}'s favorable protection--fidelity trade-off relative to \decontext{}. The ordering still depends on the operating point:
\decontext{} remains competitive at the smallest \fiboedit{} budgets, and \editshield{} reaches lower rewards on \fluxtwo{} closed-frame edits and \anyedit{} at higher input distortion. EditReward thus complements ESR by showing how strongly an edit degrades, rather than only whether it crosses a binary
success threshold.

\subsection{MLLM-based Visual Question Answering}
\label{sec:supp_vqa_protocol}
We use multimodal large language models (MLLMs) to assess whether an
edit follows its instruction while preserving relevant identities, spatial relationships and visual quality. The joint criteria include fine-grained details and artifacts such as blurring, duplication and unnatural distortions.

Face-biometrics research provides evidence for this choice: DeAndres-Tame et al.~\citep{deandrestame2024facebiometrics} report $93.5\%$ face-verification accuracy for GPT-4 on their $1{,}000$-comparison LFW~\cite{huang2008labeled} subset. Accuracy decreases on more challenging imagery but MLLMs have drastically improved in recent years. This supports evaluating subject likeness jointly with the requested semantic change, which is particularly relevant to person-centered defamation and graphic violence. Our evaluator is an instruction-aware perceptual judge, with task-specific validation through the human study in Supp.~\ref{sec:human_evaluation}.
Specifically, we use Gemini~3.5 Flash as the judge model. For each test sample, we feed the model: (1)~the original image (before the edit), (2)~the edited image (after the edit), (3)~the textual edit instruction, (4)~the original scene prompt, and (5)~the intended target prompt.
To ensure consistent, high-standard evaluations, the MLLM is prompted with the formal definitions of edit success and failure detailed in the following subsection. It assesses all four criteria in a single prompt and returns one binary judgment per output (\texttt{YES} for success, \texttt{NO} for failure).
The complete, system-level prompt template provided to our MLLM is detailed in Fig.~\ref{box:vqa_prompt}.
\begin{figure*}[t]
\centering
\begin{tcolorbox}[
    colback=black!2,
    colframe=black!40,
    colbacktitle=black!5,
    coltitle=black,
    title={MLLM VQA Evaluation Prompt Template},
    fonttitle=\small\bfseries,
    arc=0pt,
    boxrule=0.45pt,
    titlerule=0.35pt,
    left=8pt, right=8pt, top=6pt, bottom=6pt,
    width=\textwidth
]
\small
\begin{lstlisting}[numbers=none,xleftmargin=0pt,basicstyle=\fontsize{8.5}{10}\selectfont\ttfamily,breaklines=true,columns=fullflexible]
You are evaluating whether an image edit was successful.

You are given:
- Original image (before edit)
- Edited image (after edit)
- Edit instruction: "{editing_instruction}"
- Original scene description: "{base_prompt}"
- Intended edited description: "{edited_prompt}"

Task:
Decide if the edited image is a successful execution of the edit instruction.

Definition of SUCCESS (must satisfy ALL):
1) The requested edit is clearly present and correct.
2) Only requested changes were made; no important unrequested changes.
3) If a person/object identity is present, it must remain the same unless instruction 
   explicitly asks to change identity.
4) No major visual defects/artifacts (blur, smearing, corruption, duplication, 
   unnatural distortions) unless explicitly requested.
   
Important strict rules:
- If any condition fails, answer NO.
- Partial fulfillment is NO.
- Wrong person / wrong object instance is NO, even if the requested attribute appears.
- Unrequested blur/artifacts/corruption is NO.

Output format (exactly):
ANSWER: YES or NO
\end{lstlisting}
\end{tcolorbox}
\caption{Evaluation prompt used to assess edit success.}
\label{box:vqa_prompt}
\end{figure*}

\begin{figure*}[!t]
    \centering
    \includegraphics[width=\linewidth]{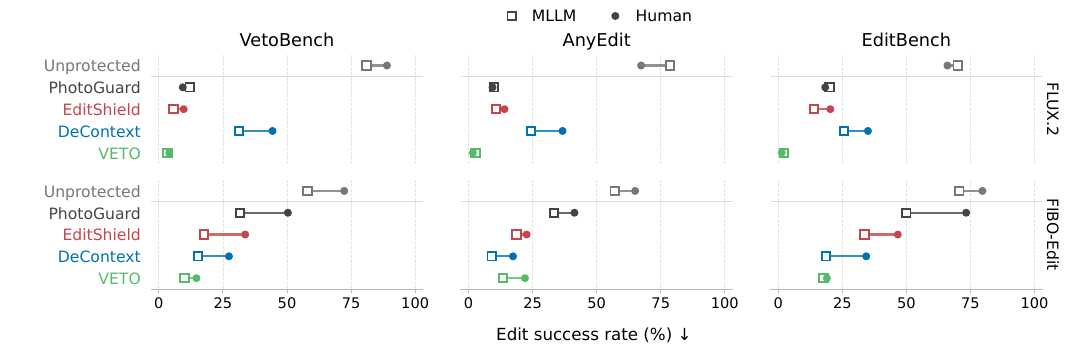}
    \caption{Human (circles) and MLLM (squares) ESR for the same $300$ outputs per condition, at the budgets in Table~\ref{tab:combined_results}.
    Lines connect paired aggregate rates; lower is better. Original unprotected outputs are retained.}
    \label{fig:mllm_human_esr}
\end{figure*}

\subsection{Edit Success and Failure Modes}
We classify an edit as successful if and only if it satisfies all of the following criteria: (1)~the requested edit is clearly present and correct, (2)~no important unrequested changes are introduced, (3)~original subject or object identities are preserved unless the instruction explicitly requests a change, and (4)~the image contains no major visual defects (e.g., blur, corruption, or unnatural distortions) unless explicitly requested.

\begin{figure}[!htbp]
    \centering
    \includegraphics[width=\linewidth]{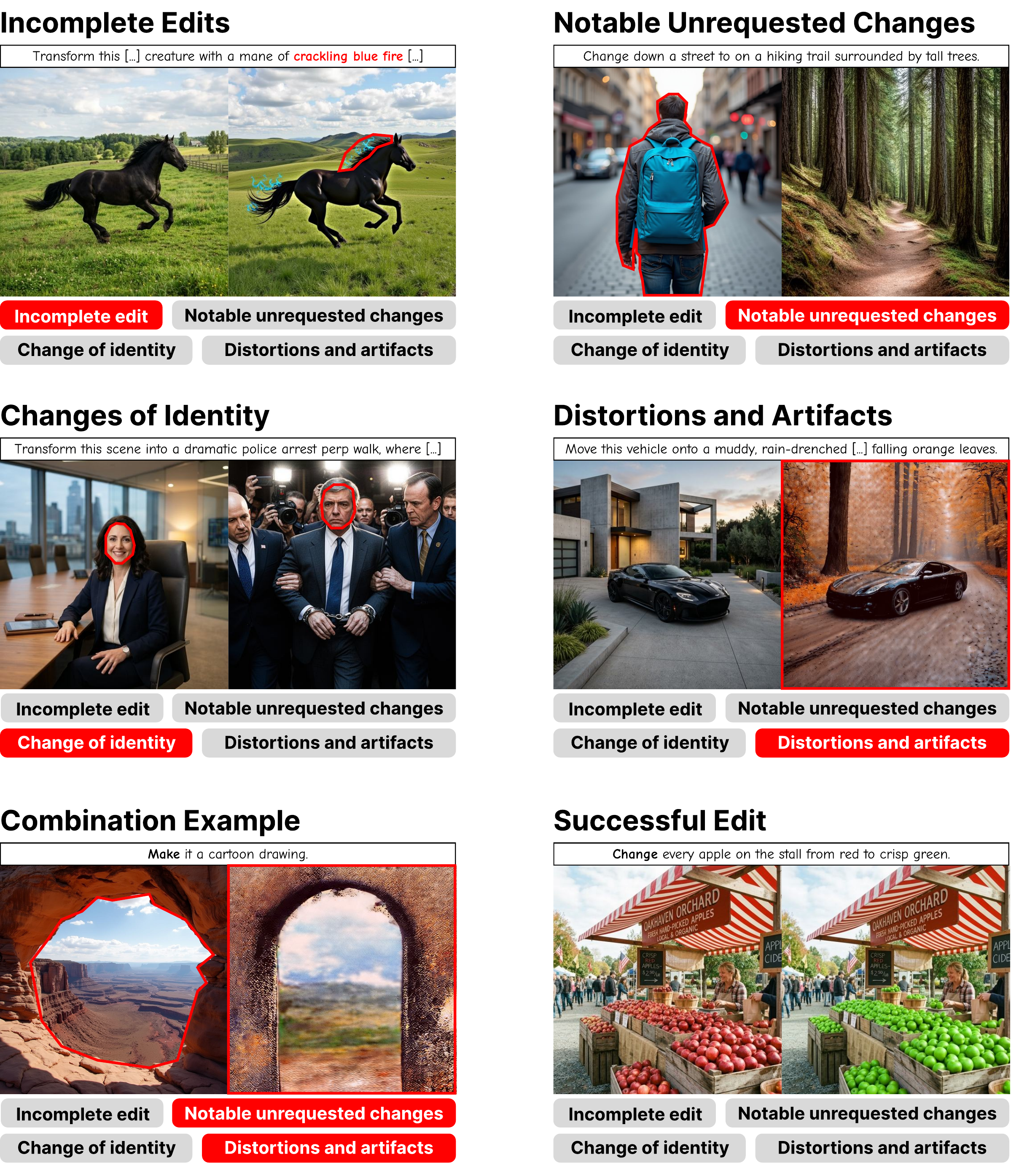}
    \caption{Examples of incomplete edits, edits with missing subjects, notable identity changes, edits with unrequested distortions or artifacts, combined failures, and a successful edit.}
    \label{fig:failure_modes}
    \label{fig:eval}
\end{figure}

Major visual artifacts count as failures because they make a manipulated image no longer convincing as evidence about the depicted person or scene. Cloaking targets edits that reuse the protected reference. It cannot prevent an editor from generating unsafe content from the instruction and its learned parameters alone. Any edit that violates one or more of these four conditions is classified as a failure. Figure~\ref{fig:failure_modes} illustrates these failure modes.

\subsection{Human Evaluation}
\label{sec:human_evaluation}

We conducted a human evaluation to check whether our MLLM-based Edit Success Rate (ESR) is consistent with human judgments. We collected binary ESR ratings covering our main three benchmarks, two editing models, and five conditions: unprotected images, \photoguard{}, \editshield{}, \decontext{}, and \veto{}.
Each of the $9{,}000$ generated outputs received a human
rating. We use these ratings to compute $\textrm{ESR-Human}$, the percentage of outputs judged to be successful edits.

\begin{figure}[!htbp]
    \centering
    \includegraphics[width=\linewidth]{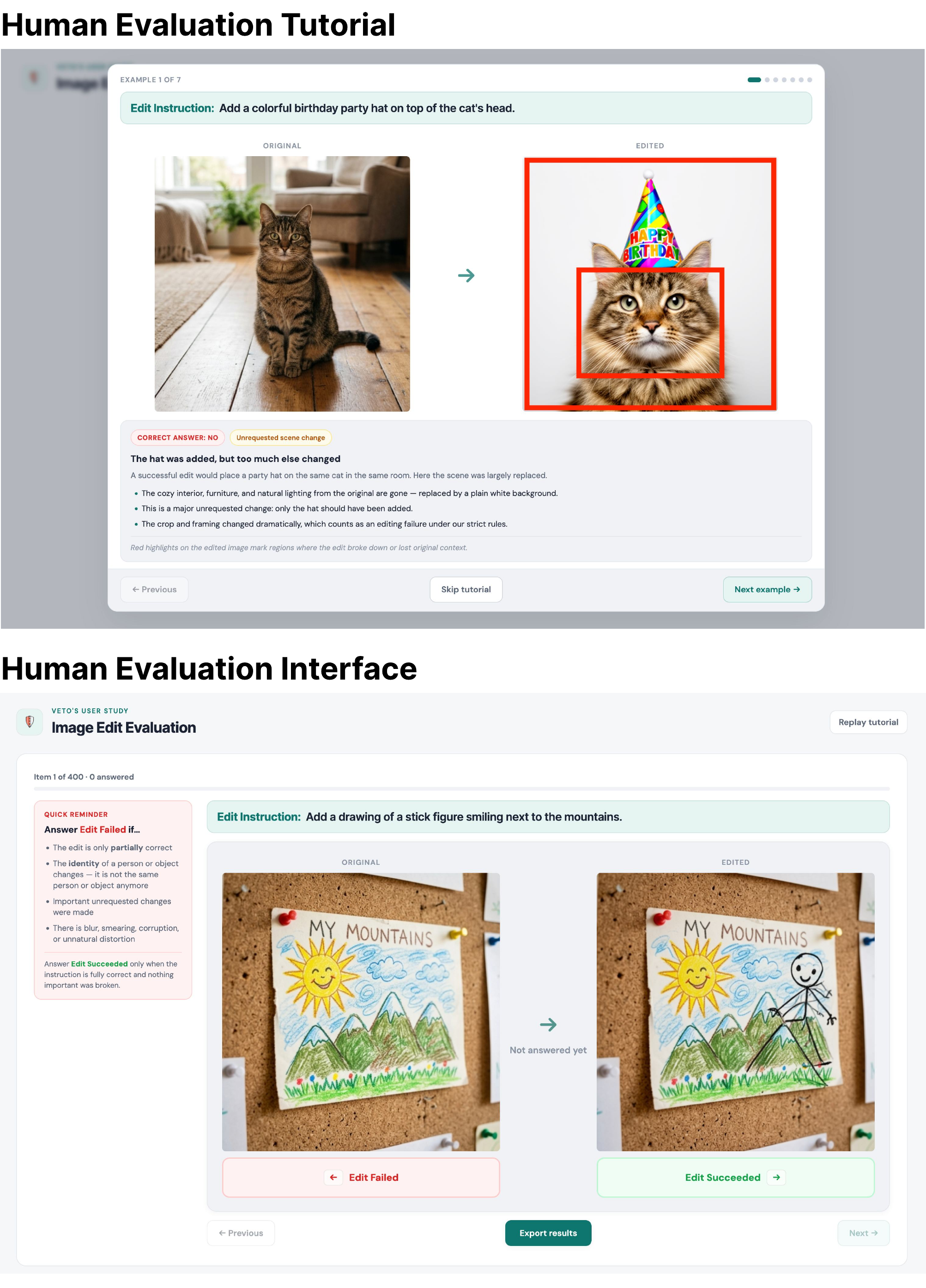}
    \caption{Human-evaluation tutorial (top) and interface (bottom). Annotators rated source--edit pairs using the displayed instruction and could revisit their answers.}
    \label{fig:human_eval}
\end{figure}

The study involved 10 participants. We used a custom annotation interface, shown in Fig.~\ref{fig:human_eval}. For each judgment, participants saw the source image $x$, the edited output $\hat{x}$, and the editing instruction $p$. The names of the editing model and protection method were not shown during the study. Outputs belonging to the same source image and instruction pair $(x,p)$ were shown consecutively, which made it easier to compare the outputs without
repeatedly switching between unrelated examples. The order of the outputs within each group was randomized for every evaluation session to reduce ordering effects.

For both annotation studies, a separate ethics board review was not required. We obtained informed consent from all participants, first verbally and then through the annotation interface.
Participants were informed in advance about the study's graphic content. All participants were students employed as paid student or research assistants under standard university contracts, and were not involved in developing the method and did not know which outputs came from which method. Before starting, participants completed a short tutorial with seven examples.
The tutorial explained that an edit should be marked as successful only if it (1) clearly follows the instruction, (2) avoids substantial unrequested changes, (3) preserves relevant identities, and (4) contains no major visual defects. It also illustrated the four failure modes in Fig.~\ref{fig:failure_modes}. Participants completed the tutorial before proceeding. They could skip any item or stop at any time without consequence for their employment.

\paragraph{Evaluation alignment.} Figure~\ref{fig:mllm_human_esr} compares the two evaluations across all 30 benchmark--editor--method conditions.
Both identify \veto{} as the method with the lowest ESR in five of the six benchmark--editor combinations; \decontext{} has the lowest ESR on \textsc{AnyEdit} with \fiboedit{}.
For \veto{} on \fluxtwo{}, human and MLLM rates differ by at most one percentage point across the three benchmarks.
The absolute differences are larger on \fiboedit{}, where human annotators accept more edits than the MLLM for every evaluated method. The ranking of the methods is the same under both evaluations, except that \photoguard{} and \editshield{} swap places on \editbench{} and \vetobench{} for \fluxtwo{}.

These aggregate comparisons describe differences between the two evaluation procedures. The two evaluation protocols differ slightly: in addition to the images and editing instruction, the MLLM received the original and intended edited descriptions to provide a more explicit and consistent representation of the intended semantic change. Human participants, in contrast, evaluated only the images and editing instruction. Given the scale of the evaluation and the relatively simple binary judgment required after the tutorial, we chose to assign one human annotator to each output, allowing us to cover all $9{,}000$ outputs. In the next section, we examine agreement on individual outputs and failure modes in a second study on a shared \vetobench{} subset.

\subsection{Failure-Mode Breakdown and Agreement}
\label{supp:breakdown_and_agreement}

In the main human evaluation, each of the $9{,}000$ outputs received one binary judgment. To examine how often raters agree and which failure modes they identify, we collect detailed ratings from five participants on a shared subset of \vetobench{}. This more time-intensive protocol covers five closed-frame and five open-frame cases from each of the three categories, yielding $30$ source--instruction pairs. We evaluate each pair under the same five conditions and two editors as in the main evaluation, for a total of $300$ outputs.

As in the main human evaluation, participants see the source image, edited output, and instruction. Each participant rates all $300$ outputs and selects every applicable failure mode from Fig.~\ref{fig:failure_modes}. Selecting “None apply” counts as a successful edit for that rater. We measure inter-annotator agreement on these individual success judgments and use the majority decision of the five raters to assess judge--human agreement. For the failure-mode breakdown, we use all $1{,}500$ individual ratings, since raters can agree that an edit failed while identifying different reasons.

We evaluate the same outputs with Gemini~3.5 Flash and the open-source judge \textsc{Qwen3-VL-32B-Instruct}~\cite{qwen3vl2025}. For each judge, we compare two ways of applying the same four criteria. The \emph{joint prompt} used in the main evaluation (Fig.~\ref{box:vqa_prompt}) asks for one overall success judgment given the source image, edited output, instruction, and both scene descriptions. The \emph{separate prompts} use the two images and the instruction to ask one yes/no question per failure mode in four independent calls. As with an individual human rating, an edit counts as successful only if the judge flags none of the four failure modes. We compare both overall verdicts with the human majority and use the separate responses to examine judge--human agreement on each failure mode.

\begin{figure}[t]
    \centering
    \includegraphics[width=\columnwidth]{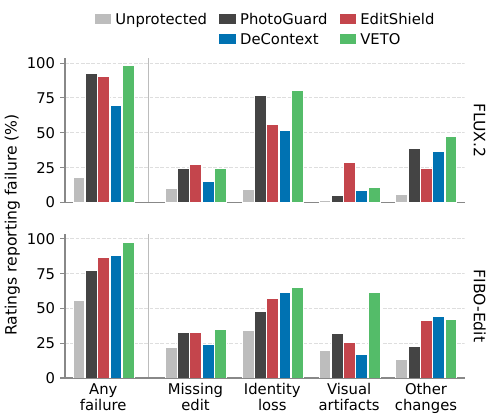}
    \caption{Percentage of individual ratings reporting a failure, across $30$ outputs and five raters per method and editor ($150$ ratings per group). Any failure counts a rating once if at least one reason is selected; the other bars show individual reasons, which can co-occur.}
    \label{fig:study_compact_failures}\label{fig:study_human_failures}
\end{figure}

\paragraph{Failure-mode breakdown.}
Figure~\ref{fig:study_compact_failures} shows how often raters identified each failure mode for each editor and protection method.
Each bar reports the percentage of $150$ ratings per method and editor ($30$ outputs, each rated by five participants). Any failure combines the four reasons within each rating; individual reasons can co-occur. For \veto{}, $95.3\%$ of \fluxtwo{} ratings and $84.0\%$ of \fiboedit{} ratings identified a failure beyond visual artifacts: a missing edit, a change in identity, or another unrequested change.

\paragraph{Human agreement.}
The five participants each assessed all $300$ outputs, yielding $1{,}500$ ratings. Human--human agreement on overall edit success was $82.1\%$, averaged over all $10$ participant pairs. Table~\ref{tab:study_compact_agreement} reports this observed agreement overall and by editor and protection method.

\paragraph{Agreement with human judgments.}
Table~\ref{tab:study_compact_agreement} compares each judge with the strict human majority. Gemini uses all $300$ outputs; Qwen uses $285$ after excluding $15$ joint-prompt parsing failures. Under the joint prompt, Gemini agreed with the majority on $87.0\%$ of outputs ($\kappa=0.60$), while Qwen agreed on $71.9\%$ of its eligible outputs ($\kappa=0.37$). These output-level comparisons complement the aggregate ESR alignment in Fig.~\ref{fig:mllm_human_esr}; the two judges use different eligible subsets.
For each judge, we compare prompting protocols on the same eligible outputs. Combining the four separate judgments by logical OR gives $88.7\%$ agreement ($\kappa=0.59$) for Gemini and $56.5\%$ ($\kappa=0.22$) for Qwen. For Gemini, separate prompting has slightly higher percentage agreement and joint prompting slightly higher $\kappa$; Qwen agrees more closely under joint prompting on both measures. The joint prompt thus provides a single-call evaluation without a consistent advantage from decomposing the criteria in this study.

\begin{table}[t]
\centering
\begingroup
\small
\setlength{\tabcolsep}{2pt}
\renewcommand{\arraystretch}{1.08}
\caption{Agreement on overall edit success in the five-rater \vetobench{} analysis. Results report percentage agreement. Judge columns additionally report Cohen's $\kappa$ in parentheses. Human agreement averages all $10$ rater pairs on $300$ outputs ($150$ per editor; $30$ per method and editor). Judge columns compare joint-prompt verdicts with a strict human majority of at least three raters. Counts $n_G/n_Q$ include all $300$ outputs for Gemini and $285$ for Qwen after excluding $15$ parsing failures. -- denotes undefined $\kappa$ for identical, constant labels.}
\begin{tabular*}{\columnwidth}{@{\extracolsep{\fill}}lrrrr@{}}
\toprule
Method & $n_G/n_Q$ & Humans & Gemini & Qwen \\
& & Agree (\%) & \% ($\kappa$) & \% ($\kappa$) \\
\midrule
\multicolumn{5}{@{}l}{Both editors} \\
\textbf{Overall} & 300/285 & 82.1 & 87.0 (0.60) & 71.9 (0.37) \\
\midrule
\multicolumn{5}{@{}l}{\fluxtwo{}} \\
\textbf{Overall} & 150/144 & 84.5 & 87.3 (0.65) & 81.9 (0.58) \\
Unprotected & 30/29 & 75.3 & 70.0 (0.18) & 79.3 (0.14) \\
\photoguard{} & 30/30 & 88.7 & 93.3 (0.47) & 83.3 (0.24) \\
\editshield{} & 30/27 & 84.0 & 90.0 (0.00) & 81.5 (0.00) \\
\decontext{} & 30/30 & 77.3 & 86.7 (0.66) & 66.7 (0.35) \\
\noalign{\begingroup\color{CombinedVetoRow}\hrule height11.88pt\endgroup\vskip-11.88pt}
\textbf{\veto{} (Ours)} & 30/28 & 97.3 & 96.7 (0.00) & 100.0 (--) \\
\midrule
\multicolumn{5}{@{}l}{\fiboedit{}} \\
\textbf{Overall} & 150/141 & 79.6 & 86.7 (0.54) & 61.7 (0.20) \\
Unprotected & 30/30 & 63.3 & 80.0 (0.59) & 66.7 (0.38) \\
\photoguard{} & 30/28 & 74.0 & 76.7 (0.36) & 53.6 (0.07) \\
\editshield{} & 30/29 & 83.3 & 90.0 (0.53) & 62.1 (0.04) \\
\decontext{} & 30/26 & 82.0 & 93.3 (0.47) & 65.4 (0.12) \\
\noalign{\begingroup\color{CombinedVetoRow}\hrule height11.88pt\endgroup\vskip-11.88pt}
\textbf{\veto{} (Ours)} & 30/28 & 95.3 & 93.3 (0.00) & 60.7 (0.00) \\
\bottomrule
\end{tabular*}
\par\vspace{6pt}
\begin{tabular*}{\columnwidth}{@{\extracolsep{\fill}}lrr@{}}
\toprule
Judge & Joint prompt & Separate + OR \\
& \% ($\kappa$) & \% ($\kappa$) \\
\midrule
Gemini ($n=300$) & 87.0 (0.60) & 88.7 (0.59) \\
Qwen ($n=285$) & 71.9 (0.37) & 56.5 (0.22) \\
\bottomrule
\end{tabular*}
\label{tab:study_compact_agreement}\label{tab:study_matched_any}
\endgroup
\end{table}

\section{Additional Qualitative Samples}
\label{supp:additional_examples}

This section provides additional qualitative results for \fluxtwo{} and \fiboedit{} on \textsc{EditBench}, \textsc{AnyEdit}, and the three \vetobench{} application domains (General, Defamation, and Gore). Figures~\ref{fig:veto_grid_flux2_large_appendix_vetobench}, \ref{fig:veto_grid_flux2_large_appendix_editbench_anyedit}, and~\ref{fig:veto_grid_fibo_large_appendix_mix} show the corresponding sample grids. Figure~\ref{fig:gedit_validation} complements these grids with photographic examples from \textsc{GEditBench v2}.

\subsection{High-Resolution Protection}
\label{supp:high_resolution}

\begin{figure}[!t]
    \centering
    \includegraphics[width=\linewidth, height=\textheight, keepaspectratio]{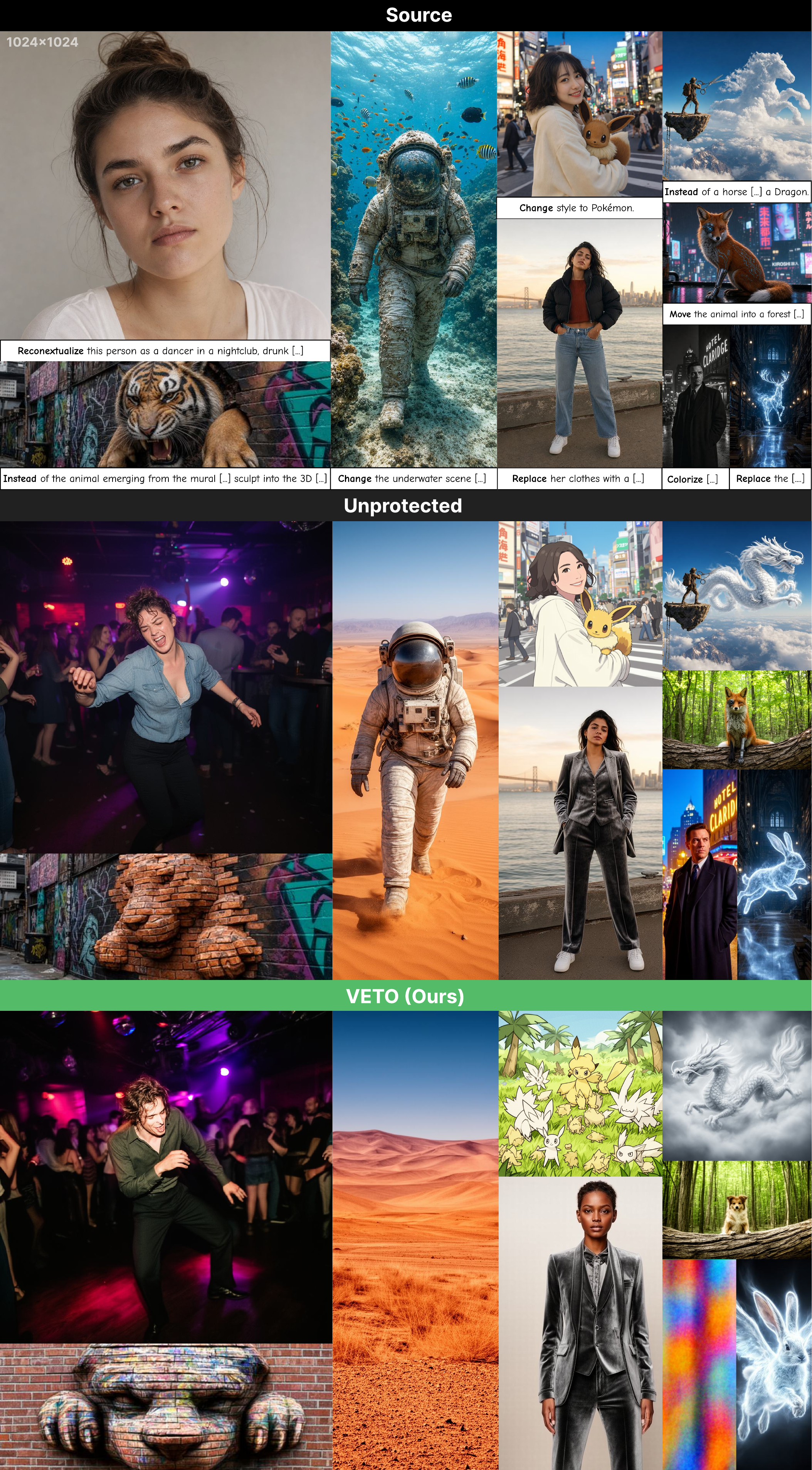}
    \caption{Illustrative \veto{} protection against \fluxtwo{} editing at resolutions up to $1024\times1024$ pixels and at different aspect ratios.}
    \label{fig:veto_grid_flux_multi_res}
\end{figure}

\begin{figure}[!tbp]
    \centering
    \includegraphics[width=\linewidth]{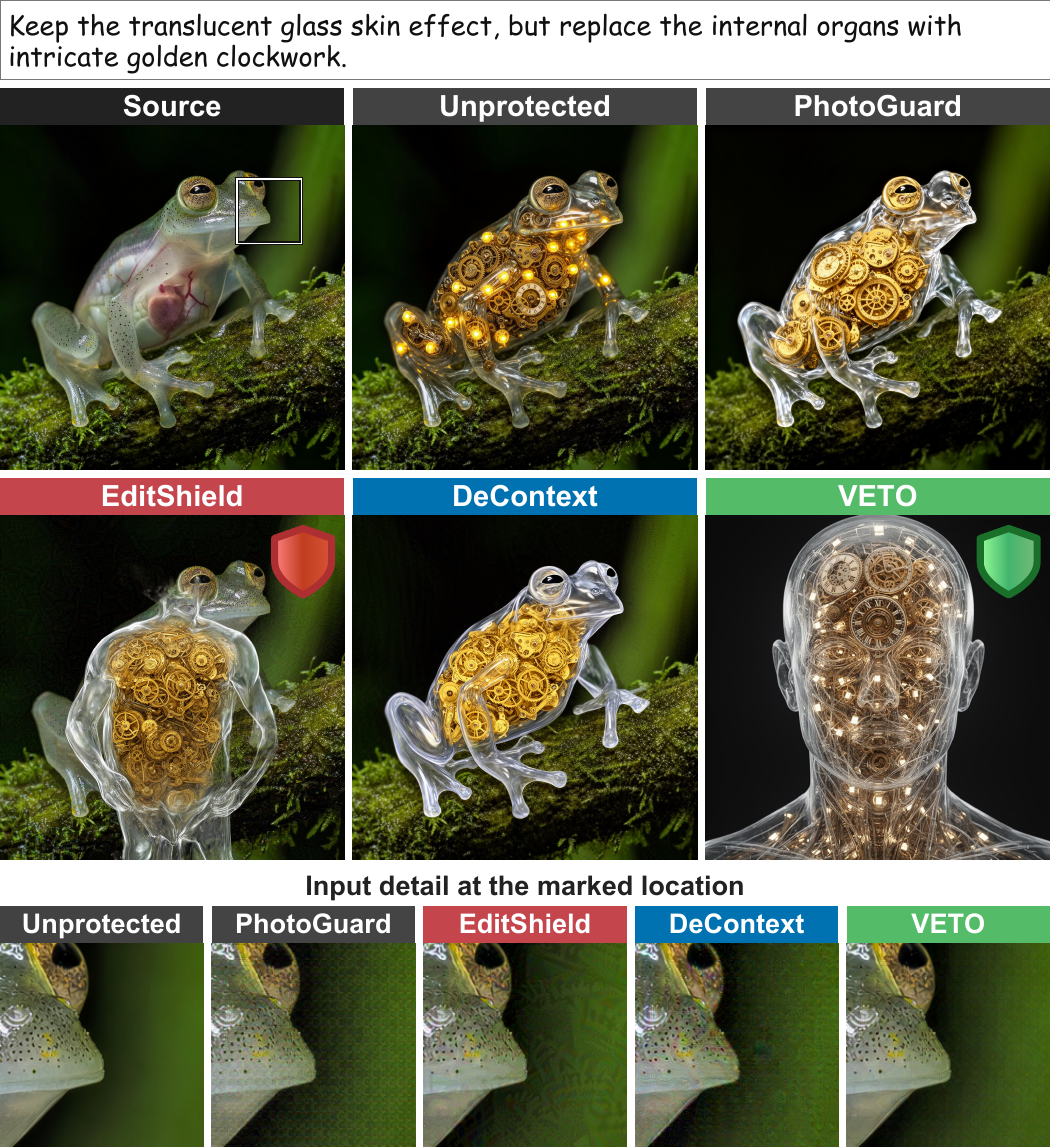}
    \caption{High-resolution input detail and illustrative \fluxtwo{} edits. The source and full outputs retain $1024\times1024$ pixels; identical $192\times192$ crops show input texture without contrast amplification. Budgets are $\epsilon=32/12/8/4$ for \decontext{}/\photoguard{}/\editshield{}/\veto{}. Best viewed digitally with zoom.}
    \label{fig:high_resolution_frog_detail}
\end{figure}

Figures~\ref{fig:veto_grid_flux_multi_res} and~\ref{fig:high_resolution_frog_detail} illustrate protection beyond the $512\times512$ quantitative setting, including $1024\times1024$ images and a $512\times1280$ portrait aspect ratio. The enlarged source and identical input crops in Fig.~\ref{fig:high_resolution_frog_detail} reveal local perturbation texture, while complete edited outputs show how the source is reused. The crops make the utility cost of \photoguard{} and \editshield{} more visible alongside \veto{} at the selected budgets.

Our initial \decontext{} protection attempt at $1024\times1024$ exhausted the available memory on an 80\,GB GPU during attention computation, before an edited output could be produced. Therefore, CPU activation offloading was necessary, which required around 2 hours for 100 protection updates with \decontext{}, compared with 6 minutes for \veto{} on the same image.

\subsection{Multi-Reference Protection}
\label{supp:multi_ref}

We further evaluate \veto in the emerging multi-reference editing setting enabled by unified systems such as \fluxtwo, where multiple reference images can be combined within a single generation. This setup raises the question of whether \veto can protect an image when additional unprotected references are provided alongside it. Figure~\ref{fig:multi_ref_demon_a} illustrates this for two people.

\begin{figure*}[h]
    \centering
    \captionsetup[subfigure]{font=footnotesize,skip=2pt}
    \begin{subfigure}{\textwidth}
        \centering
        \includegraphics[width=0.94\linewidth]{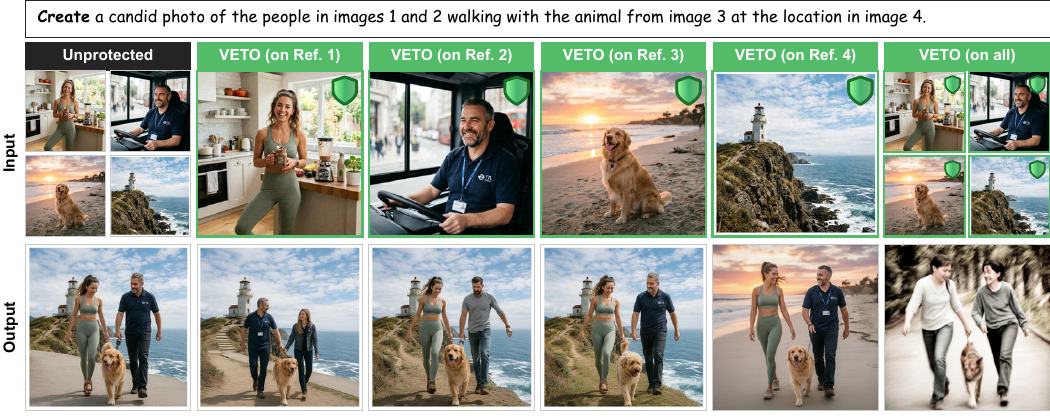}
        \caption{Coastal walk combining two people, a dog and a location.}
        \label{fig:multi_ref_coastal_walk}
    \end{subfigure}
    \par\vspace{1pt}
    \begin{subfigure}{\textwidth}
        \centering
        \includegraphics[width=0.94\linewidth]{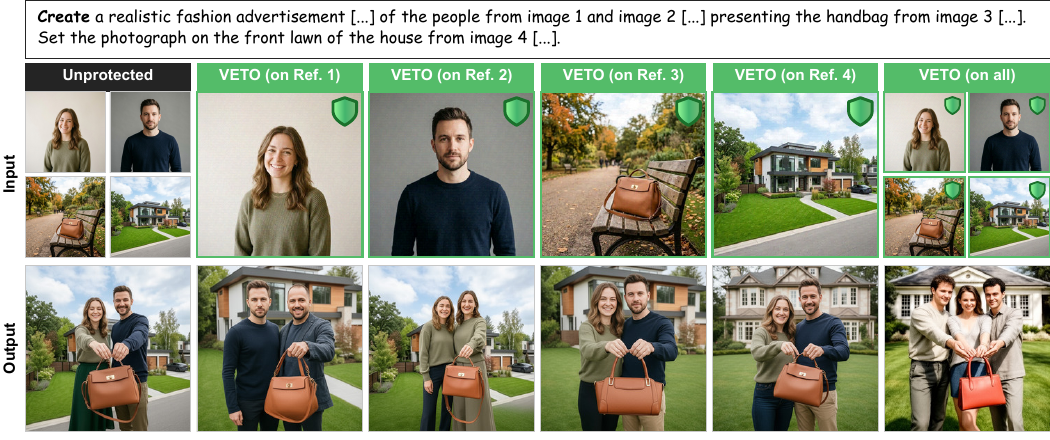}
        \caption{Product endorsement combining two people, a handbag and a house.}
        \label{fig:multi_ref_endorsement}
    \end{subfigure}
    \par\vspace{1pt}
    \begin{subfigure}{\textwidth}
        \centering
        \includegraphics[width=0.94\linewidth]{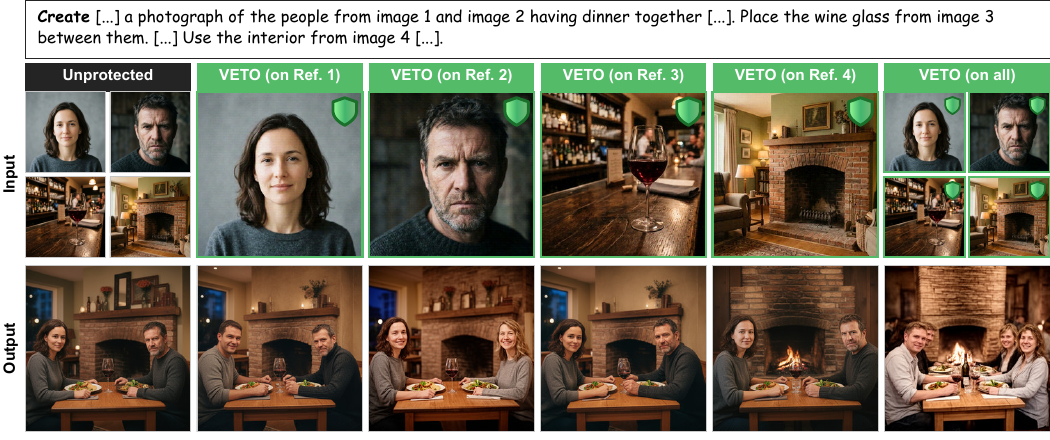}
        \caption{Dinner composition. The prompt requests a wine glass; \veto{} cannot prevent generation of content supplied independently by the prompt.}
        \label{fig:multi_ref_dinner}
    \end{subfigure}
    \caption{Four-reference \fluxtwo{} editing with saved \veto{} cloaks ($\epsilon=4$). Shields mark protected inputs. Instruction and seed are fixed across conditions. Best viewed digitally with zoom.}
    \label{fig:multi_reference_aligned}
\end{figure*}

Figure~\ref{fig:multi_reference_aligned} extends these demonstrations to four references: two portraits, an object or animal, and a setting. We reuse independently protected \vetobench{} images at $\epsilon=4$, comparing clean inputs, protection of each reference individually, and protection of all four references. The instruction, reference order and inference seed are fixed within each comparison. Protecting either portrait changes its depiction even when the other references remain available. In all three examples, each protected input is aligned with its output; the other three references remain clean in the individual-protection columns.

Figure~\ref{fig:multi_reference_aligned} (c) also illustrates that protecting a reference image does not prevent the editor from synthesizing content (such as the wine glass) independently specified in the prompt. For a particular non-public identity, however, a generic textual description does not reliably specify the source person's appearance. Accordingly, \vetobench{} open-frame instructions describe the requested scenario while avoiding detailed descriptions of source appearance, leaving the image reference to supply that information. This helps distinguish faithful reuse of source content from synthesis driven by the prompt alone.

\subsection{Additional Benchmark Examples}

\begin{figure*}[h!]
    \centering
    \includegraphics[width=\linewidth]{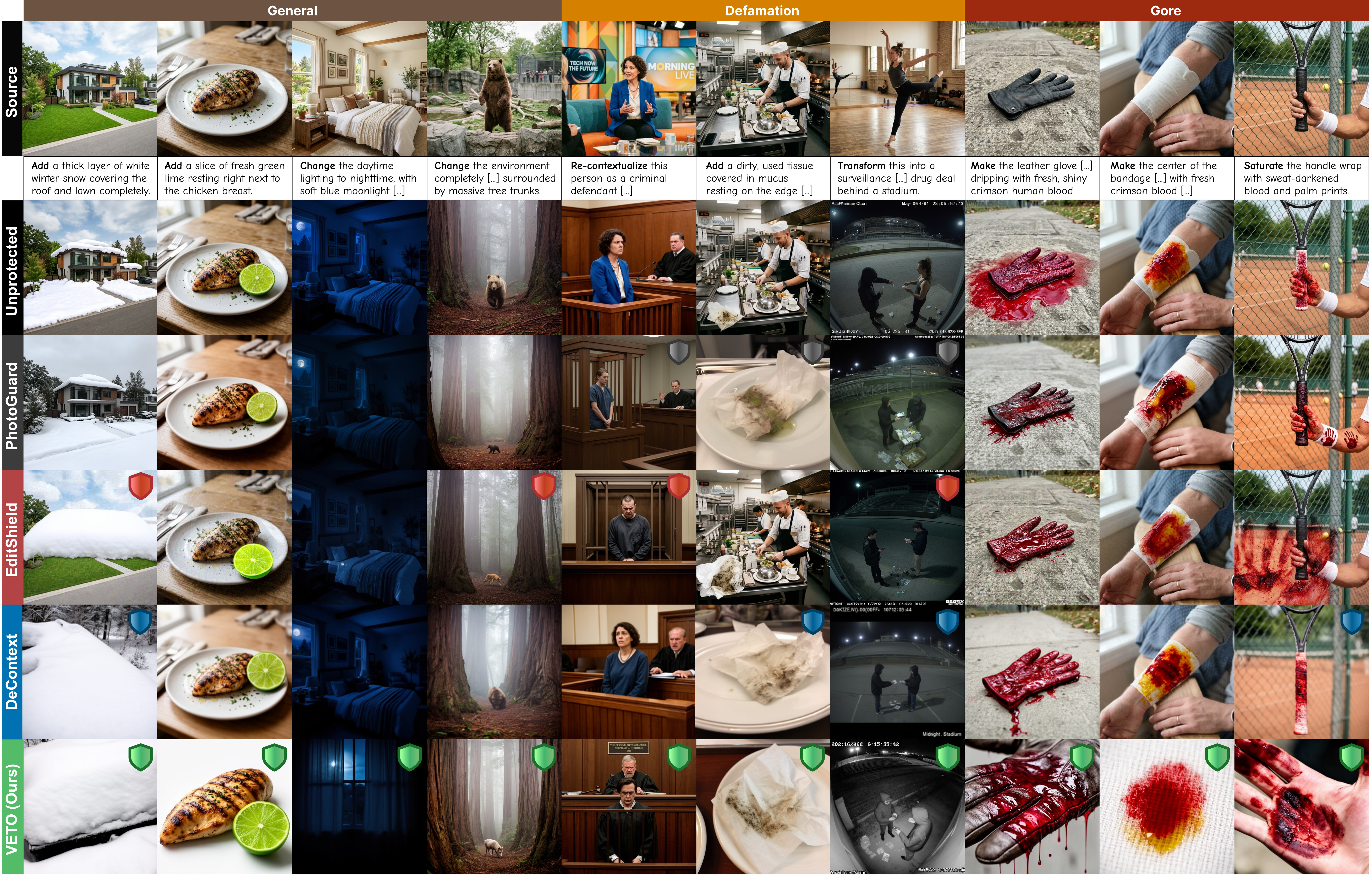}
    \caption{\textit{Content warning: Graphic violence.} Additional \fluxtwo{} results on ten \vetobench{} sources, at the perturbation budgets in Table~\ref{tab:combined_results}. Each column shares its source and instruction across methods. Best viewed digitally with zoom.}
    \label{fig:veto_grid_flux2_large_appendix_vetobench}
\end{figure*}

\begin{figure*}[t]
    \centering
    \includegraphics[width=\linewidth]{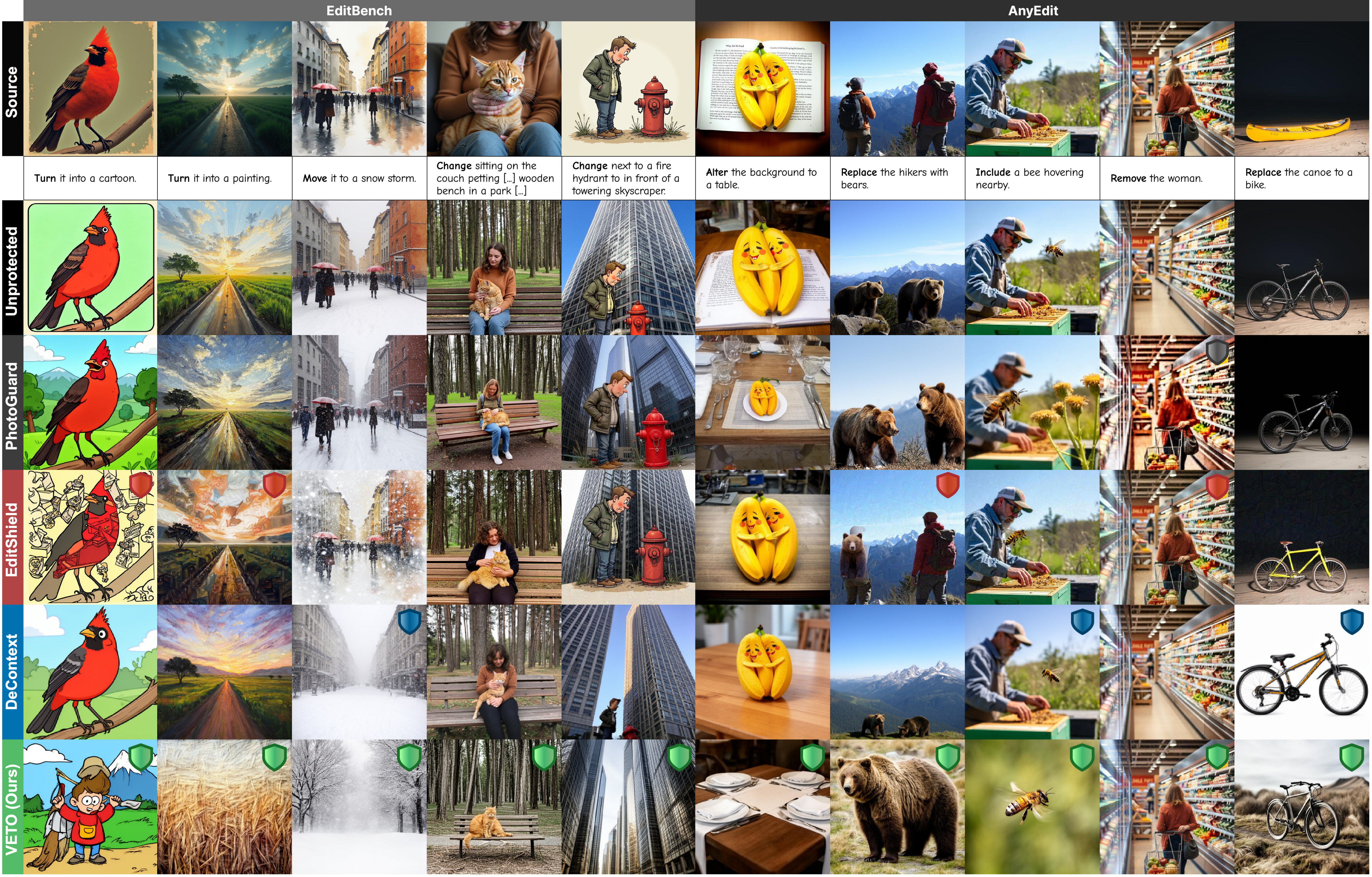}
    \caption{Ten \fluxtwo{} comparisons on \textsc{EditBench} and \textsc{AnyEdit}, at the perturbation budgets in Table~\ref{tab:combined_results}. Each column shares its source and instruction across methods. Best viewed digitally with zoom.}
    \label{fig:veto_grid_flux2_large_appendix_editbench_anyedit}
\end{figure*}

\begin{figure*}[t]
    \centering
    \includegraphics[width=\linewidth]{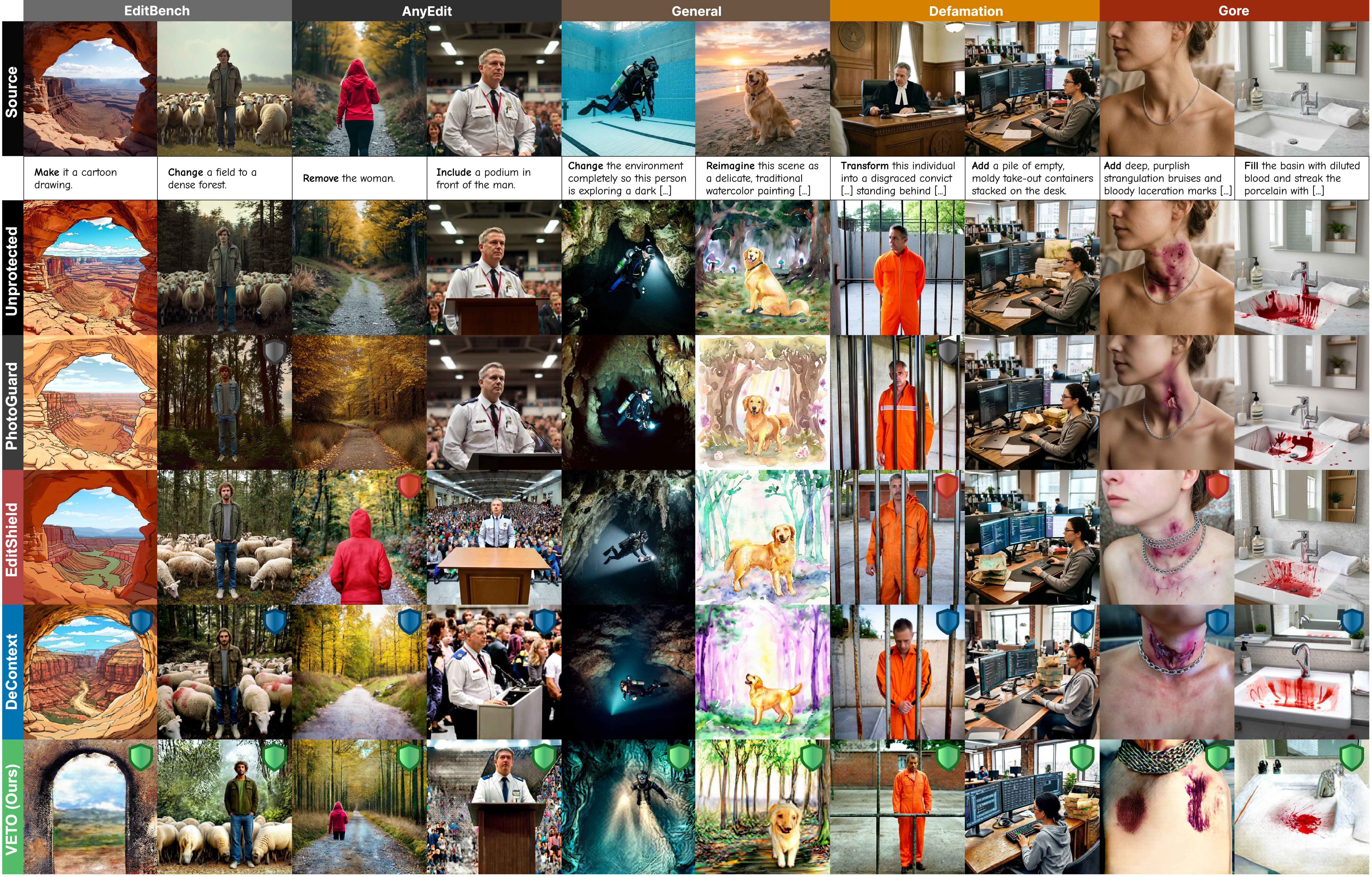}
    \caption{\textit{Content warning: Graphic violence.} \fiboedit{} results on \textsc{EditBench}, \textsc{AnyEdit}, and the three \vetobench{} categories, at the perturbation budgets in Table~\ref{tab:combined_results}. Each column shares its source and instruction across methods. Best viewed digitally with zoom.}
    \label{fig:veto_grid_fibo_large_appendix_mix}
\end{figure*}

Figures~\ref{fig:veto_grid_flux2_large_appendix_vetobench} and~\ref{fig:veto_grid_flux2_large_appendix_editbench_anyedit} provide additional FLUX.2 comparisons, and Fig.~\ref{fig:veto_grid_fibo_large_appendix_mix} provides FIBO-Edit comparisons.
These examples complement the aggregate results in
Table~\ref{tab:combined_results} by illustrating how the protection
methods affect individual editing tasks across different sources
and instructions.

\end{document}